\documentclass[10pt,twocolumn,letterpaper]{article}

\usepackage{wacv}              

\usepackage{multirow} 
\usepackage{verbatim}
\usepackage{adjustbox}
\definecolor{wacvblue}{rgb}{0.21,0.49,0.74}
\usepackage[pagebackref,breaklinks,colorlinks,allcolors=wacvblue]{hyperref}

\title{Interpretable Face Contrast Modeling for Smartphone Portrait Rendering}

\author{
\vspace{-0.5em}
Yujin Cho$^{1,2}$ \qquad
Beno\^{i}t Pochon$^{2}$ \qquad
Jean-Michel Morel$^{3}$ \qquad
Gabriele Facciolo$^{1}$\\[0.6em]
$^{1}$Universit\'e Paris-Saclay, CNRS, ENS Paris-Saclay, Centre Borelli, France\\
$^{2}$DXOMARK, France \qquad
$^{3}$Lingnan University, Hong Kong SAR, China\\
{\ttfamily\scriptsize
\{yujin.cho,gabriele.facciolo\}@ens-paris-saclay.fr \quad
bpochon@dxomark.com \quad
jeanmichelmorel@LN.edu.hk
}\\
{\ttfamily\footnotesize
\url{https://github.com/nuniniyujin/FaceContrast}
}
}

\begin{document}
\maketitle
\begin{abstract}
The quality of smartphone portrait photography has become increasingly important to both manufacturers and users. Recent devices use complex computational pipelines to improve facial rendering under varied lighting. These pipelines vary widely across manufacturers, so the same portrait can look noticeably different across smartphones. These differences affect multiple aspects of facial appearance, making perceptual preference difficult to characterize with a single measure. To address this, we propose an interpretable framework that decomposes face contrast into three components: overall facial tonal separation, eye‑region contrast, and micro‑contrast of fine skin details.
We formulate dedicated metrics for each component and validate them via pairwise human judgments, demonstrating that our approach correlates more closely with human contrast perception than traditional contrast measures. To support this analysis, we introduce a realistic mannequin‑based dataset designed for systematic facial contrast evaluation, comprising 508 full‑resolution captures from 29 smartphones under diverse lighting, with region‑specific masks and extensive pairwise perceptual annotations. As a downstream analysis, we show that combining the three component‑wise metrics provides a meaningful basis for modeling balanced facial contrast preference in a controlled setting, and we present a preliminary evaluation on a smaller real‑portrait dataset indicating that insights derived from the mannequin study can extend to real images. Our work provides a controlled first step toward defining and evaluating face contrast in portrait imaging.
\end{abstract}    
\section{Introduction}
\label{sec:intro}

\begin{figure*}[t]
    \centering
    \includegraphics[width=0.9\textwidth]
    {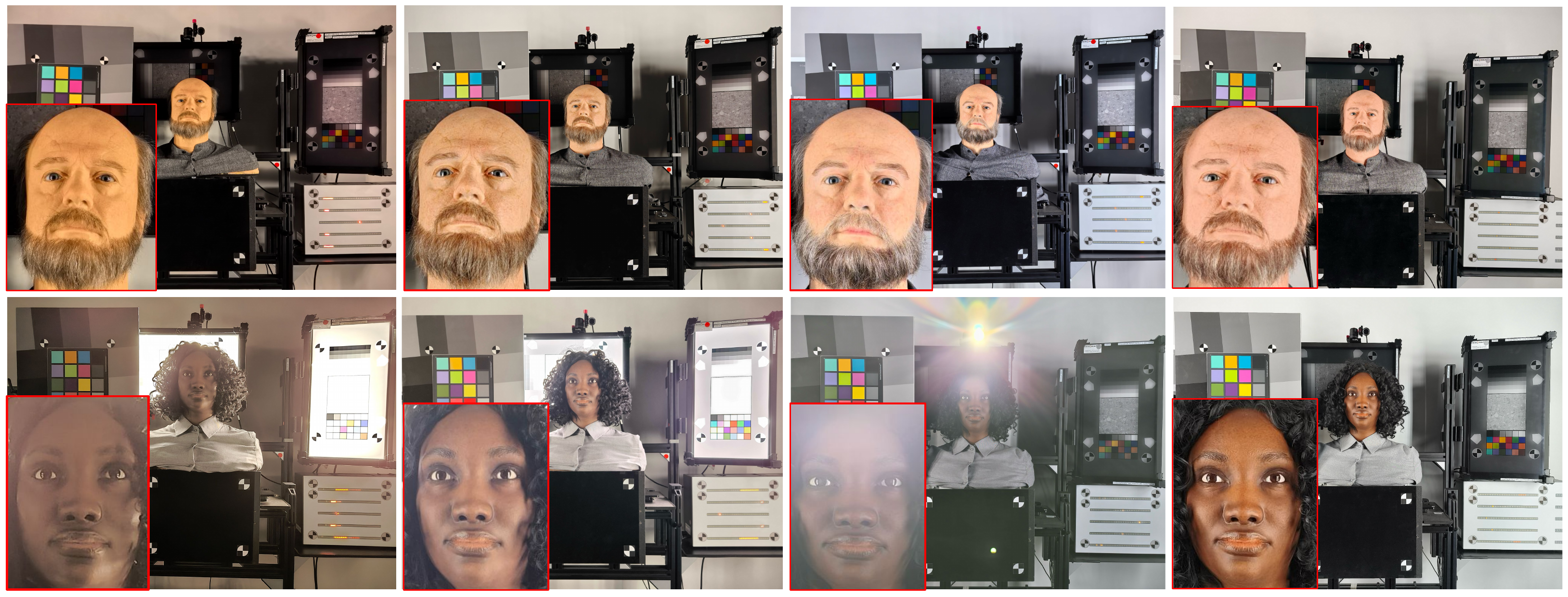} 
    \vspace{0.1mm}
    \noindent
    \makebox[0.9\textwidth][c]{
        \parbox[t]{0.2\textwidth}{\centering \small (a) 5 lux}
        \hfill
        \parbox[t]{0.2\textwidth}{\centering \small (b) 100 lux}
        \hfill
        \parbox[t]{0.2\textwidth}{\centering \small (c) 1000 lux}
        \hfill
        \parbox[t]{0.2\textwidth}{\centering \small (d) 10000 lux}
    }  
    \caption{Examples from the proposed dataset for the two mannequin subjects, Eugene and Diana, shown under four illumination levels and multiple backlight conditions ($\Delta \mathrm{EV} = 0, 4, 7,$ and $9$). Each sub-image shows the full controlled capture setup together with an inset of the face crop, highlighting how facial contrast rendering varies across devices and lighting conditions.}
    \label{fig:figure1}
    \vspace*{-1em}
\end{figure*}

\begin{figure}[t]
\centering
\includegraphics[width=0.9\columnwidth]{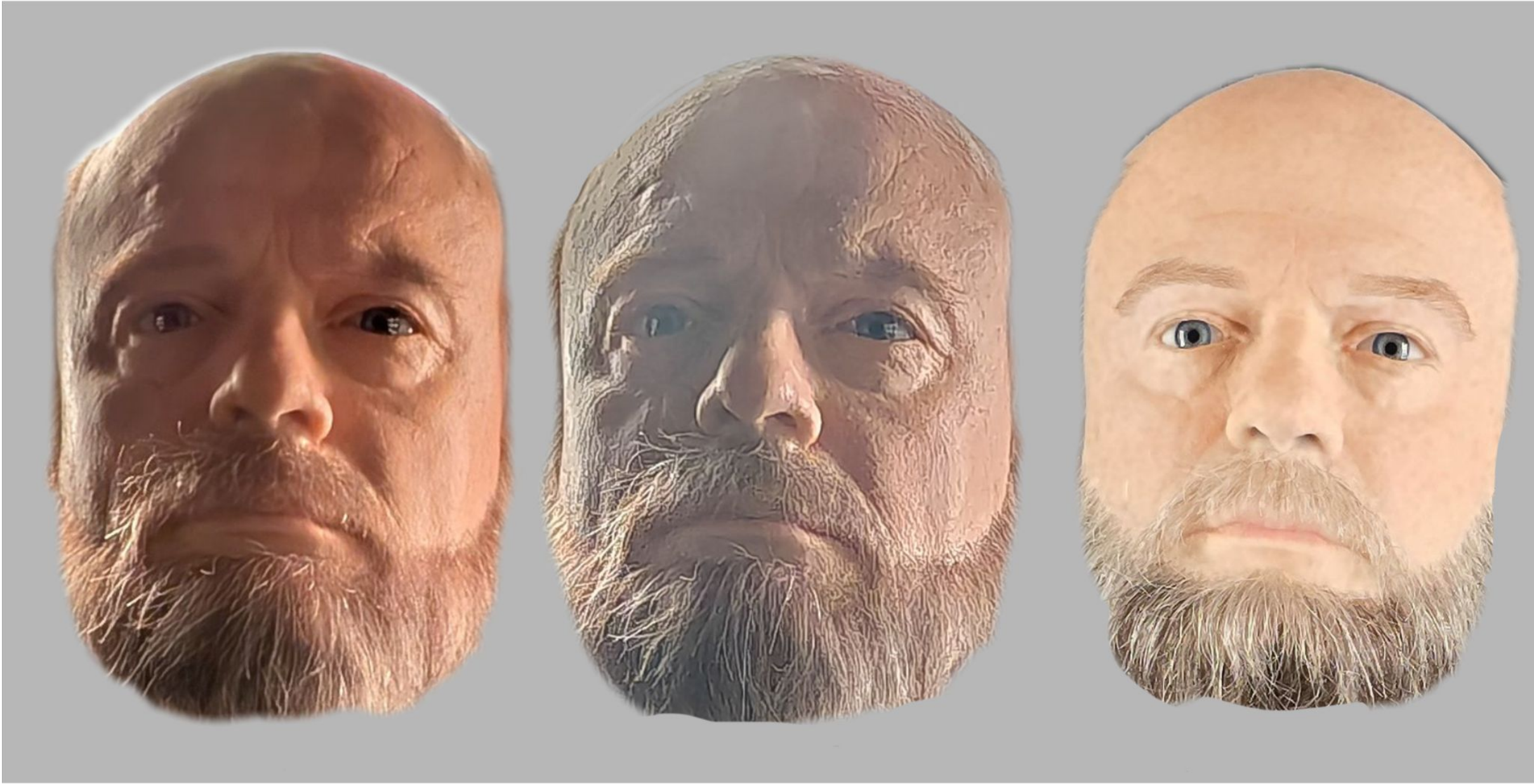}

\vspace{0.2em}
\makebox[0.9\columnwidth][c]{%
  \parbox{0.3\columnwidth}{\centering \scriptsize (a) 5 lux, $\Delta \mathrm{EV}=9$}%
  \parbox{0.3\columnwidth}{\centering \scriptsize (b) 5 lux, $\Delta \mathrm{EV}=9$}%
  \parbox{0.3\columnwidth}{\centering \scriptsize (c) 10000 lux, $\Delta \mathrm{EV}=0$}%
}

\caption{Examples of face without background captured at two illumination levels (5 and 10000 lux) with different smartphones. Facial rendering varies with lighting and device-specific processing, causing noticeable differences in face contrast.}
\label{fig:face_rendering}
\vspace*{-1.5em}
\end{figure}

With the recent growth of smartphone photography, portrait imaging has become an increasingly important application. Modern devices rely on complex Image Signal Processing (ISP) pipelines to improve facial rendering under a wide range of lighting conditions~\cite{delbracio2021mobile,chahine2023image}. These pipelines combine operations such as multi-frame fusion, tone mapping, denoising, sharpening, and face-specific enhancement, 
and they vary widely across manufacturers~\cite{delbracio2021mobile,morikawa2021image,monod2021analysis}.
As a result, the same portrait can appear noticeably different across devices, while the underlying processing remains largely a black box. 
These differences are further accentuated in portrait photography, where humans are particularly apt at distinguishing subtle differences in faces~\cite{tanaka1993parts,maurer2002many,kanwisher1997fusiform}.
Together, these processing and perceptual factors make portrait quality difficult to explain with a single perceptual factor. Instead, multiple attributes such as exposure, detail preservation, noise, and local enhancement are all crucial~\cite{chahine2023image,cho2026adapting,ntire26raim_piqa}.
One particularly important attribute is face contrast. It influences dynamic range, mid-tone rendering, local contrast, and fine-detail visibility across the face. In practice, variations in face contrast can change the apparent depth of the face, the clarity of the eyes, and the visibility of the skin texture, even when the overall content of the scene remains the same, as illustrated in Figure~\ref{fig:figure1} and~\ref{fig:face_rendering}. Yet face contrast remains underexplored as a standalone perceptual attribute. Prior work has examined specific portrait quality factors, such as noise and texture~\cite{chahine2023image,cho2026adapting}, but has not isolated face contrast itself as a distinct target of analysis.
This partially stems from the complex evaluation of face contrast in photography.  Whereas general contrast measures are well studied~\cite{michelson1995studies}, face contrast does not have a single widely accepted definition.
Existing work on facial contrast~\cite{russell2009sex,porcheron2013aspects,russell2016facial} has mostly focused on feature-to-skin luminance or color differences, which do not fully capture how contrast is rendered in photographic portraits. Moreover, the appearance of face contrast is strongly affected by imaging conditions and processing, including illumination, tone reproduction, presence of flare and veil, local enhancement, and texture rendering~\cite{peli1990contrast,haun2013perceived}. As a result, face contrast has mostly been studied under idealized conditions, 
limiting its generalization across diverse lighting and smartphone rendering conditions.

Existing approaches do not fully address this gap. General natural image quality assessment methods mainly target global perceptual quality~\cite{yang2022maniqa, madhusudana2021st, ke2021musiq}, whereas works related to face image quality~\cite{babnik2023diffiqa, kolf2024grafiqs} are often primarily designed for recognition utility rather than to understand how facial appearance is rendered. 
Our goal is different: rather than predicting global portrait quality or selecting the more preferred rendering, we aim to define and quantify the perceived amount of face contrast in a way that is interpretable and directly tied to portrait rendering. In this work, we propose to decompose face contrast into three perceptual components: \textit{overall contrast}, \textit{eye contrast}, and \textit{micro-contrast}. \textit{Overall contrast} 
captures coarse tonal separation across the face. \textit{Eye contrast} is similar, but treated separately due to the importance of eyes in face perception~\cite{peterson2012looking, royer2018greater}. Finally, \textit{micro-contrast} captures the visibility of fine facial details in skin regions, focusing on how texture is rendered in the final portrait image rather than on intrinsic skin texture alone. The three components are intended to characterize the perceived amount of contrast in different facial regions and spatial scales. 
Since more contrast is not necessarily better, we further consider balanced contrast, a holistic preference judgment over the relative balance of these components. This downstream analysis examines how the three component-wise measures relate to contrast preference in the controlled mannequin setting. To study these component-wise contrast measures under controlled variations, we build on prior mannequin-based studies of portrait texture quality~\cite{nicolas2021portrait,ventura2024video} and use realistic mannequins to compare facial contrast across devices and lighting conditions. The main contributions of this paper are as follows:

\begin{itemize}    

    \item We introduce a controlled mannequin-based dataset with full-resolution captures, region-specific masks, and pairwise human perceptual annotations, enabling controlled evaluation of face contrast across devices and lighting conditions.
    
    \item We decompose face contrast in smartphone portrait imaging into three perceptual components: overall contrast, eye contrast, and micro-contrast, and propose dedicated metrics to estimate their perceived amounts using pairwise psychophysical judgments.
    
    \item As a downstream analysis on the controlled mannequin dataset, we show that the proposed three-component representation provides a meaningful basis for modeling balanced contrast preference.

    \item We further evaluate the proposed component-wise metrics on real portrait images, providing preliminary evidence of applicability beyond the controlled setting. 
\end{itemize}

\section{Related Work}

Previous work related to our study includes general image quality assessment, portrait quality assessment, face image quality for recognition, and classical contrast modeling. General image quality assessment (IQA) methods aim to predict overall perceptual image quality, either with reference images~\cite{wang2004ssim,mantiuk2005predicting,mantiuk2011hdrvdp,zhang2018unreasonable} or in a no-reference setting. In the no-reference setting, early hand-crafted models rely on natural-scene statistics~\cite{mittal2012making,mittal2012no}, while more recent approaches~\cite{bosse2017deep, zhang2020blind, Su_2020_CVPR, ying2020patches, yang2022maniqa, madhusudana2021st, ke2021musiq,wang2022exploring, agnolucci2024arniqa,chen2024topiq} are learning-based. These learning-based approaches are typically trained on human-rated image-quality datasets~\cite{sheikh2006statistical,ponomarenko2009tid2008,ponomarenko2013color}. More recently, multi-modal large language models (MLLMs) based IQA methods use quality-aware prompting and scoring with vision-language representations~\cite{li2026q,wu2023q,you2025teaching,depictqa_v2}. These methods model global perceptual quality from image content and distortion patterns, but they are not designed to isolate face contrast as a specific perceptual attribute. 
Portrait quality assessment methods~\cite{chahine2023image,chahine2025generalized,chahine2024deep,ntire26raim_piqa} study portrait-specific attributes such as preservation of face detail, exposure, naturalness, noise, and global portrait quality. A related line of work studies face image quality for recognition~\cite{jo2023ifqa, kolf2024grafiqs, babnik2024ediffiqa,babnik2023diffiqa,boutros2023cr}, where the goal is to estimate whether an image is suitable for matching or verification. While both directions are relevant, neither is primarily intended to evaluate face contrast itself.

Contrast has long been studied in imaging, and classical measures such as Michelson contrast, Weber contrast, and RMS contrast remain widely used~\cite{michelson1995studies,moulden1990standard,prins2016psychophysics}. These formulations are well established for controlled stimuli, simple image patterns, and global image statistics. Previous work has also shown that contrast becomes harder to define for complex images, where perception depends on spatial structure and local image content~\cite{peli1990contrast}. This limitation is especially relevant for portraits, where perceived facial contrast depends not only on overall luminance range but also on the spatial distribution of tonal variation and the visibility of fine details. There have also been attempts to define facial contrast more explicitly in color science, cosmetics, and face perception~\cite{russell2009sex,porcheron2013aspects,russell2016facial}. In this literature, facial contrast is  defined as the luminance or color difference between facial features, such as the eyes, lips, or eyebrows, and the surrounding skin, often through adapted Michelson formulations. Although these definitions are a starting point for our study, they present limitations when exploring portrait photos taken with various smartphones under different lighting conditions. 
Unfortunately, this ideal lighting scenario is not consistent with the way consumers take portraits, prompting the definition of more robust metrics.


\section{Contrast Definition} 

We define face contrast as a multi-scale perceptual property that cannot be fully captured by a single global statistic. We thus decompose it into three complementary components: overall contrast, eye contrast, and micro-contrast. This decomposition was developed through consultations with professional photographers to provide meaningful compositions that can vary independently. It also provides a practical and interpretable framework for face contrast analysis.  
Figure~\ref{fig:face_rendering} illustrates the diversity of facial contrast. Although images~(a) and~(b) were captured under the same lighting condition, they differ markedly because of device-specific face rendering: (a) shows high contrast from cast shadows, whereas (b) exhibits overly harsh skin texture. By contrast, image~(c) shows high eye contrast with smooth skin. This diversity reinforces that contrast is an inherently complex attribute and further motivates the decomposition into three components.

\begin{itemize}
\item \textbf{Overall contrast} describes the global tonal separation across the face, reflecting how clearly bright and dark are separated~\cite{michelson1995studies,chen2023face}.
\item \textbf{Eye contrast} captures the local luminance separation within the eye region, which is semantically and visually critical in portrait perception~\cite{peterson2012looking,royer2018greater}.

\item \textbf{Micro-contrast} quantifies the visibility of fine facial details, such as pores in the skin and small local structures~\cite{chahine2023image,cho2026adapting}.

\item \textbf{Balanced contrast} describes the perceived harmony among different aspects of facial contrast. It reflects whether global tonal separation, eye definition, and fine-detail visibility are rendered in a natural and visually pleasing balance.

\end{itemize}

Overall, these definitions capture properties of contrast but are not directly quantifiable. This prompts us to focus on human judgment as the reference for the level of contrast, which we then try to model with specific metrics.

\section{Dataset construction}
 
\begin{figure}[t]
    \centering
    \includegraphics[width=0.6\columnwidth]{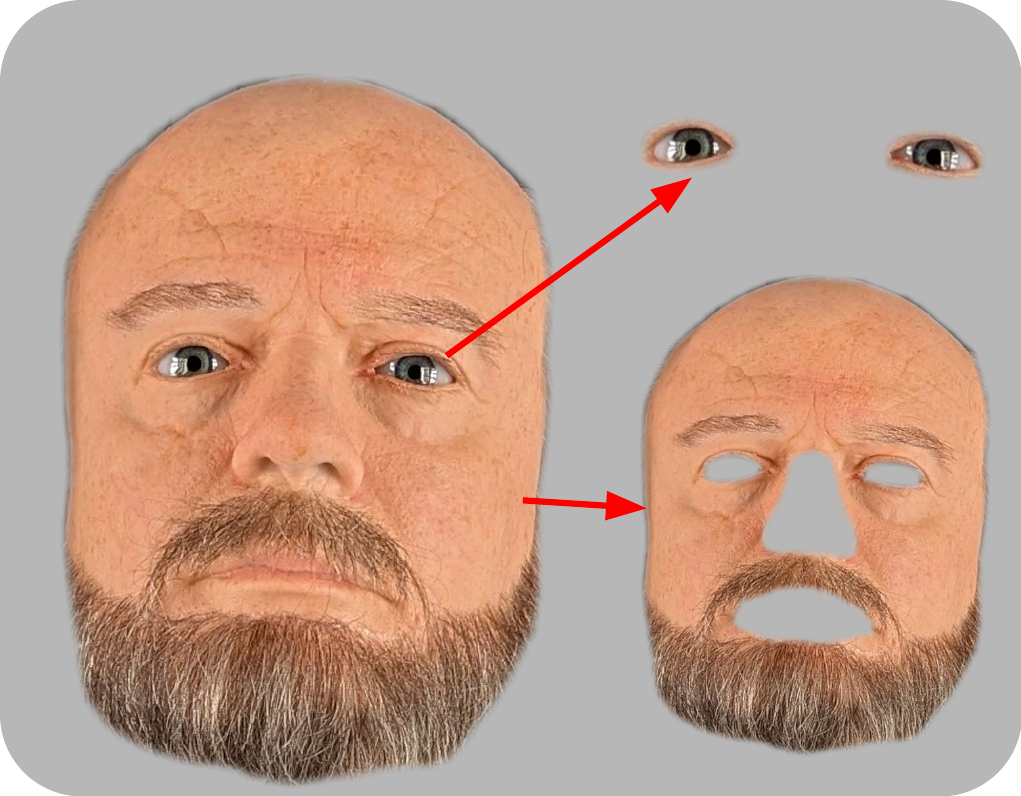}
    \caption{Illustration of the face mask ($\Omega_f$) , eye mask ($\Omega_e$), and skin mask  ($\Omega_s$) used throughout our work.}
    \label{fig:mask_examples}
    \vspace*{-1.5em}
\end{figure}

We construct a portrait dataset using industry-grade realistic mannequins in a laboratory setup with controlled illumination, scene geometry, and camera motion. The complete dataset contains 508 full-resolution images acquired with 29 smartphones from multiple manufacturers under four illumination conditions (Figure~\ref{fig:figure1}). 
For perceptual annotation,  we supplement the original captures with DxO PhotoLab9 edits that increase or decrease tonal or micro-contrast. This yields 800 images with 2,400 region-specific masks (Fig.~\ref{fig:mask_examples})~\cite{yu2018bisenet}. Region-specific masks allow each contrast metric to be computed in its corresponding perceptually relevant region, while reducing interference from other components. 
The dataset uses two high-fidelity mannequins with different skin tones, Eugene and Diana, which reliably trigger the portrait rendering pipelines of the tested smartphones. Additional details on the mannequins, including materials and reflectance measurements, are provided in the supplementary material (Sec.~A.3).
The images are organized into five groups based on lighting conditions, as shown in Table~\ref{tab:lighting_conditions}. Groups~1--4 correspond to four distinct lighting conditions, while Group~5 combines non-overlapping samples from these conditions. Each of Groups~1--4 contains 75 images per mannequin, while Group~5 contains 100 images per mannequin.


\begin{table}[t]
\centering
\caption{Lighting conditions used to construct the five comparison groups. $\Delta \mathrm{EV}$ denotes the calibrated exposure difference between the illumination on the mannequin and the light produced by the LED panel.}
\label{tab:lighting_conditions}
\small
\resizebox{0.9\columnwidth}{!}{%
\begin{tabular}{lll}
\hline
\textbf{Group} & \textbf{Light Uniformity} & \textbf{$\Delta \mathrm{EV}$ / Illuminance (lux)} \\ \hline
Group1 & Flat & $\Delta \mathrm{EV}$ = 0, 5--100 lux \\
Group2 & Flat & $\Delta \mathrm{EV}$ = 0, 1000--10000 lux \\
Group3 & Slightly non-uniform & $\Delta \mathrm{EV}$ = 4, 100--1000 lux \\
Group4 & Strongly non-uniform & $\Delta \mathrm{EV}$ = 7, 9; 5--1000 lux \\
Group5 & Mixed & Groups~1--4 \\ \hline
\end{tabular}%
}
\vspace*{-1em}
\end{table}

\vspace{-1mm}
\subsection{Annotation Protocol and JOD Scaling}

To align the proposed metrics with perception, we collect pairwise human judgments for each contrast component separately. Each pair is formed from images of the same mannequin identity, keeping facial structure and surface texture fixed across the two images so that observers can focus on differences in photographic rendering rather than differences between subjects. This setup also supports device-wise analysis of contrast rendering, reported in the supplementary material (Sec.~F.4). Pairwise judgments allow for more precise annotations than ambiguous absolute scoring~\cite{zhang2021uncertainty}. 
During each annotation, explicit guidelines are provided so that contrast is assessed separately from preference, sharpness, or general image quality. (see the supplementary material Sec.~C).
The annotation questions are phrased to match each target contrast attribute: 
\begin{enumerate}
    \item \textbf{Overall contrast:} ``Which image has the larger range from darks to highlights?'' 
    \item \textbf{Eye contrast:} ``Choose the image with the darker pupil and brighter sclera (white of the eye).''
    \item \textbf{Micro-contrast:} ``Choose the image where the tiny details on the skin surface are more visible.''
    \item \textbf{Balanced contrast:} ``Which image has more balanced face contrast?'' 
    
\end{enumerate}
These task-specific prompts help to disentangle the three contrast components and encourage consistent perceptual judgments among observers. Balanced contrast is collected as a separate annotation task and is used only for the downstream preference analysis, not to define the three contrast components. Across all four tasks, trained image-quality experts completed 20,000 pairwise comparisons over 160 hours in total. 
To improve annotation efficiency, image pairs are sampled using an active pairwise comparison strategy~\cite{jamieson2011active}, so that the collected comparisons focus on informative pairs rather than exhaustively enumerating all possible combinations.  For each contrast task, pairwise choices are converted into interval-scale values in Just Objectionable Difference (JOD) units using a Thurstone Case V paired-comparison model. A difference of 1 JOD between two images means that 75\% of observers would vote for the image with the higher JOD value~\cite{perez2019pairwise}. The JOD scale is estimated independently for each mannequin and comparison set. For Groups~1--4, a separate scale is estimated for each lighting condition, whereas Group~5 uses a single mixed-lighting scale per mannequin.




\section{Proposed Face Contrast Measures}
\label{sec:proposedmethods}
We propose metrics for the three face-contrast components defined above: overall contrast, eye contrast, and micro-contrast. These interpretable, perceptually grounded measures are evaluated against human annotations in the following section.
Following prior work on facial contrast~\cite{russell2009sex, porcheron2013aspects, russell2016facial}, we analyze facial lightness in the CIELAB color space. Specifically, we use the $L^*$ channel and normalize it to $[0,1]$, which provides an approximately perceptual lightness representation while separating tonal variation from chromatic variation. We denote by $L$ the normalized lightness image, and $\Omega_f$, $\Omega_e$, and $\Omega_s$ denote the face, eye, and skin masks, respectively. 

\subsection{Overall Contrast \texorpdfstring{($C_{\mathrm{overall}}$)}{(C\_overall)}}

Overall contrast captures large-scale luminance variation throughout the face, including shading, illumination gradients, and tonal transitions between bright and dark regions. In portraits, this component contributes strongly to the perceived depth and dimensionality of the face. 
Traditional global contrast measures such as Michelson contrast~\cite{michelson1995studies} do not fully capture this effect as they reduce the image to a single global luminance statistic and ignore the spatial distribution of contrast. We therefore measure overall contrast using a multi-scale formulation based on locally normalized luminance variation.
For a pixel $x \in \Omega_f$ in the face mask, we define the local contrast response at scale $r$ as
\begin{equation}
c_r(x) = \frac{\sigma_r(x)}{\mu_r(x) + \epsilon},
\end{equation}

where $\mu_r(x)$ and $\sigma_r(x)$ denote the local mean and standard deviation of the lightness image $L$, respectively, computed over an $r\times r$ square patch centered at $x$, and $\epsilon=10^{-6}$ is used for numerical stability.
This formulation is related to classical local contrast normalization in complex images~\cite{peli1990contrast}. 
By normalizing the local standard deviation by the local mean, it reduces sensitivity to absolute luminance and emphasizes relative tonal variation across the face.
As a result, the metric is less dominated by raw brightness level and better reflects how tonal contrast is rendered locally across the face. To capture contrast across multiple spatial scales, we pool the local response over the face mask and average it across multiple scales $\mathcal{R}$, where $|\Omega_f|$ denotes the number of pixels in the face mask:
\begin{equation}
C_{\mathrm{overall}} =
\frac{1} {|\mathcal{R}|}\sum_{r\in\mathcal{R}}
\left(
\frac{1}{|\Omega_f|}\sum_{x \in \Omega_f} c_{r}(x)
\right).
\end{equation}
This multi-scale design is consistent with prior work~\cite{haun2013perceived} showing that contrast and illumination effects are perceived across multiple spatial scales. 
Smaller windows respond to local tonal variation and fine shading transitions, while larger windows capture broader tonal structure across regions such as the cheeks, forehead, and jaw. In our experiments, we set $\mathcal{R} = {16, 32, 64, 128, 256}$, with $|\mathcal{R}|=5$, to capture both local and global facial contrast structures~\cite{jobson1997multiscale}. (see Sec~D.2 in supplementary material.)


\subsection{Eye Contrast \texorpdfstring{($C_{\mathrm{eye}}$)}{(C\_eye)}}

The eyes are structurally distinct from the rest of the face because they contain both very dark regions (pupil) and very bright ones (sclera). Their contrast is therefore not always well captured by global facial contrast alone. Previous work on face perception~\cite{peterson2012looking,royer2018greater} also supports treating the eye region as a separate component of face contrast due to its singular role in face recognition. In addition, directional lighting can also strongly affect the separation between the pupil, iris, and sclera, making the eyes appear flatter or grayer. We thus treat eye contrast as a separate component and measure it from
the dominant dark--bright separation within the eye region. We divide the luminance values within the eye mask $\Omega_e$ into two classes using the Otsu thresholding method~\cite{otsu1975threshold}. The left-and right-eye masks are merged before applying a single Otsu threshold.
Let $D$ and $B$ denote the resulting dark and bright pixel sets, respectively. We define eye contrast as
\begin{equation}
C_{\mathrm{eye}} = \mu_B - \mu_D,
\end{equation}
where $\mu_B$ and $\mu_D$ are the mean lightness values of the bright and dark classes.

\subsection{Micro-Contrast \texorpdfstring{($C_{\mathrm{micro}}$)}{(C\_micro)}}

Micro-contrast is intended to capture the visibility of fine-scale facial texture in skin regions while remaining relatively insensitive to broader shading and illumination. In photography, micro-contrast is a commonly used but imprecise term, sometimes used interchangeably with texture visibility~\cite{wieczorek2018sharpness,jirsa2022contrast}. 
Here, we use it more specifically to describe the extent to which fine facial details are rendered in the final portrait image,
distinguishing it from a general preference for stronger sharpening or texture enhancement. As micro-contrast depends on both surface texture and the camera pipeline, our controlled mannequin setup fixes the underlying facial texture within each identity. This allows us to isolate how different devices render the same fine-scale details.
We now describe each stage of the proposed $C_{\mathrm{micro}}$ computation.\\

\noindent\textbf{Log-DoG Decomposition.} To reduce the effect of illumination and isolate texture-scale variation, we first operate in the logarithmic domain.  Let $L$ denote the lightness image, and define the logarithmic transformed image as
$L_{\log} = \log(L + \epsilon)$,
where $\epsilon$ is a small constant for numerical stability.
This transform converts multiplicative illumination changes into additive ones, making the representation more stable to brightness differences and allowing the metric to emphasize relative texture-scale luminance variations while reducing its sensitivity to global illumination gradients~\cite{prins2016psychophysics}. We then apply a Difference of Gaussians (DoG) operator to extract a band of spatial frequencies associated with micro texture, $D_{\log}=G_{\sigma_{\mathrm{fine}}}\ast L_{\log}
-G_{\sigma_{\mathrm{coarse}}}\ast L_{\log}$,
where $G_{\sigma}$ denotes Gaussian smoothing with standard deviation $\sigma$. Inspired by Gaussian scale-space sampling~\cite{morel2011sift}, we use $\sigma_{\mathrm{fine}} = \frac{1}{\sqrt{2}}$ and $\sigma_{\mathrm{coarse}} = \frac{3}{\sqrt{2}}$, which provide a band-pass response that suppresses broad shading while preserving fine facial detail~\cite{petro2014multiscale}. The ratio $\sigma_{\mathrm{coarse}}/\sigma_{\mathrm{fine}} = 3$ follows a previously established center--surround DoG configuration~\cite{hansen2008recurrent}. The use of DoG is motivated by early vision models of center--surround filtering and edge detection~\cite{marr1980theory}, and can be interpreted as an approximation of Peli-style local contrast normalization~\cite{peli1990contrast,haun2013perceived}.\\



\noindent\textbf{Local Energy Pooling.} The DoG response $D_{\log}$ is a signed signal, and its local mean may be close to zero even when strong texture is present. To obtain a positive measure of local textural strength, we compute the local RMS energy over a finite neighborhood. For each pixel $x \in \Omega_s$, let $N(x)$ denote an $11 \times 11$ window centered at $x$. The local energy is defined as
\begin{equation}
E_{\mathrm{local}}(x)
=
\sqrt{
\frac{1}{|N(x)|}
\sum_{y \in N(x)} D_{\log}(y)^2
}.
\end{equation}

 This local energy representation is consistent with micro-contrast as a measure of fine-grained local variation, as well as with the classical Filter--Rectify--Filter (FRF) framework for texture perception and early visual processing. The latter has been used to explain how the visual system estimates texture energy and texture boundaries~\cite{landy1991texture,landy200473,adelson1985spatiotemporal}.\\

\noindent\textbf{Perceptual Response Compression.} Following Stevens' power law~\cite{stevens1957psychophysical}, perceived magnitude is often modeled as a power function of stimulus intensity. In the specific case of contrast perception, Legge and Foley~\cite{legge1980contrast} showed that the visual system applies a non-linear transducer that progressively compresses higher contrast levels. In other words, the physical strength of the local texture does not translate linearly into the perceived texture contrast. To model perceptual saturation, we apply a power-law nonlinearity to the local energy:
\begin{equation}
S_{\mathrm{local}}(x)
=
\left(E_{\mathrm{local}}(x)\right)^{0.6}.
\end{equation}

The exponent $0.6$ is chosen to reflect supra-threshold compression of perceived contrast, reducing the dominance of strong local responses. This nonlinearity is applied to local energy rather than directly to luminance values, consistent with response-compression models operating on the outputs of spatially filtered visual channels~\cite{heeger1992normalization}.\\

\noindent\textbf{Robust Spatial Pooling.} The final micro-contrast measure is obtained by robustly pooling the perceptual local responses over the skin mask $\Omega_s$ to reduce the influence of specular highlights, sensor noise, and mask boundary artifacts.  Let $q_{10}$ and $q_{90}$ denote the 10th and 90th percentiles of the set $\{S_{\mathrm{local}}(x)\mid x \in \Omega_s\}$.
We define
\begin{equation}
\begin{aligned}
C_{\mathrm{micro}}
&=
\frac{1}{|\mathcal{K}|}
\sum_{x \in \mathcal{K}} S_{\mathrm{local}}(x), \\
\mathcal{K}
&=
\left\{
x \in \Omega_s \;\middle|\;
q_{10} \le S_{\mathrm{local}}(x) \le q_{90}
\right\}.
\end{aligned}
\end{equation}

This formulation improves robustness to outlier responses while preserving a perceptually meaningful measure of micro-contrast in skin regions. Parameter sensitivity analysis in Sec.~D.4 and Fig.~S13 show that the metric remain stable under reasonable variations.


\section{Experiments and Results}

We evaluate the proposed methods under two complementary settings. (i) \emph{Within-lighting evaluation} is conducted separately for Groups~1--4 using a device-disjoint train/validation split comprising 19 training devices and 10 validation devices.  It measures generalization to unseen devices within a fixed lighting condition. The complete device allocation is reported in the supplementary material (Sec.~F.2).
(ii) \emph{Mixed-lighting evaluation} is conducted on Group~5, which contains images from multiple lighting conditions that do not overlap with Groups~1--4 and are assessed on a common JOD scale. For fitted balanced contrast preference predictors, Groups~1--4 are used for parameter estimation and Group~5 is used exclusively for evaluation.  
We evaluate the Overall, Eye, and Micro contrast measures by computing Pearson's linear correlation coefficient (PLCC) and
Spearman's rank correlation coefficient (SRCC) against their corresponding component-specific JOD scores, as reported in Table~\ref{tab:all_contrast_protocols}. For balanced-contrast preference, the three contrast components are instead jointly fitted to predict balanced-contrast JOD, with results reported in Table~\ref{tab:balanced_contrast_before_after}.

\subsection{Amount of Contrast Analysis}


\noindent\textbf{Overall Contrast.} We compare the proposed metric with classical luminance statistics and prior facial contrast measures~\cite{russell2009sex,porcheron2013aspects,russell2016facial}. It achieves the highest nominal correlations, with PLCC/SRCC of 0.792/0.780 for \emph{within-lighting} and 0.688/0.693 for \emph{mixed-lighting evaluation}. Global luminance measures are competitive within lighting but degrade under mixed lighting. Feature-to-skin measures show weaker correlations, as they quantify luminance or chromatic differences between facial features and the surrounding skin rather than face-wide shading and tonal variation. These results suggest that overall facial contrast depends on the spatial distribution of relative tonal variation across the face. \\

\begin{table}[t]
\centering
\caption{Comparison of proposed and reference measures for overall, eye, and micro-contrast under within-lighting device-disjoint
evaluation on Groups~1--4 and mixed-lighting evaluation on Group~5. Results are based on 500 bootstrap resamples.}

\label{tab:all_contrast_protocols}
\small
\setlength{\tabcolsep}{5pt}

\resizebox{1\columnwidth}{!}{
\begin{tabular}{llcccc}
\hline
\multirow{2}{*}{Contrast} & \multirow{2}{*}{Metric} & 
\multicolumn{2}{c}{Within-lighting} & \multicolumn{2}{c}{Mixed-lighting} \\
& & PLCC & SRCC & PLCC & SRCC \\
\hline
\multirow{7}{*}{Overall}
& \textbf{Proposed overall contrast} & \textbf{0.792} & \textbf{0.780} & \textbf{0.688} & \textbf{0.693} \\

& Global log-luminance std. & 0.723 & 0.708 & \underline{0.604} & \underline{0.566} \\
& Global luminance std. & 0.675 & 0.646 & 0.551 & 0.562\\ 
& Michelson contrast (p2--p98)~\cite{michelson1995studies} & \underline{0.731} & \underline{0.710} & 0.583 & 0.542 \\
& Facial feature contrast~\cite{russell2009sex} & -0.026 & -0.005 & -0.302 & -0.263 \\
& Normalized facial feature contrast~\cite{porcheron2013aspects} & -0.019 & -0.016 & -0.254 & -0.259 \\
& Facial contrast vector magnitude~\cite{russell2016facial} & 0.137 & 0.226 & -0.281 & -0.103 \\
\hline
\multirow{4}{*}{Eye}
& \textbf{Otsu-based eye contrast~\cite{otsu1975threshold}} & \textbf{0.879} & \textbf{0.855} & \underline{0.892} & 0.891 \\
& Dark--bright separation ($h_i(1-l_o)$) & 0.852 & 0.832 & \textbf{0.901} & \textbf{0.901} \\
& Percentile range ($h_i-l_o$) & 0.847 & \textbf{0.855} & 0.882 & \underline{0.892} \\
& Michelson normalized range~\cite{michelson1995studies} & 0.680 & 0.704 & 0.703 & 0.677 \\
\hline
\multirow{8}{*}{Micro}
& \textbf{Proposed micro-contrast} & \textbf{0.849} & \textbf{0.842} & \textbf{0.797} & \textbf{0.787} \\
& GLCM contrast~\cite{haralick2007textural} & 0.682 & 0.753 & 0.594 & 0.706 \\
& GLCM dissimilarity~\cite{haralick2007textural} & \underline{0.812} & \underline{0.829} & \underline{0.763} & \underline{0.783} \\
& LBP entropy~\cite{ojala2002multiresolution} & 0.257 & 0.232 & 0.371 & 0.305 \\
& Wavelet energy (L1)~\cite{mallat2002theory} & 0.804 & 0.819 & 0.759 & 0.775 \\
& CPBD~\cite{narvekar2011no} & 0.336 & 0.305 & 0.358 & 0.283 \\
& S3 sharpness~\cite{vu2011bf} & 0.490 & 0.478 & 0.604 & 0.564 \\
& Michelson contrast (p2--p98)~\cite{michelson1995studies} & 0.223 & 0.241 & 0.282 & 0.256 \\
\hline
\end{tabular}
}
\vspace{-1em}
\end{table}


\noindent\textbf{Eye Contrast.} We compare the Otsu-based formulation with three reference measures derived from the luminance distribution within the eye mask. Let $h_i$ and $l_o$ denote the 98th and 2nd percentiles of eye-region lightness, respectively. The Otsu-based formulation achieves PLCC/SRCC values of 0.879/0.855 for \emph{within-lighting} and 0.892/0.891 for \emph{mixed-lighting evaluation}. The percentile-based measures perform comparably; dark--bright separation is slightly higher under \emph{mixed-lighting}, while Otsu performs slightly better overall \emph{within-lighting}. 
On average however, Otsu achieves higher performance without needing any parameter tuning to remove outliers. Overall, the results show that dark--bright luminance separation is a strong indicator of perceptual eye contrast. \\

\noindent\textbf{Micro-Contrast.} We compare the proposed formulation with conventional texture
descriptors, including GLCM~\cite{haralick2007textural}, LBP
entropy~\cite{ojala2002multiresolution}, and wavelet energy~\cite{mallat2002theory}, as well as the perceptual sharpness
measures CPBD~\cite{narvekar2011no} and S3~\cite{vu2011bf}.
Although these methods were not specifically designed for facial
micro-contrast, they provide relevant baselines for assessing alignment with human judgments of fine facial texture. The proposed metric achieves the highest nominal correlations in both settings, reaching PLCC/SRCC values of 0.849/0.842 for \emph{within-lighting} and 0.797/0.787 for \emph{mixed-lighting} evaluation. GLCM dissimilarity and Wavelet energy also provides good performance, suggesting that local gray-level co-occurrence pattern and spatial-frequency energy capture perceptually relevant skin micro-contrast. However, CPBD and S3 show lower correlations, indicating that perceived micro-contrast is not fully characterized by image sharpness alone.
LBP entropy and Michelson contrast~\cite{michelson1995studies} perform substantially worse, indicating that structural diversity or luminance range are insufficient to capture perceived micro-contrast.


\subsection{Balanced Contrast Preference Analysis}

\begin{table*}[t]
\captionsetup[table]{justification=raggedright,singlelinecheck=false}
\caption{Balanced contrast preference prediction before and after task adaptation. We compare the proposed component-based method with hand-crafted, general IQA, portrait IQA, and face-quality baselines.
``3 (raw)'' denotes the fitted coefficients of $ax^2+bx+c$
for a scalar score. Our combined model uses a second-order polynomial over three contrast components with 9 trainable parameters. Entries are omitted when a before- or after-adaptation counterpart is not applicable. All results are computed from 500 bootstrap resamples.}
\label{tab:balanced_contrast_before_after}

\fontsize{6.5}{7.2}\selectfont
\setlength{\tabcolsep}{2.3pt}
\renewcommand{\arraystretch}{0.87}

\centering
\resizebox{0.8\textwidth}{!}{%
\begin{tabular}{lllcc cccc cccc}
\toprule
\multirow{3}{*}{Type}
& \multirow{3}{*}{Method}
& \multirow{3}{*}{Expl.}
& \multirow{3}{*}{Total}
& \multirow{3}{*}{Train.}
& \multicolumn{4}{c}{Before Adaptation}
& \multicolumn{4}{c}{After Adaptation} \\
\cmidrule(lr){6-9} \cmidrule(lr){10-13}

& & & & &
\multicolumn{2}{c}{Within-lighting}
& \multicolumn{2}{c}{Mixed-lighting}
& \multicolumn{2}{c}{Within-lighting}
& \multicolumn{2}{c}{Mixed lighting} \\
\cmidrule(lr){6-7}
\cmidrule(lr){8-9}
\cmidrule(lr){10-11}
\cmidrule(lr){12-13}

& & & & &
PLCC & SRCC & PLCC & SRCC &
PLCC & SRCC & PLCC & SRCC \\
\midrule


\multirow{6}{*}{\shortstack[l]{Hand-crafted}}
& BRISQUE~\cite{mittal2012no}
& \checkmark
& 3 (raw) & 3 (raw)
& 0.115 & 0.063 & 0.223 & 0.190
& 0.169 & 0.134 & 0.235 & 0.316 \\

& NIQE~\cite{mittal2012making}
& \checkmark
& 3 (raw) & 3 (raw)
& 0.411 & 0.430 & 0.533 & 0.603
& 0.397 & 0.405 & 0.461 & 0.577 \\

& Ours (Overall)
& \checkmark
& 3 (raw) & 3 (raw)
& 0.567 & 0.602 & 0.579 & 0.586
& 0.689 & 0.637 & 0.647 & 0.589 \\


& Ours (Eye)
& \checkmark
& 3 (raw) & 3 (raw)
& 0.456 & 0.447 & 0.623 & 0.596
& 0.647 & 0.570 & 0.645 & 0.634 \\

& Ours (Micro)
& \checkmark
& 3 (raw) & 3 (raw)
& 0.465 & 0.523 & 0.574 & 0.586
& 0.546 & 0.542 & 0.669 & 0.628 \\

& \textbf{Ours (combined)}
& \checkmark
& 9 \text{ (raw)}
& 9 \text{ (raw)}
& -- & -- & -- & --
& \underline{0.728} & \underline{0.739}
& \underline{0.773} & \underline{0.778} \\


\midrule


\multirow{6}{*}{\shortstack[l]{General IQA}}
& ARNIQA~\cite{agnolucci2024arniqa}
& \texttimes
& 23.56M & 4K
& 0.252 & 0.251 & 0.474 & 0.509
& 0.355 & 0.319 & 0.308 & 0.296 \\

& CLIPIQA~\cite{wang2022exploring}
& \texttimes
& 102.06M & 1K
& 0.474 & 0.495 & 0.540 & 0.581
& 0.635 & 0.612 & 0.759 & 0.739 \\

& HyperIQA~\cite{Su_2020_CVPR}
& \texttimes
& 27.43M & 2K
& 0.285 & 0.268 & 0.462 & 0.478
& 0.338 & 0.320 & 0.377 & 0.385 \\

& MANIQA~\cite{yang2022maniqa}
& \texttimes
& 135.8M & 296K
& 0.412 & 0.404 & 0.608 & 0.619
& 0.547 & 0.598 & 0.709 & 0.772 \\

& MUSIQ~\cite{ke2021musiq}
& \texttimes
& 27.13M & 4K
& 0.382 & 0.402 & 0.597 & 0.641
& 0.353 & 0.315 & 0.523 & 0.488 \\

& TOPIQ~\cite{chen2024topiq}
& \texttimes
& 45.20M & 133K
& 0.373 & 0.355 & 0.587 & 0.592
& 0.661 & 0.642 & 0.753 & 0.751 \\

\midrule
\multirow{3}{*}{\shortstack[l]{MLLM IQA}}
& Q-Align~\cite{wu2023q}
& \texttimes
& 8.2B & 0
& 0.507 & 0.511 & 0.615 & 0.632
& -- & -- & -- & --\\

& DeQA-Score~\cite{you2025teaching}
& \texttimes
& 8.2B & 0
& 0.609 & 0.572 & 0.744 & 0.753
& -- & -- & -- & -- \\

& Q-Insight~\cite{li2026q}
& \texttimes
& 8.29B & 0
& 0.571 & 0.590 & 0.639 & 0.698
& -- & -- & -- & -- \\

\midrule
\multirow{1}{*}{\shortstack[l]{Portrait IQA}}
& Full-HyperIQA~\cite{chahine2025generalized}
& \texttimes
& 140.22M & 24K
& 0.364 & 0.367 & 0.584 & 0.568
& 0.395 & 0.394 & 0.571 & 0.555 \\

\midrule


\multirow{5}{*}{\shortstack[l]{Face Quality}}
& CR\_FIQA\_L~\cite{boutros2023cr}
& \texttimes
& 65.29M & 66K
& 0.150 & 0.146 & 0.319 & 0.220
& 0.533 & 0.545 & 0.708 & 0.720 \\

& DifFIQA\_R~\cite{babnik2023diffiqa}
& \texttimes
& 65.29M & 66K
& 0.394 & 0.390 & 0.565 & 0.499
& 0.554 & 0.538 & 0.763 & 0.781 \\

& GraFIQs\_R50~\cite{kolf2024grafiqs}
& \texttimes
& 43.7M & 66K
& -0.141 & -0.132 & -0.297 & -0.199
& 0.539 & 0.500 & 0.658 & 0.654 \\

& eDifFIQA\_S~\cite{babnik2024ediffiqa}
& \texttimes
& 24.11M & 66K
& 0.496 & 0.490 & 0.588 & 0.505
& 0.609 & 0.598 & 0.791 & 0.805 \\

& \textbf{IFQA (VIT)}~\cite{jo2023ifqa}
& \texttimes
& 657.04M & 164K
& 0.330 & 0.283 & 0.410 & 0.461
& \textbf{0.858} & \textbf{0.856}
& \textbf{0.852} & \textbf{0.868} \\

\bottomrule
\end{tabular}%
}
\vspace*{-1em}
\end{table*}

As a downstream analysis, we examine whether the three proposed contrast components provide a useful basis for modeling balanced contrast preference. In this task, annotators selected the image with the most \emph{balanced contrast}, reflecting an overall judgment rather than any single component in isolation.
Our best variant is a second-order polynomial regressor over the three proposed contrast components (see supplementary Sec~D.6).
As shown in Table~\ref{tab:balanced_contrast_before_after}, this model reaches PLCC/SRCC values of 0.728/0.739 under \emph{within-lighting} and 0.773/0.778 under \emph{mixed-lighting}. 
Each component alone yields non-trivial but suboptimal performance, indicating that balanced contrast preference is not explained by a single attribute. We further compare against five related families: hand-crafted IQA~\cite{mittal2012making,mittal2012no}, general IQA~\cite{agnolucci2024arniqa,wang2022exploring,Su_2020_CVPR,yang2022maniqa,ke2021musiq,chen2024topiq}, MLLM-based IQA~\cite{wu2023q,you2025teaching,li2026q}, portrait IQA~\cite{chahine2025generalized}, and face-quality assessment~\cite{boutros2023cr,babnik2023diffiqa,kolf2024grafiqs,babnik2024ediffiqa,jo2023ifqa}. The IQA methods primarily target global perceptual image quality, whereas portrait IQA focuses on portrait quality. Face-quality models assess facial image quality, often in the context of biometric utility or recognition-related tasks.
We evaluate both off-the-shelf and after lightweight task adaptation to balanced contrast preference.  
Before adaptation, most learning-based methods perform markedly worse, indicating a gap between their original objectives and balanced contrast assessment. MLLM-based IQA models show relatively strong zero-shot performance. Given the limited training data, we evaluate these large models zero-shot to avoid unreliable adaptation.

For a fairer comparison, we fine-tune remaining learned baselines using lightweight trainable heads while keeping the pretrained representation largely frozen. After adaptation, performance improves substantially for several learned baselines, especially face-quality models. 
Among all methods, fine-tuned IFQA~\cite{jo2023ifqa} achieves the highest performance. IFQA was pretrained using human preference judgments of restored faces, which may make its representation particularly suitable for this task. Nevertheless, our proposed component-based model remains the strongest among learning-free methods and the second-best method overall. 
Combining our metrics surpasses most neural-network-based baselines, showing that the proposed decomposition captures a substantial part of balanced contrast preference in the controlled setting. The remaining gap to fine-tuned IFQA suggests that balanced contrast judgments may also depend on additional cues beyond the three explicit contrast components.
This latter aspect is where neural-network approaches shine: using large amounts of pretraining data, they are able to efficiently model human preference. While our method performs best in low-data regimes or when interpretability is desired, learning-based approaches can be used to capture the remaining subtlety in human contrast preferences. We expect the interaction between the two paradigms to yield improved models of contrast preference.

\begin{table}[t]
\centering
\caption{
Real-portrait generalization results on 444 images across 10 scenes. PLCC and SRCC are computed separately within each scene and then
macro-averaged across scenes. Ours is applied in a zero-shot manner, while IFQA is fine-tuned on the mannequin-based training annotations
for each corresponding task.
}
\label{tab:real_generalization}
\setlength{\tabcolsep}{5pt}
\resizebox{0.8\columnwidth}{!}{%
\begin{tabular}{l|cc|cc}
\hline
\multirow{2}{*}{\textbf{Target}} 
& \multicolumn{2}{c|}{\textbf{Ours}} 
& \multicolumn{2}{c}{\textbf{IFQA (fine-tuned)}} \\
\cline{2-5}
& \textbf{PLCC} & \textbf{SRCC} 
& \textbf{PLCC} & \textbf{SRCC} \\
\hline
Overall contrast & 0.792 & 0.825 & \textbf{0.858} & \textbf{0.862} \\
Eye contrast     & \textbf{0.753} & \textbf{0.768} & 0.633 & 0.629 \\
Micro-contrast   & \textbf{0.717} & \textbf{0.731} & 0.693 & 0.678 \\
\hline
\end{tabular}
}
\vspace*{-0.5em}
\end{table}

\subsection{Generalization to Real Portrait Data}

To examine whether the mannequin-based findings extend to real portraits, we collected a smaller, less controlled dataset of 444 images from 10 subjects/scenes. 
It contains 29 unique capture devices, 18 of which are not present in the mannequin dataset. We generate contrast-adjusted variants using DxO PhotoLab 9 to obtain at least 44 images per scene for reliable pairwise comparison. It includes more varied facial appearances and poses, including three-quarter views (Supplementary Fig.~S14). We collected pairwise annotations for overall contrast, eye contrast, and micro-contrast using the same questions as in the controlled study and converted them to JOD scores, requiring 80 annotation hours. We compare the proposed zero-shot metrics with IFQA fine-tuned on the mannequin annotations (Table~\ref{tab:real_generalization}). Our metrics outperform fine-tuned IFQA for eye and micro-contrast, while IFQA performs better for overall contrast. Without fitting on mannequin or real-portrait annotations, the proposed metrics retain meaningful correlations on real portraits. We additionally report scene-wise performance with uncertainty in Supplementary Table~S9. 
These results suggest that the mannequin dataset can bridge controlled contrast analysis and real-portrait evaluation.
Broader evaluation across identities, skin tones, poses, and capture conditions remains important future work.

\begin{figure}[t]
    \centering
    \includegraphics[width=0.8\columnwidth]{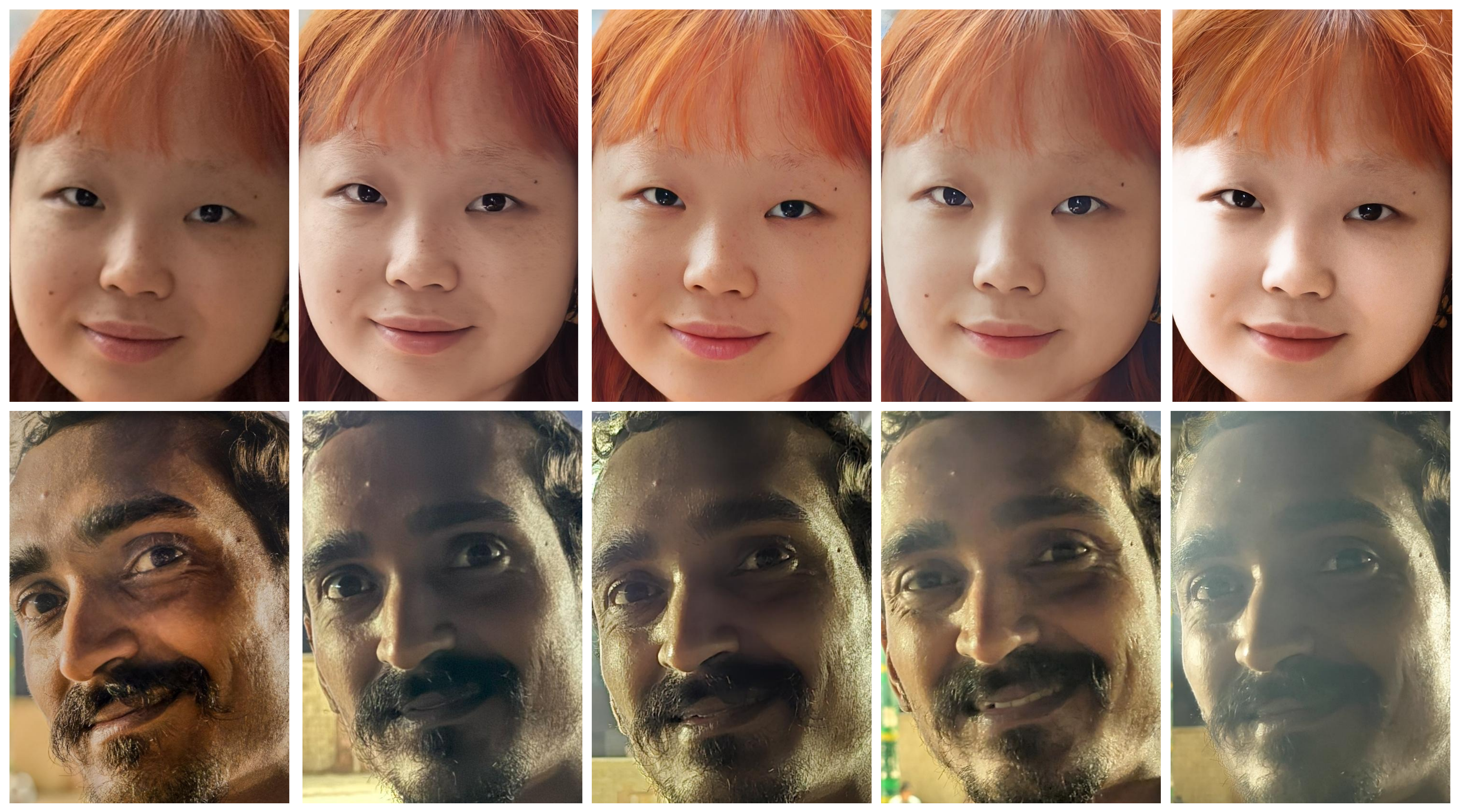}
    \caption{Examples of real portraits used in our preliminary real-world analysis. Images span multiple subjects, devices, and edited contrast variants.}
    \label{fig:overall_real}
    \vspace*{-1em}
\end{figure}

\section{Conclusion}
\label{sec:conclusion}

We constructed a realistic mannequin-based dataset captured using multiple smartphones under diverse lighting conditions. For controlled face contrast evaluation, we paired the dataset with region-specific masks and psychophysical annotations. Rather than treating face contrast as a single global quantity, we decomposed it into three perceptual components: overall contrast, eye contrast, and micro-contrast, and proposed an interpretable metric for each attribute. The key improvement over previous metrics lies in leveraging the multi-scale and multi-frequency aspects of contrast perception.
Our experimental results demonstrate that the three proposed metrics align closely with human contrast perception, and perform strongly against conventional contrast, texture and sharpness measures.
These results suggest that face contrast in portrait imaging is better understood as a structured, multi-scale perceptual property rather than as a simple global statistic. As a downstream analysis, we examined the preference for balanced facial contrast in the controlled mannequin setting.
The measured overall, eye, and micro-contrast components provide a meaningful basis for explaining holistic contrast preference: each component alone is insufficient, while their combination explains substantially more variation in human judgments.
A fine-tuned face-quality model achieved slightly higher correlations, suggesting that balanced contrast judgments may be influenced by additional cues beyond the three explicit contrast components. We also evaluated the proposed metrics on real-world portraits, showing that our zero-shot metrics generalize beyond the controlled mannequin setting. We hope that the proposed dataset and interpretable formulation will support future work on interpretable portrait quality modeling and face-contrast evaluation.

\section{Acknowledgements} This work was financed by a grant from ANRT (Association Nationale de la Recherche et de la Technologie) N$^\circ$ 2023/0335. This work used GENCI-IDRIS HPC and storage resources under allocations AD011017622, AD011014895R2, AD011014895R1, and AD011014895R  on Jean Zay. 
{
    \small
    \bibliographystyle{ieeenat_fullname}
    \bibliography{main}
}

\clearpage
\makeatletter
\global\@colht\textheight
\global\@colroom\textheight
\global\vsize\textheight
\makeatother

\appendix
\maketitlesupplementary

\setcounter{table}{0}
\renewcommand{\thetable}{S\arabic{table}}
\renewcommand{\theHtable}{supp.\arabic{table}}
\setcounter{figure}{0}
\renewcommand{\thefigure}{S\arabic{figure}}
\renewcommand{\theHfigure}{supp.\arabic{figure}}
\setcounter{equation}{0}
\renewcommand{\theequation}{\arabic{equation}}
\renewcommand{\theHequation}{supp.\arabic{equation}}


\section{Annotation Protocol}

\begin{figure*}[t]
    \centering
    \includegraphics[width=\textwidth]{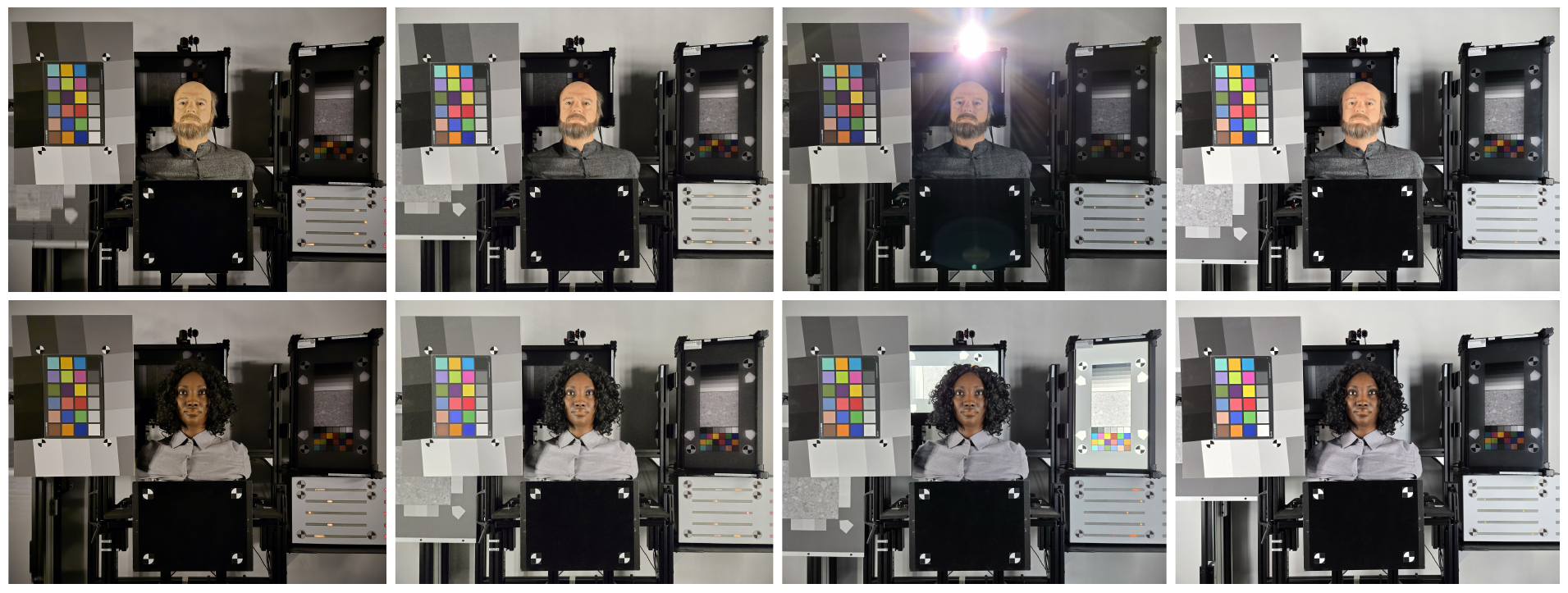}
 
    \vspace{0.2mm}
    \noindent
    \makebox[\textwidth][c]{%
        \parbox[t]{0.24\textwidth}{\centering \small (a) 5 lux}
        \hfill
        \parbox[t]{0.24\textwidth}{\centering \small (b) 100 lux}
        \hfill
        \parbox[t]{0.24\textwidth}{\centering \small (c) 1000 lux}
        \hfill
        \parbox[t]{0.24\textwidth}{\centering \small (d) 10000 lux}
    }  
    \caption{Examples from the full-resolution dataset. The two mannequins are Eugene and Diana, and the images were captured using a Samsung Galaxy Z Fold6. The dataset includes multiple $\Delta \mathrm{EV}$ conditions (0, 4, 7, and 9), but here we show only $\Delta \mathrm{EV}=0$ and $\Delta \mathrm{EV}=4$ under backlit conditions.}
    \label{fig:dataset_examples}
    \vspace*{-1em}
\end{figure*}

\subsection{Annotation Viewing Conditions}

All annotations were performed under controlled viewing conditions to limit unintended variation across observers. Images were presented on a calibrated 32-inch 4K BenQ monitor at a fixed viewing distance of 65\,cm. The monitor was configured to standard sRGB conditions with a D65 white point, while ambient illumination was regulated to minimize glare and visual fatigue, following~\cite{chahine2023image}. Face crops were displayed at a standardized scale according to the preprocessing procedure described in the main paper. This setup was intended to keep facial detail within a perceptually appropriate range for normal viewing and fine-level comparison.

\subsection{JOD Scaling}

For each contrast task, the collected pairwise choices were converted into interval-scale Just Objectionable Difference (JOD) values using a probabilistic paired-comparison scaling framework~\cite{perez2019pairwise}. Let $q_i$ denote the latent perceptual score associated with image $i$. For a pair of images $(i,j)$, the observer is assumed to choose image $i$ over image $j$ with probability determined by the difference between their latent scores. Following the Thurstone Case V observer model~\cite{davidson1976bibliography, thurstone1994law}, the internal perceptual response for each image is modeled as a noisy version of its latent score:
\begin{equation}
r_i = q_i + \epsilon_i,
\end{equation}
where $\epsilon_i$ is a random noise term. Under a Gaussian observer model, the probability that image $i$ is preferred over image $j$ is
\begin{equation}
P(i \succ j) = \Phi\!\left(\frac{q_i - q_j}{\sigma}\right),
\end{equation}
where $\Phi(\cdot)$ denotes the cumulative distribution function of the standard normal distribution and $\sigma$ controls the observer uncertainty. The latent scores $\{q_i\}$ are estimated from the measured pairwise responses by maximizing the likelihood of the observed comparisons. The resulting fitted values define an interval perceptual scale, which we report in JOD units. Under this scaling, a difference of 1~JOD corresponds approximately to 75\% agreement in pairwise comparisons.

Intuitively, this procedure maps binary pairwise decisions onto a continuous perceptual axis. Images with similar JOD values are difficult to distinguish perceptually for the target attribute, whereas larger JOD differences correspond to stronger perceived differences. These JOD values are then used as perceptually grounded references for evaluating the proposed contrast metrics. 
To quantify uncertainty in the recovered perceptual scale, we estimate confidence intervals using the analytical error-estimation framework of Zolliker and Bara\'nczuk~\cite{zolliker2010error}. Because our annotation protocol uses active pairwise sampling, these intervals should be interpreted as approximate scale-level uncertainty estimates rather than exact finite-sample guarantees. We use them as a practical indicator of the stability of the recovered JOD values under the incomplete comparison design.

\begin{figure}[t]
    \centering

    \includegraphics[width=0.95\columnwidth]{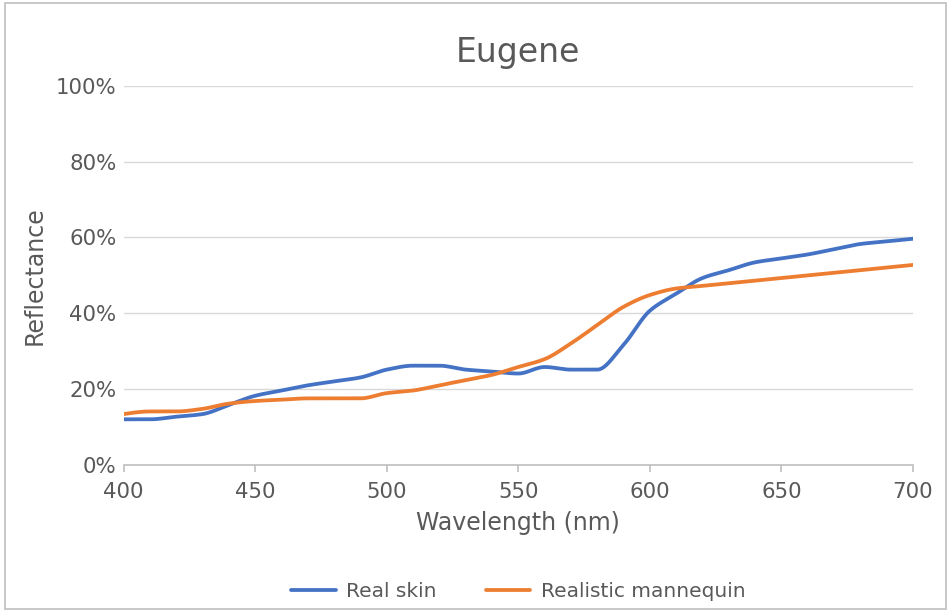}
    \centerline{\small (a)}

    \vspace{0.5em}

    \includegraphics[width=0.95\columnwidth]{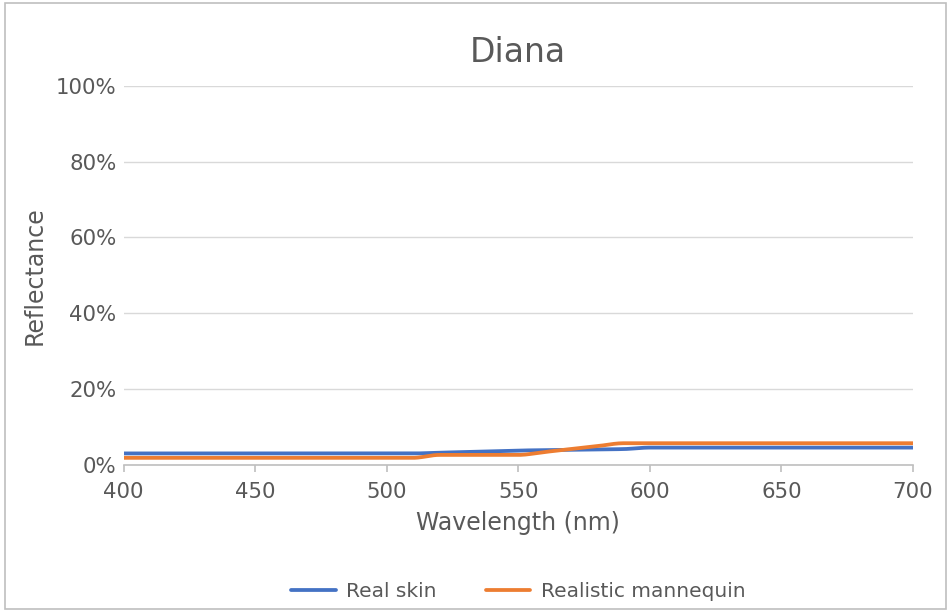}
    \centerline{\small (b)}

    \caption{Spectral reflectance comparison between real human skin and the corresponding realistic mannequin surface for (a) Eugene and (b) Diana. The mannequins follow similar wavelength-dependent reflectance trends to the real subjects.}
    \label{fig:reflectance}
    \vspace*{-1em}
\end{figure}

\subsection{Controlled Portrait Test Bench}

Figure~\ref{fig:dataset_examples} shows examples of images captured using a controlled portrait-imaging test bench. The setup is designed to ensure repeatable acquisition while reproducing realistic portrait scenes. It includes a realistic mannequin, a composite chart, a flare light source, a multispectral lighting system, a ColorChecker with grayscale patches, and an automated moving object. Together, these components allow us to vary facial appearance, illumination, flare, tonal and color references, and temporal scene conditions.
The use of a mannequin allows us to capture the same face repeatedly while avoiding variations in expression, pose, and subject motion. Its skin-like reflectance and fine surface details also make it suitable for evaluating how different smartphones render facial appearance under controlled conditions. The mannequins are realistic replicas of human subjects and are specifically designed for image-quality evaluation. They were tested to reliably trigger smartphone portrait modes. The mannequins are constructed from polyurethane foam with proprietary skin-like pigmentation, silicone components, real hair, and glass eyes. Figure~\ref{fig:reflectance} shows that the realistic mannequins closely follow the reflectance trends of the corresponding real human skin across the visible wavelength range.
The composite chart and the ColorChecker with grayscale patches provide reference structures for checking exposure and color consistency throughout the acquisition process. Illumination is controlled using a multispectral lighting system so that both the spectrum and intensity of the scene can be varied systematically. In this work, lux denotes the measured illuminance on the mannequin, while $\Delta \mathrm{EV}$ denotes the calibrated exposure difference between the light on the mannequin and the light produced by the LED panel. Thus, lux controls the physical amount of ambient light reaching the setup, whereas $\Delta \mathrm{EV}$ controls the relative strength of the LED backlight with respect to the mannequin illumination. By varying both factors, we obtain portrait images spanning flat and non-uniform lighting conditions, as well as a broad range of facial contrast appearances. A dedicated flare light source is used to introduce controlled stray-light and veiling-glare conditions, which can reduce tonal separation and affect perceived face contrast. Finally, the automated moving object introduces temporal variation into the scene, allowing the test bench to account for motion effects in multi-frame capture and ISP processing.

\section{Annotation Guidelines Overview}

The goal of this annotation task is to break down face contrast into a set of visually meaningful aspects that can be reliably evaluated across different smartphone images and lighting conditions. The appearance of facial contrast is affected by many factors, such as shading, skin reflectance, tone adjustment, noise, and fine details, so we collect annotations that separately describe each of these elements.

\begin{figure*}[t]
    \centering
    \begin{subfigure}{0.9\linewidth}
        \centering
        \includegraphics[width=\linewidth]{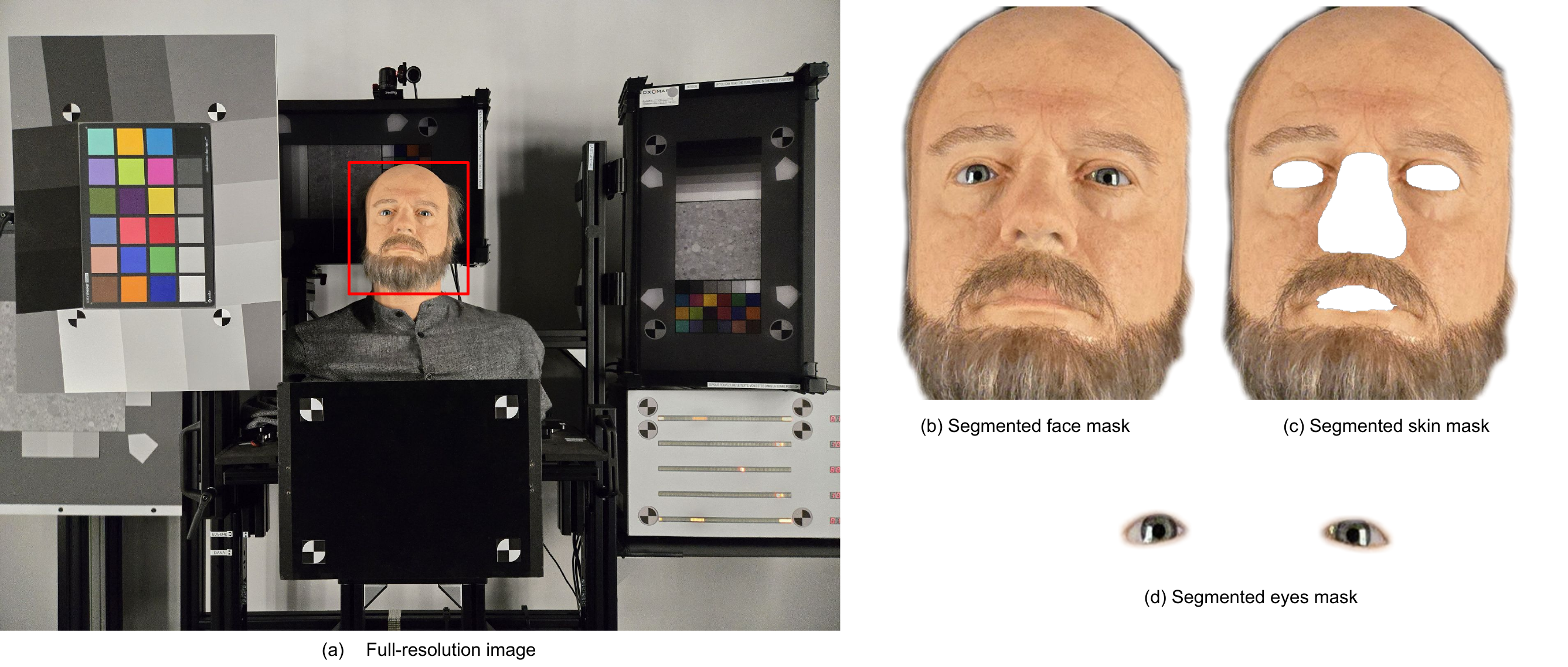}
        \caption{Bright skin tone mannequin (Eugene), taken under 100~lux -- LED4000K light.}
        \label{fig:guidelines_top}
    \end{subfigure}
    \vspace{0.5em}
    \begin{subfigure}{0.9\linewidth}
        \centering
        \includegraphics[width=\linewidth]{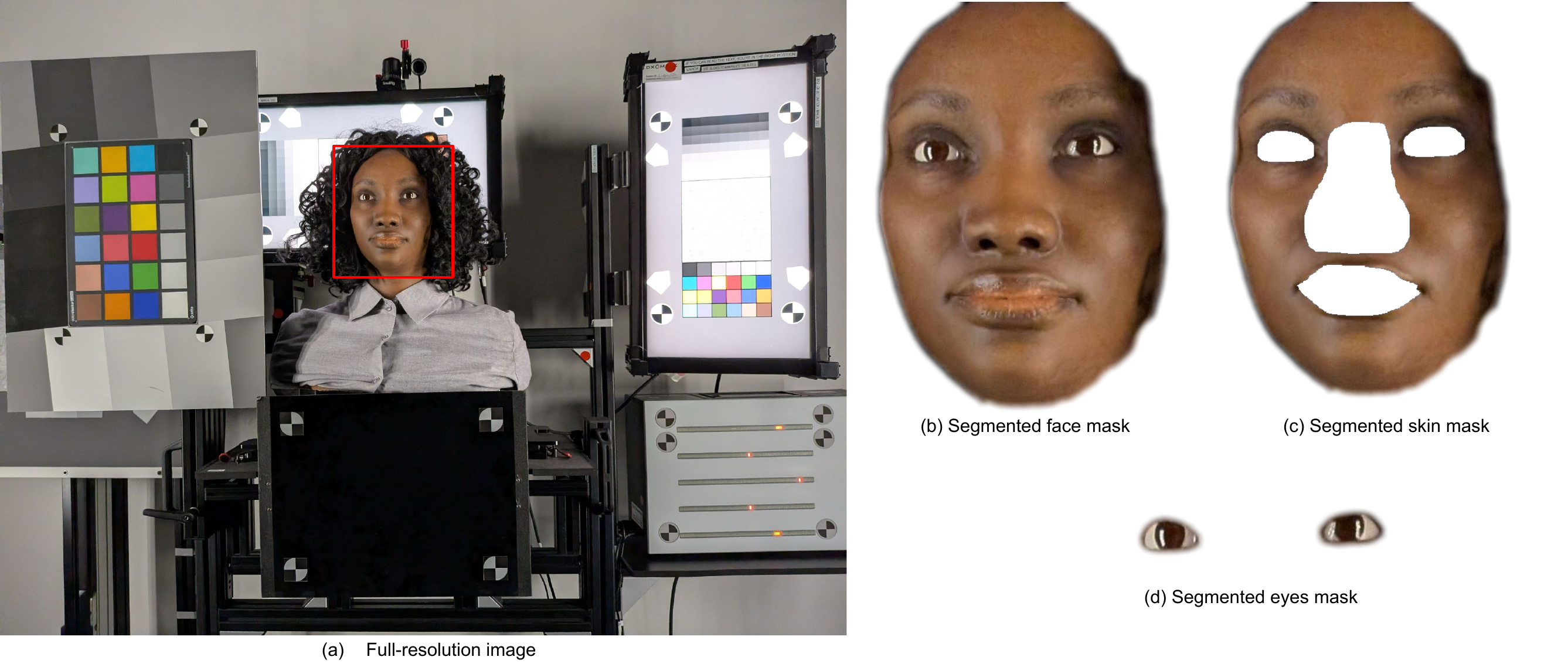}
        \caption{Dark skin tone mannequin (Diana), taken under 100~lux -- LED4000K light with $\Delta \mathrm{EV}=4$.}
        \label{fig:guidelines_bottom}
    \end{subfigure}
    \caption{Example scenes used to illustrate the annotation protocol for the two mannequin subjects. (a) Eugene under 100~lux, LED~4000K lighting. (b) Diana under 100~lux, LED~4000K lighting with $\Delta \mathrm{EV}=4$. The examples highlight the facial regions used to guide the overall-contrast, eye-contrast, and micro-contrast annotation tasks.}
    \label{fig:guidelines}
\end{figure*}

To make the evaluation simple and consistent, all images come from a fixed portrait scene using two realistic mannequins:
(i) \textbf{Eugene}, with a bright/white skin tone, and
(ii) \textbf{Diana}, with a dark skin tone.
Because the mannequins have different skin tones, we do not compare them to each other. All comparisons are made only within the same mannequin
(Eugene vs.\ Eugene, and Diana vs.\ Diana). This ensures that the evaluation focuses on differences created by the smartphone camera rather than differences in skin tone.

During the annotation, you will see masks that highlight specific parts of the face: (i) the face-only area, (ii) the skin-only region, and
(iii) the eye region, as shown in Figure~\ref{fig:guidelines}.
These masks are provided to guide your attention toward the areas we want you to evaluate and to reduce distractions from the background or other elements in the scene.

\begin{enumerate}
\item Which image has more contrast? For this task, we make three separate annotations. Please note that we do not evaluate \textit{color contrast} or \textit{texture contrast} in the photographic sense. Here, \textit{color contrast} refers to contrast created by differences in color, while \textit{texture contrast} refers to contrast created by differences in surface texture. Instead, we focus only on tonal contrast for the \textit{overall contrast} and \textit{eye-region contrast} questions, and on \textit{micro-contrast} for the skin-detail evaluation. Also, we are \emph{not} choosing which image looks better. The goal is simply to decide which image has \emph{more contrast}. In some cases, the segmented face mask may include small black pixels along the mask boundary. These black pixels are segmentation artifacts and should be completely ignored during the evaluation.

\item Which image has more balanced contrast? For this task, you will see the full face crop region (red box in Figure~\ref{fig:guidelines}). Here, annotators select the image they prefer between the two images.
\end{enumerate}

\subsection{Overall Contrast}

\begin{figure}[t]
    \centering
    \begin{subfigure}{0.48\linewidth}
        \centering        \includegraphics[width=0.5\linewidth]{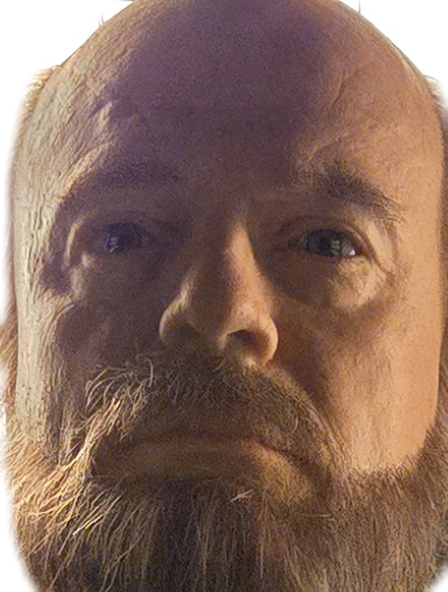}
        \caption{5~lux, LED~2700K,\\ $\Delta \mathrm{EV}=9.0$}
        \label{fig:illum_212030}
    \end{subfigure}
    \hfill
    \begin{subfigure}{0.48\linewidth}
        \centering        \includegraphics[width=0.5\linewidth]{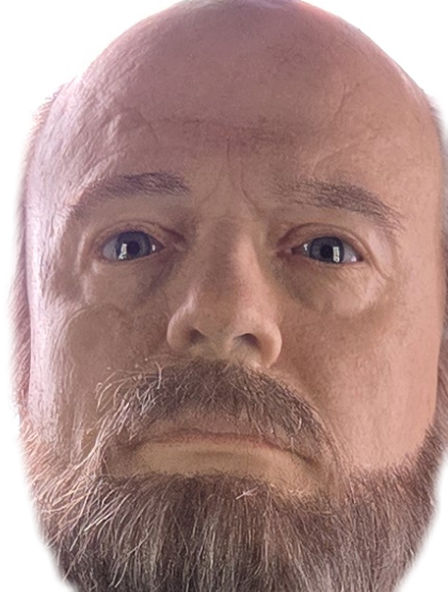}
        \caption{100~lux, LED~4000K, \\$\Delta \mathrm{EV}=7.0$}
        \label{fig:illum_212028}
    \end{subfigure}
    \vspace{0.2em}
    \hfill
    \begin{subfigure}{0.48\linewidth}
        \centering       \includegraphics[width=0.5\linewidth]{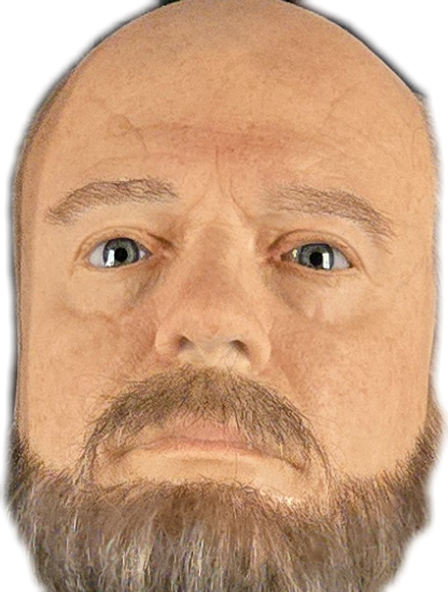}
        \caption{100~lux,LED~4000K, \\$\Delta \mathrm{EV}=0.0$}
        \label{fig:illum_212026}
    \end{subfigure}
    \begin{subfigure}{0.48\linewidth}
        \centering        \includegraphics[width=0.5\linewidth]{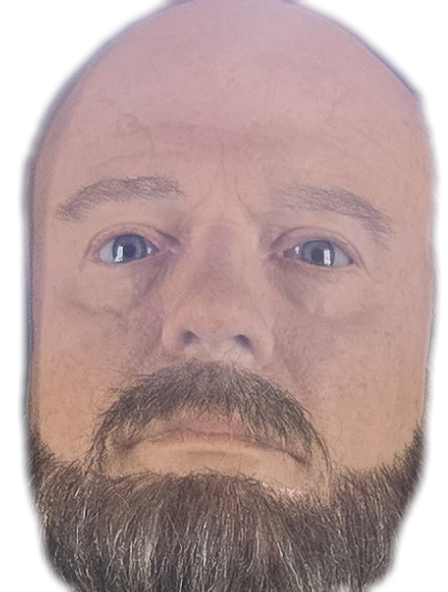}
        \caption{1000~lux, SME\_D65,\\ $\Delta \mathrm{EV}=0.0$}
        \label{fig:illum_212022}
    \end{subfigure}
    \caption{Examples of the same scene captured under different illumination 
    levels and exposure settings.}
    \label{fig:illum_examples_four}
\end{figure}

In this task, annotators are shown whole-face masked images without any background. As shown in Figure~\ref{fig:illum_examples_four}, they compare several face images taken under different illumination and exposure settings. When we talk about overall contrast, we mainly mean tonal contrast: the difference between the dark and bright parts of the face. Bright regions such as the sclera and dark regions such as the eye sockets naturally contribute to the overall tonal range of the face. This is especially important for dark-skinned mannequins, where the eyes may contain the strongest bright--dark separation. However, this task is distinct from the Eye-Region Contrast task: Overall Contrast measures global tonal separation across the whole face, whereas Eye-Region Contrast evaluates only the local tonal structure inside the eye region (e.g., sclera vs.\ iris vs.\ pupil).

Lighting can strongly influence overall contrast. For example, in Figures~\ref{fig:illum_212030} and~\ref{fig:illum_212028}, strong directional lighting creates a clear separation between bright and dark areas, so these images have higher overall contrast. In contrast, the image in Figure~\ref{fig:illum_212026} does not have strong lighting differences, so the contrast is more moderate and even. It is important to remember that we are not choosing the image that looks the most pleasing or natural. For example, Figure~\ref{fig:illum_212026} may look the most pleasing overall, but our goal here is simply to decide which image has more contrast. Finally, in some cases, such as Figure~\ref{fig:illum_212022}, the face may look flat because of flare or veiling light. This reduces the separation between bright and dark areas, so we consider it to have low overall contrast.

\begin{figure}[t]
        \centering        \includegraphics[width=\columnwidth]{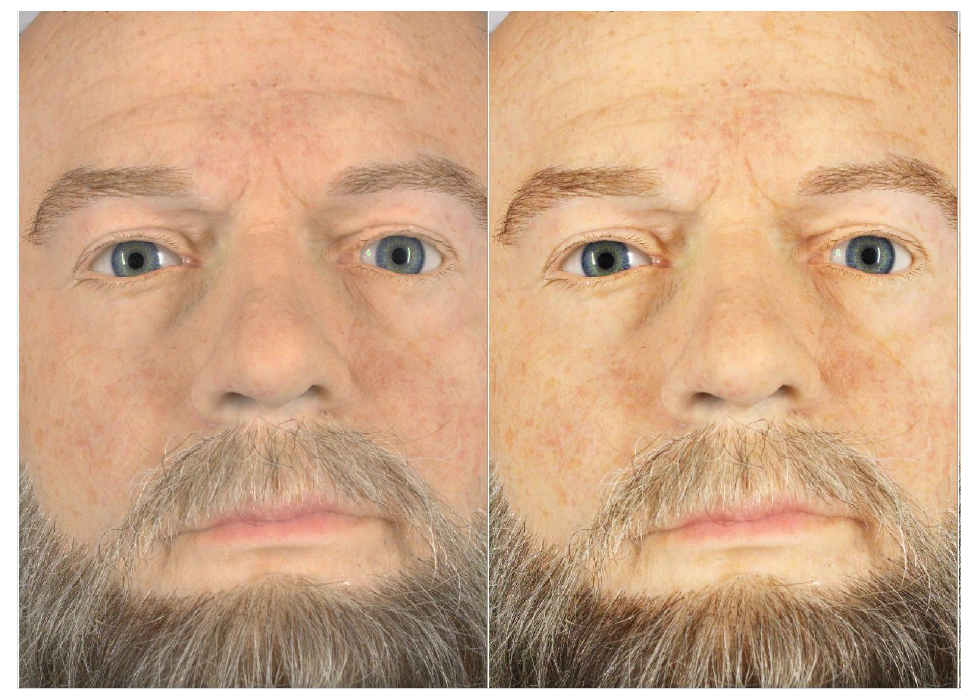}
        \caption{Tonal contrast (Overall contrast) example
        }
        \label{fig:tonal_contrast_example}
\end{figure}

Figure~\ref{fig:tonal_contrast_example} shows an example of a tonal contrast difference. The right image has higher tonal contrast than the left image. In the right image, the pupil is darker and the sclera is brighter. In the beard area, the white hairs appear brighter and the dark hairs appear darker. The separation between individual hairs is also clearer.

\subsection{Eye-Region Contrast}

In this task, annotators are shown only the eye masks, without any background.
When judging contrast in the eyes, consider the following: contrast is higher when the pupil appears darker and the white of the eye appears brighter. Sometimes, there may be small white reflections on the pupil caused by light. In such cases, a brighter reflection also contributes to a higher-contrast appearance. Please note that natural eye color, such as blue or brown, should not influence the decision. Eye color is not related to the contrast we want to measure.

\subsection{Micro-Contrast}

In this task, annotators are shown only the skin masks, without any background. The goal is to judge how clearly the small skin textures appear. Please focus on details such as pores, fine lines, and eyebrow or forehead texture. When these tiny details look sharper and more visible, the image has higher micro-contrast. When they look soft or blurred, the micro-contrast is lower. For the Eugene mannequin, the beard area may also be visible. If the beard texture appears clear, this also counts as higher micro-contrast. Please note that micro-contrast is not about color or lighting. For example, an image taken in a dark environment may look very noisy, making it hard to see the skin texture. In that case, even if the color or brightness is different, the micro-contrast should still be judged as low because the fine details are not visible.

Figure~\ref{fig:micro_contrast_example} shows an example of micro-contrast. The right image has higher micro-contrast. When assessing micro-contrast, however, we focus on local details such as the eyebrows, forehead, pores, and beard. In the right image, the pores are more textured and stand out more. The separation between dark and bright hairs is more visible, and the wrinkles on the forehead stand out more clearly. The right image therefore shows richer local detail.

\subsection{Balanced Contrast Preference}

In this task, annotators are shown the face crop region with the background. They select the image they prefer from the two images shown in the comparison tool.

\section{Detailed Annotation Guidelines}
The annotations should be performed according to the specific guidelines. To assist the annotators during this process, we associate each guideline with a question that helps clarify what needs to be evaluated. Moreover, we provide brief practical guidance for annotators when choosing one image from the two displayed. Contrast rendition can be assessed according to the following guidelines and criteria:

\subsection{Overall Contrast Guideline}

Overall contrast means tonal contrast: the intrinsic tonal depth of the image. The main question is: ``Does the face show a wide range of tones—from deep dark areas to clean highlights, with distinct midtones in between?''
This refers to the global tonal rendering of the image, independent of lighting direction.

In the pairwise comparison tool, the short version is displayed: \textit{``Which image has the larger range from darks to highlights?''}

\vspace{0.5cm}

\noindent We now describe how to evaluate each aspect in detail.
Tonal contrast should be judged by looking at how different the bright, medium, and dark parts of the face appear.

\vspace{0.5cm}

\begin{figure}[t]
        \centering       \includegraphics[width=0.99\linewidth]{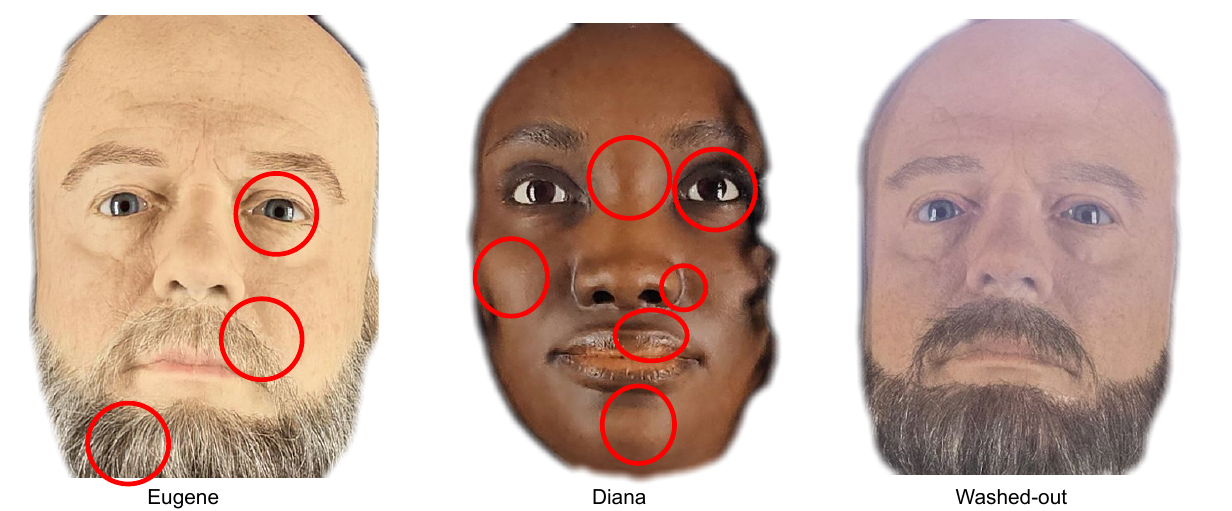}
        \caption{ Dark and bright regions to inspect for Eugene and Diana, with a washed-out example for overall contrast.}
        \label{fig:wheretolookt}
\end{figure}

\begin{figure}[t]
        \centering
    \includegraphics[width=\columnwidth]{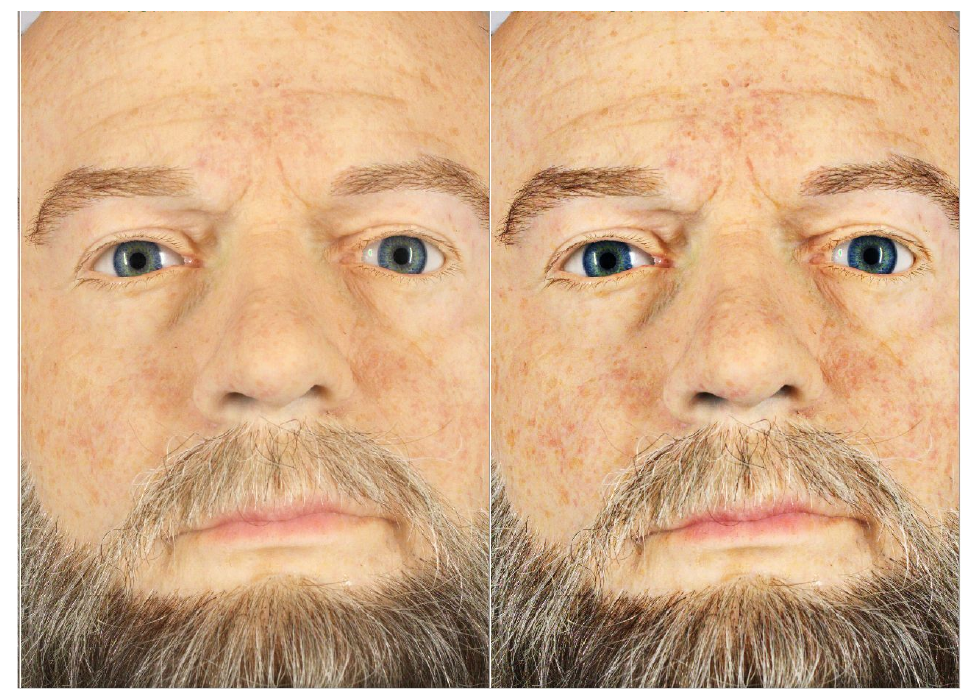}
        \caption{Example of micro-contrast: the image on the right exhibits higher micro-contrast than the image on the left.}
        \label{fig:micro_contrast_example}
\end{figure}

\begin{itemize}
    \item \textbf{Bright vs.\ Dark Areas on the Face:}  
    Look for clearly bright areas and clearly dark areas on the face, as shown in Figure~\ref{fig:wheretolookt}.
    Faces with higher contrast show obvious differences between light and dark regions.
    \begin{itemize}
        \item For the \textbf{Eugene mannequin}, compare the white hair to the dark hair, and check the eyes.
        \item For the \textbf{Diana mannequin}, look at bright skin areas (chin, cheeks) versus darker areas, and the eyes. Look at the shadow around the lips and the highlight on the lips too.
    \end{itemize}
    If the bright parts look truly bright and the dark parts look clearly dark, the contrast is higher.
    If everything looks similar in tone, the contrast is low.

    \item \textbf{Does the Face Look Flat or Washed Out?}  
    Check whether the face looks clear and solid, or flat and foggy.
    A low-contrast face often looks grayish, dull, or veiled, as if a thin haze covers it.
    You should still be able to see normal facial details between the bright and dark areas.
    If the middle tones disappear and everything blends together, the contrast is low.
\end{itemize}

\subsection{Eye-Region Contrast Guideline}

\noindent Main question: ``Do the black and white parts of the eyes look truly black and truly white, or do they look gray and washed out?''
High eye contrast means that the black pupil, the iris, and the white of the eye are easy to tell apart.

In the pairwise comparison tool, the short version is displayed: \textit{``Select the image with the darker pupil and brighter sclera (white of the eye).''}

\vspace{0.5cm}

\begin{figure}[t]
        \centering        \includegraphics[width=\columnwidth]{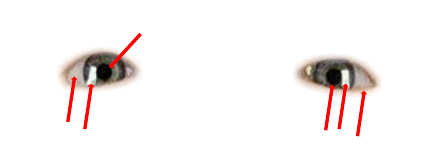}
        \caption{Regions to inspect for eye contrast: pupil darkness, catch-light brightness, and sclera brightness.}
        \label{fig:wheretolookat3}
\end{figure}

\textbf{Check these two things (for both mannequins), as shown in Figure~\ref{fig:wheretolookat3}:}
\begin{itemize}
    \item \textbf{White of the Eye:}  
    Look at the white part of the eye (sclera).
    Does it look white and bright, or does it look gray, yellow, or dull?
    Whiter whites mean higher eye contrast.

    \item \textbf{Black of the Eye:}  
    Look at the black center of the eye (pupil).
    Does it look deep black, or does it look gray or milky?
    Darker blacks mean higher eye contrast.
\end{itemize}

\vspace{0.5cm}

\textbf{Mannequin-specific reminders:}
\begin{itemize}
    \item \textbf{Eugene (blue eyes):}  
    Ignore how strong the blue color looks.
    Only judge whether the pupil looks clearly black compared to the rest of the eye.

    \item \textbf{Diana (brown eyes):}  
    Check whether the black pupil is darker than the brown iris.
    If the pupil and iris blend together into the same dark tone, eye contrast is low.
\end{itemize}

\subsection{Micro-Contrast Guideline}

Main question: ``Can you clearly see the tiny details on the skin surface?'' When evaluating micro-contrast, we are looking for local regions where we can see details and texture. High micro-contrast means that the texture looks very crisp. Low micro-contrast causes the skin to appear overly smooth, plastic, or softened. Note that higher micro-contrast does not mean that the image looks more pleasant.

In the pairwise comparison tool, the short version is displayed:
\textit{``Choose the image where the tiny details on the skin surface are more visible.''}

\vspace{0.5cm}

\begin{figure}[t]
        \centering
        \includegraphics[width=\columnwidth]{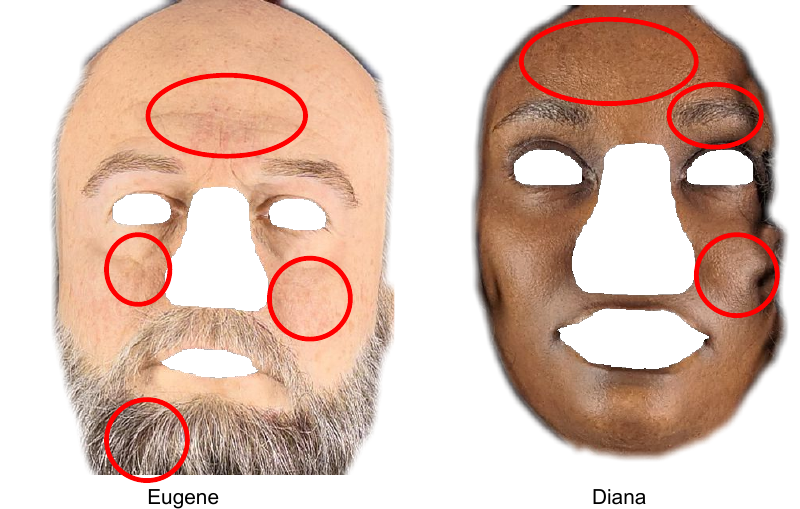}
        \caption{Regions to inspect for micro-contrast in Eugene and Diana.}
        \label{fig:wheretolookat2}
\end{figure}

For Eugene, we focus on these questions by looking at the regions of interest shown in Figure~\ref{fig:wheretolookat2}:
\begin{itemize}
    \item \textbf{Beard Separation:} Look closely at the beard. Can you see a clear, sharp separation between the white hairs and the dark hairs, or do they blend into a gray mix?
    \item \textbf{Mustache Detail:} Are the individual hairs of the mustache well preserved and crisp, or does the mustache look like a blurry, soft patch?
    \item \textbf{Cheek / Forehead Texture:} Can you see the tiny holes (pores) on the cheeks, or is the skin smoothed out?
\end{itemize}

\vspace{0.5cm}

For Diana, we focus on these questions:
\begin{itemize}
    \item \textbf{Forehead Detail:} Look at the forehead area. Can you see actual skin texture and fine lines, or does the skin look too smooth (like plastic)?
    \item \textbf{Eyebrow Definition:} Since there is no beard, look at the eyebrows. Can you count individual eyebrow hairs, or do they look like a painted smudge?
    \item \textbf{Over-smoothing Check:} Does the skin look naturally textured, or has the camera softened it too much, making it look artificial?
\end{itemize}

\subsection{Balanced Contrast Preference Guideline}
\noindent Main question: \textit{``Which image has more balanced contrast?''}
Balanced contrast refers to the image whose facial contrast appears most natural and visually well balanced overall. Annotators should consider whether the face has sufficient tonal separation without appearing too flat, too harsh, over-sharpened, or unnatural. Unlike the previous three tasks, this judgment is based on overall preference rather than the amount of a single contrast component.

\section{Ablation Study}

\subsection{Visualization of Overall Contrast}

\begin{figure*}[t]
\centering
\includegraphics[width=0.99\linewidth]
{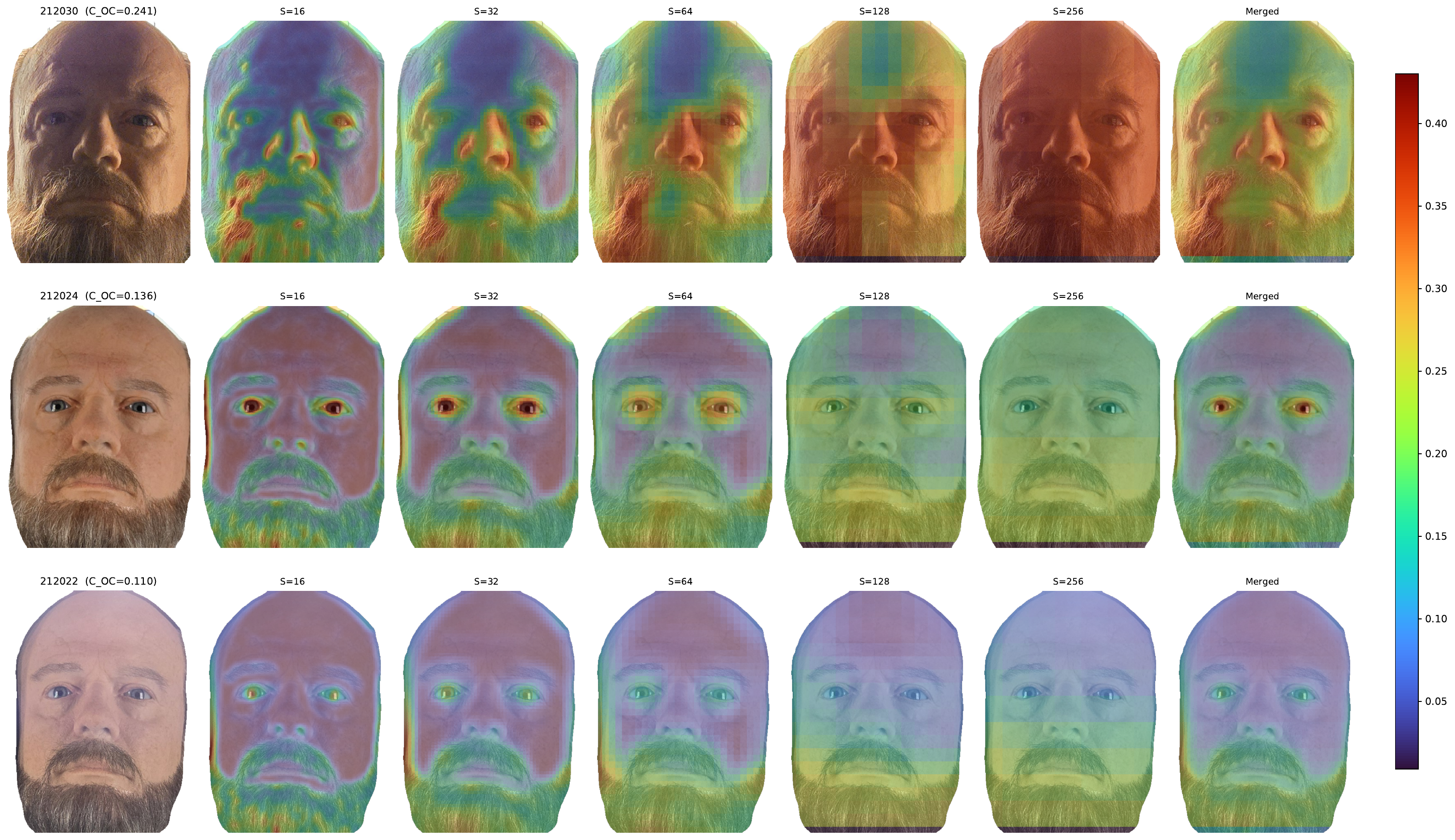}
\caption{Visualization of the multi-scale overall-contrast formulation for three example images. For each image, we show the original face mask, the local normalized contrast map computed at each scale $r \in \mathcal{R}$, where $\mathcal{R}=\{16,32,64,128,256\}$, and the resulting merged contrast map. Smaller scales emphasize fine local variations, while larger scales capture broader facial shading and illumination structure.}

\label{fig:multiscale_example}
\end{figure*}

Figure~\ref{fig:multiscale_example} visualizes the multi-scale formulation used for overall contrast. For each example image, we show the local normalized contrast map computed at different window sizes, together with the resulting merged contrast map. Smaller windows respond more strongly to fine local variations, whereas larger windows emphasize broader facial shading and illumination structure. By combining these responses across scales, the proposed formulation captures overall facial contrast in a way that is more spatially informative than a single global luminance statistic.

\subsubsection{Baseline Formulations}

Traditional measures typically define contrast as the luminance difference between discrete facial features (eyes, eyebrows, lips) and the surrounding skin.

\begin{enumerate}
    \item \textbf{Russell (2009)~\cite{russell2009sex}:} Utilizes a simple luminance range (difference) to define contrast $C_{Russell}$:
    \begin{equation}
        C_{\text{Russell}} = L_{\text{skin}} - L_{\text{features}},
    \end{equation}
    where $L_{\text{skin}}$ and $L_{\text{features}}$ are the mean luminance values of the skin and facial feature pixels, respectively. This metric measures the absolute luminance gap without normalization.

    \item \textbf{Pochéron et al. (2013)~\cite{porcheron2013aspects}:} Adopted the Michelson contrast formulation to ensure the metric is scaled relative to the overall brightness:
    \begin{equation}
        C_{\text{Pochéron}} = \frac{L_{\text{skin}} - L_{\text{features}}}{L_{\text{skin}} + L_{\text{features}}}.
    \end{equation}
    This normalized version accounts for varying global illumination levels but still relies on region-wide averages.

    \item \textbf{Russel et al. (2016)~\cite{russell2016facial}:} Expanded the Michelson approach by calculating contrast for multiple individual features (eyes, lips, eyebrows) and combining them into a single \textit{Facial Contrast Vector Magnitude} $|\vec{C}_{J}|$:
    \begin{equation}
        |\vec{C}_{\text{Russel}}| = \sqrt{C_{\text{eyes}}^2 + C_{\text{lips}}^2 + C_{\text{eyebrows}}^2},
    \end{equation}
    where each $C_{i}$ is calculated using the Michelson formula as in Pochéron et al. (2013).
\end{enumerate}

\subsection{Overall Contrast Ablation Study}
Table~\ref{tab:overall_scale_ablation} examines the sensitivity of the proposed overall-contrast measure to the choice of spatial scales. We consider both leave-one-scale-out variants of the reference set, $\mathcal{R}=\{16,32,64,128,256\}$, and alternative scale sets obtained by shifting or densifying the sampled scales.
Removing either of the two finest scales (16 or 32 pixels) consistently decreases both PLCC and SRCC under the \emph{Within-lighting} and \emph{Mixed-lighting} evaluations. Removing the intermediate 64-pixel scale results in only a small change, whereas removing the two coarsest scales yields higher nominal correlations. In particular, excluding the 256-pixel scale gives the highest nominal correlations in the leave-one-scale-out analysis, reaching PLCC/SRCC values of $0.815/0.806$ within lighting and $0.729/0.729$ under mixed lighting, compared with $0.792/0.780$ and $0.688/0.693$, respectively, for the reference configuration. The bootstrap distributions nevertheless exhibit substantial overlap, indicating that the observed differences should be interpreted as a tendency rather than a definitive advantage of a particular scale set.

A similar trend is observed when perturbing the complete scale set. Shifting all scales toward smaller spatial support yields moderately higher nominal correlations, while shifting them toward larger support reduces performance. Increasing the sampling density using half-octave spacing produces results close to the reference configuration, suggesting that the metric does not benefit substantially from a denser sampling of the same spatial range.
Overall, these results indicate that the proposed measure is reasonably robust to moderate changes in the scale set, while also suggesting that the coarsest spatial support may be less informative for the face crops considered in our dataset. We retain the originally specified $R={16,32,64,128,256}$ configuration in the main experiments to avoid retrospectively selecting the scale set based on the evaluation annotations, and report the present analysis as a sensitivity study rather than using it for post-hoc parameter tuning.

\begin{table*}[t]
\centering
\caption{
Scale-set ablation of the proposed overall-contrast measure.
Scale values denote square-patch side lengths in pixels.
Results are reported as the mean $\pm$ standard deviation over
500 bootstrap resamples.
The symbol $\dagger$ indicates the reference configuration used in
the main experiments, while bold values indicate the highest nominal
correlation within each analysis block.
}
\label{tab:overall_scale_ablation}
\small
\setlength{\tabcolsep}{5pt}
\renewcommand{\arraystretch}{1.05}
\resizebox{0.8\textwidth}{!}{%
\begin{tabular}{lcccc}
\toprule
\multirow{2}{*}{Scale setting}
& \multicolumn{2}{c}{Within-lighting condition}
& \multicolumn{2}{c}{Mixed-lighting condition} \\
\cmidrule(lr){2-3}\cmidrule(lr){4-5}
& PLCC & SRCC & PLCC & SRCC \\
\midrule

\multicolumn{5}{l}{\textbf{Leave-one-scale-out analysis}} \\

$\{16,32,64,128,256\}^{\dagger}$
& $0.792 \pm 0.070$
& $0.780 \pm 0.081$
& $0.688 \pm 0.133$
& $0.693 \pm 0.097$ \\

Without 16
& $0.778 \pm 0.077$
& $0.755 \pm 0.092$
& $0.662 \pm 0.147$
& $0.671 \pm 0.107$ \\

Without 32
& $0.781 \pm 0.076$
& $0.765 \pm 0.090$
& $0.669 \pm 0.145$
& $0.678 \pm 0.104$ \\

Without 64
& $0.788 \pm 0.074$
& $0.773 \pm 0.088$
& $0.681 \pm 0.140$
& $0.687 \pm 0.102$ \\

Without 128
& $0.798 \pm 0.067$
& $0.791 \pm 0.077$
& $0.701 \pm 0.127$
& $0.705 \pm 0.091$ \\

Without 256
& $\mathbf{0.815 \pm 0.054}$
& $\mathbf{0.806 \pm 0.065}$
& $\mathbf{0.729 \pm 0.107}$
& $\mathbf{0.729 \pm 0.074}$ \\

\midrule

\multicolumn{5}{l}{\textbf{Scale-set perturbation}} \\

Smaller: $\{12,24,48,96,192\}$
& $\mathbf{0.808 \pm 0.059}$
& $\mathbf{0.796 \pm 0.070}$
& $\mathbf{0.713 \pm 0.119}$
& $\mathbf{0.714 \pm 0.085}$ \\

Reference: $\{16,32,64,128,256\}^{\dagger}$
& $0.792 \pm 0.070$
& $0.780 \pm 0.081$
& $0.688 \pm 0.133$
& $0.693 \pm 0.097$ \\

Larger: $\{20,40,80,160,320\}$
& $0.787 \pm 0.075$
& $0.770 \pm 0.085$
& $0.665 \pm 0.147$
& $0.677 \pm 0.104$ \\

\shortstack[l]{Half-octave dense:\\
$\{16,23,32,45,64,91,128,181,256\}$}
& $0.797 \pm 0.068$
& $0.788 \pm 0.074$
& $0.692 \pm 0.130$
& $0.696 \pm 0.094$ \\

\bottomrule
\end{tabular}%
}
\end{table*}

\subsection{Eye Contrast Ablation Study}

As a simple baseline, eye-region contrast can be defined using a Michelson-style normalized range computed from percentile extrema within the eye mask:
\begin{equation}
    L_{\min} = P_{1}(L_{\mathrm{eye}}), \qquad
    L_{\max} = P_{99}(L_{\mathrm{eye}}),
\end{equation}
and
\begin{equation}
    C_{\mathrm{eye}}
    = \frac{L_{\max} - L_{\min}}
           {L_{\max} + L_{\min}}.
\end{equation}
This formulation measures the separation between the darkest and brightest structures in the eye region and can provide a reasonable approximation when the pupil/iris and sclera are clearly separated.

\begin{figure*}[t]
\centering
\includegraphics[width=0.99\linewidth]{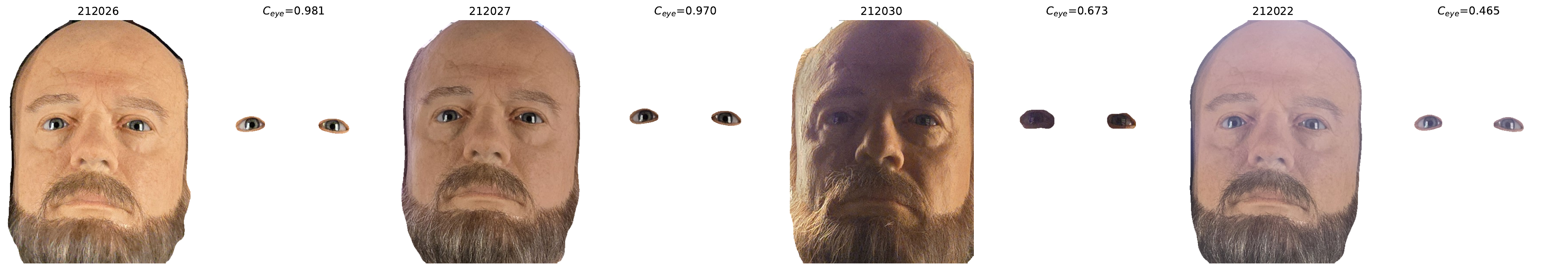}
\caption{Face images and corresponding eye-region crops ordered by the Michelson eye-contrast measure~\cite{michelson1995studies}. While this formulation provides a reasonable ranking in some cases, it can fail in low-contrast regimes. In particular, the perceptual ordering between Images~212022 and~212030 is not preserved. For visualization, sRGB images are shown, but the metric is computed on the $L^*$ channel only.}
\label{fig:eye_examples}
\end{figure*}

However, this normalized range becomes less reliable in low-contrast cases. When both the bright and dark structures are compressed into a narrow luminance range, the resulting score may no longer reflect the perceived separation between the pupil, iris, and sclera. As a result, the Michelson-style ordering can become perceptually unstable. Figure~\ref{fig:eye_examples} illustrates such a failure case: although Image~212030 appears to have lower eye contrast than Image~212022, the Michelson~\cite{michelson1995studies} measure does not preserve this perceptual ordering.

\begin{figure*}[t]
    \centering
    \includegraphics[width=0.99\linewidth]{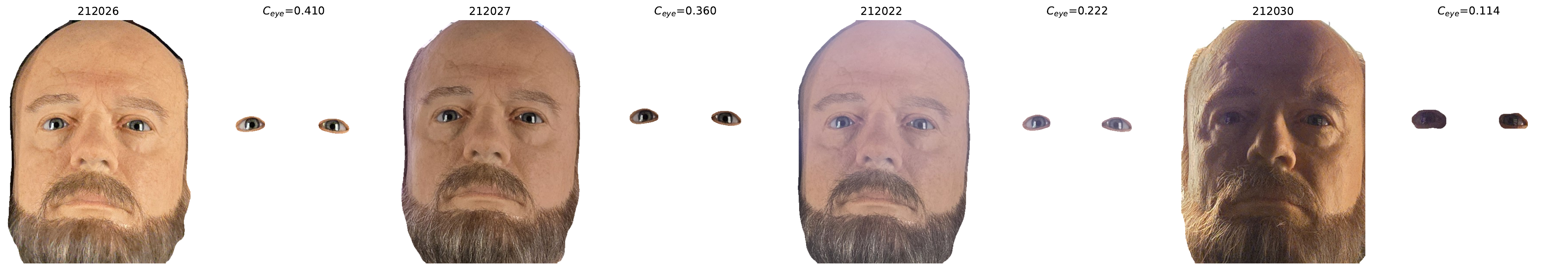}
    \caption{Low-contrast eye-region examples ordered by the proposed Otsu-based eye-contrast measure~\cite{otsu1975threshold}. Although these images exhibit weaker dark--bright separation than high-contrast cases, the proposed formulation still yields a perceptually meaningful ranking.}
    \label{fig:otsu_low_contrast}
\end{figure*}

For this reason, we adopt the Otsu-based formulation in the main method. By partitioning the luminance values within the eye mask into dominant dark and bright classes, the proposed measure better reflects the perceptually meaningful separation between the darker pupil/iris region and the brighter sclera and highlights. Figure~\ref{fig:otsu_low_contrast} further shows that even in a challenging low-contrast regime, the Otsu-based~\cite{otsu1975threshold} formulation still yields a perceptually plausible ranking.

\subsection{Micro-Contrast Ablation Study}

We first evaluate the contribution of the three main stages in the proposed micro-contrast formulation: local energy pooling, perceptual response compression, and robust spatial pooling.
Tables~\ref{tab:micro_ablation_protocolA}
and~\ref{tab:micro_ablation_protocolB} report the component ablation results under the within-lighting and mixed-lighting evaluations, respectively. In these experiments, we use the reference configuration
$\sigma_f=1/\sqrt{2}\approx0.7$,
$\sigma_c/\sigma_f=3$,
$W=11$, $\gamma=0.6$, and
$[q_{10},q_{90}]$ pooling.
Each component is removed individually while all other applicable parameters are held fixed. Specifically, removing robust spatial
pooling replaces the $[q_{10},q_{90}]$ trimmed mean with an untrimmed spatial mean. Removing perceptual response compression sets
$\gamma=1$, while retaining the reference scale parameters, local RMS integration with an $11\times11$ window, and robust pooling. Removing local energy pooling replaces the spatially localized RMS response with a single global DoG statistic computed using the same fine and coarse scales.

Under \emph{Within-lighting
condition}, the complete formulation achieves a PLCC/SRCC of $0.848\pm0.024$/$0.841\pm0.028$. Removing robust spatial pooling reduces the correlations to $0.825\pm0.026$/$0.814\pm0.029$, demonstrating that percentile trimming improves robustness to isolated responses caused by
specular highlights, shadow noise, and mask-boundary artifacts. Removing perceptual response compression reduces the result to
$0.802\pm0.029$/$0.815\pm0.029$. The larger reduction in PLCC than in SRCC suggests that the nonlinear transducer contributes
particularly to the linear calibration between the objective response and subjective JOD, while having a smaller effect on rank ordering.
The largest degradation occurs without local energy pooling, for which performance decreases to $0.664\pm0.069$/$0.668\pm0.077$. The same trend is observed under Mixed-lighting condition. The complete formulation achieves $0.797\pm0.011$/$0.787\pm0.019$, compared with
$0.775\pm0.011$/$0.763\pm0.000$ without robust spatial pooling and $0.736\pm0.019$/$0.763\pm0.001$ without perceptual response compression. Replacing local energy pooling with a global DoG
statistic further reduces performance to
$0.631\pm0.102$/$0.629\pm0.077$. These results consistently indicate that micro-contrast is better represented as a spatially localized
texture response than as a single global high-frequency statistic.

\begin{figure*}[t]
    \centering
    \includegraphics[width=0.98\textwidth]
    {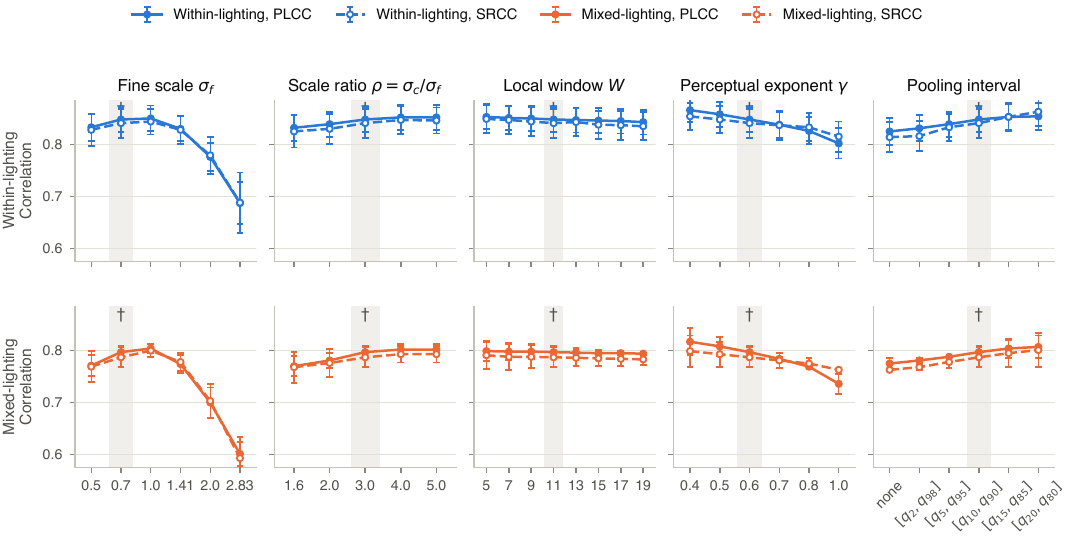}
    \caption{
    Parameter sensitivity of the proposed micro-contrast measure under the \emph{Within-lighting condition} (top) and mixed-lighting (bottom) evaluations.
    We vary the fine scale $\sigma_f$, scale ratio
    $\rho=\sigma_c/\sigma_f$, local integration window $W$, perceptual exponent $\gamma$, and robust pooling interval. Solid and dashed
    lines denote PLCC and SRCC, respectively, and error bars represent the standard deviation over 500 bootstrap resamples. The gray bands
    and $\dagger$ symbols indicate the reference configuration used in the main experiments:
    $\sigma_f=1/\sqrt{2}\approx0.7$, $\rho=3$, $W=11$,
    $\gamma=0.6$, and $[q_{10},q_{90}]$ pooling.
    }
    \label{fig:micro_parameter_sensitivity}
    \vspace*{-0.8em}
\end{figure*}

In addition to the component ablation, we evaluate the sensitivity of the formulation to its numerical parameters. Figure~\ref{fig:micro_parameter_sensitivity} reports a one-factor-at-a-time sensitivity analysis over the fine scale $\sigma_f$, coarse-to-fine scale ratio  $\rho=\sigma_c/\sigma_f$,
local integration window $W$, perceptual exponent $\gamma$,
and robust pooling interval.
These parameters were fixed before the evaluation according to the adopted scale-space formulation, perceptual model, and implementation
design, rather than selected by maximizing performance on the evaluation sets.

The settings $\sigma_f=1/\sqrt{2}\approx0.7$ and $\sigma_f=1.0$ produce comparable correlations, whereas performance decreases at $\sigma_f=\sqrt{2}$ and coarser scales. This trend indicates that
relatively fine spatial scales are more suitable for representing the facial details associated with subjective micro-contrast. Performance
is also stable when the coarse-to-fine scale ratio is varied from $\rho=3$ to $\rho=5$: relative to $\rho=3$, the largest absolute
change across the four reported correlations is only $0.006$. We therefore retain the reference ratio $\rho=3$.

\paragraph{Local window size.}
Changing the local integration window from $5\times5$ to $19\times19$ produces limited variation. Under the \emph{Within-lighting}
evaluation, PLCC ranges from $0.843$ to $0.853$ and SRCC from $0.835$ to $0.849$. Under the mixed-lighting evaluation, the corresponding
ranges are $0.794$--$0.799$ and $0.783$--$0.791$. By contrast, replacing local RMS integration with a single global DoG statistic
reduces PLCC and SRCC under both settings, as reported in Tables~\ref{tab:micro_ablation_protocolA}
and~\ref{tab:micro_ablation_protocolB}. These results demonstrate the importance of spatially localized energy integration while showing that performance is only weakly affected by its exact window size. We retain the intermediate $11\times11$ window.

\paragraph{Perceptual response compression.}
Response compression improves correlation relative to the uncompressed setting ($\gamma=1$). The highest nominal correlations
are obtained with $\gamma=0.4$, while the reference value $\gamma=0.6$ gives slightly lower results. The differences between these settings are $0.018$ in PLCC and $0.013$ in SRCC under the \emph{Within-lighting} evaluation, and $0.020$ and $0.012$, respectively,
under the \emph{Mixed-lighting} evaluation. We retain $\gamma=0.6$ because it was specified from the adopted perceptual response model before
the sensitivity analysis rather than selected retrospectively.

\paragraph{Percentile trimming.}
Percentile trimming improves performance relative to untrimmed spatial averaging. Replacing the untrimmed mean with $[q_{10},q_{90}]$ pooling
increases PLCC/SRCC from $0.825$/$0.814$ to $0.848$/$0.841$ under the \emph{Within-lighting} evaluation and from $0.775$/$0.763$ to $0.797$/$0.787$ under the \emph{Mixed-lighting} evaluation. Narrower intervals provide further nominal improvements; however, the reference interval $[q_{10},q_{90}]$ retains the central $80\%$ of spatial responses and provides a less aggressive form of outlier rejection.

Overall, the component ablations show that local energy integration, perceptual response compression, and robust spatial pooling each contribute to agreement with subjective judgments. The parameter
sweeps further demonstrate limited sensitivity to the local window size and scale ratio, together with gradual and interpretable trends for the fine scale, perceptual exponent, and pooling interval.
Although some neighboring configurations achieve higher nominal correlations, the reference configuration provides consistently strong performance under both evaluation settings without post-hoc parameter
selection.

\begin{table}[t]
\centering
\caption{
Ablation study of the proposed micro-contrast formulation under \emph{Within-lighting condition}. Results are reported as the mean $\pm$ standard deviation over 500 bootstrap resamples.
}
\label{tab:micro_ablation_protocolA}
\small
\setlength{\tabcolsep}{4pt}
\resizebox{\columnwidth}{!}{%
\begin{tabular}{lcc}
\toprule
Variant & PLCC & SRCC \\
\midrule
Proposed micro-contrast
& \textbf{$0.849 \pm 0.024$}
& \textbf{$0.842 \pm 0.028$} \\

w/o robust spatial pooling
& $0.825 \pm 0.026$
& $0.814 \pm 0.029$ \\

w/o perceptual response compression
& $0.802 \pm 0.029$
& $0.815 \pm 0.029$ \\

w/o local energy pooling
& $0.664 \pm 0.069$
& $0.668 \pm 0.077$ \\
\bottomrule
\end{tabular}
}
\end{table}

\begin{table}[t]
\centering
\caption{
Ablation study of the proposed micro-contrast formulation under
\emph{Mixed-lighting condition}. Results are reported as the mean $\pm$ standard deviation
over 500 bootstrap resamples.
}
\label{tab:micro_ablation_protocolB}
\small
\setlength{\tabcolsep}{4pt}
\resizebox{\columnwidth}{!}{%
\begin{tabular}{lcc}
\toprule
Variant & PLCC & SRCC \\
\midrule
Proposed micro-contrast
& \textbf{$0.797 \pm 0.011$}
& \textbf{$0.787 \pm 0.019$} \\

w/o robust spatial pooling
& $0.775 \pm 0.011$
& $0.763 \pm 0.000$ \\

w/o perceptual response compression
& $0.736 \pm 0.019$
& $0.763 \pm 0.001$ \\

w/o local energy pooling
& $0.631 \pm 0.102$
& $0.629 \pm 0.077$ \\
\bottomrule
\end{tabular}
}
\end{table}

\subsection{Per-Mannequin Analysis}

Table~\ref{tab:per_mannequin_within} reports the within-lighting results
separately for Diana and Eugene. All three proposed metrics remain positively correlated with their corresponding perceptual judgments for both mannequins. Eye and micro-contrast show similar performance across the two identities,
while overall contrast exhibits greater variation.

\begin{table}[t]
\centering
\caption{
Per-mannequin performance of the proposed contrast metrics under the
within-lighting evaluation. Correlations are macro-averaged over
Groups~1--4 for each mannequin. Uncertainty denotes the standard
deviation over 500 bootstrap resamples of the four group-level
correlations.
}
\label{tab:per_mannequin_within}
\footnotesize
\setlength{\tabcolsep}{4pt}
\resizebox{\columnwidth}{!}{%
\begin{tabular}{lcccc}
\toprule
\multirow{2}{*}{Contrast}
& \multicolumn{2}{c}{Diana}
& \multicolumn{2}{c}{Eugene} \\
\cmidrule(lr){2-3}\cmidrule(lr){4-5}
& PLCC & SRCC & PLCC & SRCC \\
\midrule
Overall contrast
& $0.880 \pm 0.017$
& $0.875 \pm 0.026$
& $0.705 \pm 0.129$
& $0.691 \pm 0.145$ \\

Eye contrast
& $0.934 \pm 0.015$
& $0.919 \pm 0.007$
& $0.824 \pm 0.064$
& $0.784 \pm 0.101$ \\

Micro-contrast
& $0.863 \pm 0.040$
& $0.862 \pm 0.043$
& $0.835 \pm 0.026$
& $0.822 \pm 0.033$ \\
\bottomrule
\end{tabular}%
}
\end{table}

\subsection{Balanced Contrast Preference Ablation Study}

We compare three balanced contrast preference-model variants using the proposed contrast measures as predictors: a non-negative linear regressor, a second-order polynomial regressor, and a pairwise-trained MLP. The second-order polynomial augments the three contrast components with their squared and cross-component interaction terms. 
The pairwise-trained MLP uses the same augmented representation and independently maps each image to a scalar balanced contrast score. It is trained using pairwise ranking supervision together with a balanced contrast JOD regression regularizer, following the training protocol described in the next subsection. 
Table~\ref{tab:proposed_fitting_variants} compares the three variants under the within-lighting and mixed-lighting evaluations.

The two evaluations differ in how the balanced contrast preference predictors are fitted and evaluated. In the within-lighting evaluation, a separate model is fitted for each mannequin and each of Groups~1--4 using the corresponding device-disjoint training split, and evaluated on the held-out devices from the same group. Because the balanced contrast JOD scale is estimated independently for each group and mannequin, each model is fitted and evaluated within its corresponding JOD scale. In the mixed-lighting evaluation, a single model per mannequin is fitted using Groups~1--4 and evaluated on the held-out Group~5 without refitting or using its balanced contrast labels. For each mannequin, the training balanced contrast JOD values therefore originate from four independently estimated group-specific scales, whereas Group~5 provides a single mixed-lighting JOD scale.
Within-lighting results are computed separately for the eight Group--mannequin evaluation subsets and then macro-averaged. Mixed-lighting results are computed separately for Diana and Eugene on Group~5 and then macro-averaged over the two mannequins.

As shown in Table~\ref{tab:proposed_fitting_variants}, the second-order polynomial regressor achieves the highest overall performance, reaching PLCC/SRCC values of $0.728/0.739$ within lighting and $0.773/0.778$ under mixed lighting. The pairwise-trained MLP is competitive but remains slightly below the polynomial model overall, while the non-negative linear regressor performs worse in both settings. We therefore use the second-order polynomial regressor in the main balanced contrast preference analysis, as it remains compact and explicitly parameterized while capturing quadratic effects and cross-component interactions.

\begin{table}[t]
\centering
\captionsetup[table]{justification=raggedright,singlelinecheck=false}
\caption{
Comparison of balanced contrast preference predictors under the
within-lighting and mixed-lighting evaluations.
}
\label{tab:proposed_fitting_variants}
\footnotesize
\setlength{\tabcolsep}{3pt}
\resizebox{\columnwidth}{!}{%
\begin{tabular}{lccc}
\toprule
Method & PLCC & SRCC & RMSE \\
\midrule
\multicolumn{4}{l}{\textbf{Within-lighting evaluation}} \\
Non-negative linear regressor
& $0.635 \pm 0.085$
& $0.654 \pm 0.073$
& $1.111 \pm 0.090$ \\
Second-order polynomial regressor
& $\mathbf{0.728 \pm 0.055}$
& $\mathbf{0.739 \pm 0.054}$
& $\mathbf{1.063 \pm 0.115}$ \\
Pairwise-trained MLP
& $0.708 \pm 0.055$
& $0.734 \pm 0.044$
& $1.118 \pm 0.114$ \\
\midrule
\multicolumn{4}{l}{\textbf{Mixed-lighting evaluation}} \\
Non-negative linear regressor
& $0.715 \pm 0.044$
& $0.682 \pm 0.016$
& $1.094 \pm 0.029$ \\
Second-order polynomial regressor
& $\mathbf{0.773 \pm 0.054}$
& $\mathbf{0.778 \pm 0.046}$
& $\mathbf{0.975 \pm 0.006}$ \\
Pairwise-trained MLP
& $0.759 \pm 0.062$
& $0.756 \pm 0.049$
& $0.983 \pm 0.024$ \\
\bottomrule
\end{tabular}%
}
\end{table}

\paragraph{Sensitivity to the training JOD scales.}
Given the independently estimated Group--mannequin balanced contrast JOD scales described above, we test whether differences in their locations and dispersions affect mixed-lighting performance. We refit the second-order polynomial model after z-normalizing the balanced contrast JOD training targets separately within each Group--mannequin comparison set. The normalization is applied only to the Groups~1--4 training targets; Group~5 remains unchanged.
As shown in Table~\ref{tab:preference_jod_normalization}, the mixed-lighting correlations are essentially unchanged: PLCC and SRCC change from $0.773/0.778$ with the original balanced contrast JOD targets to $0.774/0.775$ with Group--mannequin z-normalized targets. This indicates that the mixed-lighting correlations are not materially sensitive to differences in the locations and dispersions of the independently estimated source JOD scales. We therefore retain the original balanced contrast JOD targets used in the main experiments.

\begin{table}[t]
\centering
\caption{
Sensitivity of the second-order polynomial balanced contrast preference
predictor to normalization of the Groups~1--4 training targets.
}
\label{tab:preference_jod_normalization}
\footnotesize
\setlength{\tabcolsep}{5pt}

\resizebox{0.9\columnwidth}{!}{%
\begin{tabular}{lcc}
\toprule
Training target & PLCC & SRCC \\
\midrule
Raw balanced contrast JOD
& $0.773 \pm 0.054$
& $0.778 \pm 0.046$ \\
Group--mannequin z-normalized JOD
& $0.774 \pm 0.050$
& $0.775 \pm 0.038$ \\
\bottomrule
\end{tabular}%
}

\end{table}

\subsubsection{Fitted Coefficients}

Table~\ref{tab:preference_coefficients} reports the coefficients of the non-negative linear and second-order polynomial models in standardized feature space. The ``Fitting data'' column identifies the data used to estimate each set of coefficients. The linear coefficients are constrained to be non-negative, whereas the second-order polynomial coefficients are estimated without sign constraints.
The models are fitted under independently estimated balanced contrast JOD scales. In addition, feature standardization is performed separately for each training set. We report the coefficients to make the fitted models and their parameter estimation explicit rather than to infer relative perceptual importance across lighting groups or mannequins.

\begin{table*}[t]
\centering
\caption{
Coefficients of the linear and second-order polynomial preference regressors
in standardized feature space.
For within-lighting evaluation, models are fitted separately within
Groups~1--4. For mixed-lighting evaluation, models are fitted on
Groups~1--4 and applied without refitting to the held-out Group~5 set.
}
\label{tab:preference_coefficients}
\scriptsize
\setlength{\tabcolsep}{3.0pt}

\begin{tabular}{llllccccccccc}
\toprule
Evaluation & Fitting data & Subject & Model &
$\alpha_O$ & $\alpha_E$ & $\alpha_M$ &
$\alpha_{O^2}$ & $\alpha_{E^2}$ & $\alpha_{M^2}$ &
$\alpha_{OE}$ & $\alpha_{OM}$ & $\alpha_{EM}$ \\ 
\midrule

\multirow{16}{*}{Within-lighting}
& \multirow{4}{*}{G1}
& \multirow{2}{*}{Diana}
& Linear
& 0.3737 & 0.3475 & 0.1114
& -- & -- & -- & -- & -- & -- \\ 

&
&
& Polynomial
& 0.6424 & 0.3075 & 0.2604
& -0.4039 & 0.0006 & -0.3136
& -0.0219 & 0.1000 & 0.2704 \\ 

&
&
\multirow{2}{*}{Eugene}
& Linear
& 0.1839 & 0.0000 & 0.2491
& -- & -- & -- & -- & -- & --  \\ 

&
&
& Polynomial
& 0.4860 & -0.2704 & 0.4312
& -0.4301 & -0.0327 & -0.4807
& -0.4350 & 0.7321 & 0.5574  \\ 

\cmidrule(lr){2-13}

& \multirow{4}{*}{G2}
& \multirow{2}{*}{Diana}
& Linear
& 0.0000 & 1.3586 & 0.3842
& -- & -- & -- & -- & -- & -- \\ 

&
&
& Polynomial
& -0.3201 & 1.3867 & 0.8643
& -2.0840 & -4.3643 & -1.9313
& 5.7795 & 1.2028 & 1.0930  \\ 

&
&
\multirow{2}{*}{Eugene}
& Linear
& 0.8423 & 0.3132 & 0.2065
& -- & -- & -- & -- & -- & --  \\ 

&
&
& Polynomial
& 1.0003 & 0.0538 & 0.2690
& -0.4833 & -0.3169 & -0.1009
& 0.4215 & -0.0670 & 0.0423 \\  

\cmidrule(lr){2-13}

& \multirow{4}{*}{G3}
& \multirow{2}{*}{Diana}
& Linear
& 0.6623 & 0.4001 & 0.0000
& -- & -- & -- & -- & -- & -- \\  

&
&
& Polynomial
& 0.8821 & 0.5117 & -0.0024
& -1.0349 & -0.8659 & -0.2808
& 1.4400 & 0.2754 & -0.0865 \\ 

&
&
\multirow{2}{*}{Eugene}
& Linear
& 0.2555 & 0.0448 & 0.4794
& -- & -- & -- & -- & -- & -- \\ 

&
&
& Polynomial
& -0.0577 & 0.0732 & 1.0654
& -0.8306 & -0.2800 & -0.8362
& -0.7581 & 1.4209 & 0.3522 \\ 

\cmidrule(lr){2-13}

& \multirow{4}{*}{G4}
& \multirow{2}{*}{Diana}
& Linear
& 0.7120 & 0.7386 & 0.0406
& -- & -- & -- & -- & -- & -- \\ 

&
&
& Polynomial
& 0.8128 & 0.6468 & 0.1072
& -0.1729 & -0.0588 & 0.0460
& 0.0686 & -0.0714 & 0.3098 \\ 

&
&
\multirow{2}{*}{Eugene}
& Linear
& 0.0000 & 0.7800 & 0.3131
& -- & -- & -- & -- & -- & -- \\ 

&
&
& Polynomial
& 0.0294 & 1.0096 & 0.3282
& -0.2183 & -0.2613 & 0.1306
& -0.6179 & 0.0959 & 0.4796 \\ 

\midrule

\multirow{4}{*}{Mixed-lighting}
& \multirow{4}{*}{G1--4}
& \multirow{2}{*}{Diana}
& Linear
& 0.1750 & 0.6530 & 0.3610
& -- & -- & -- & -- & -- & -- \\ 

&
&
& Polynomial
& 0.6543 & 0.0039 & 0.5306
& -0.5738 & -0.4526 & -0.3662
& 0.7366 & 0.2660 & 0.1194 \\ 

&
&
\multirow{2}{*}{Eugene}
& Linear
& 0.1768 & 0.3320 & 0.3802
& -- & -- & -- & -- & -- & --  \\ 

&
&
& Polynomial
& 0.4816 & 0.0155 & 0.3340
& -0.3749 & -0.3212 & -0.2911
& -0.1104 & 0.4249 & 0.1800 \\ 

\bottomrule
\end{tabular}
\end{table*}

\subsubsection{Training Details for Learned Balanced contrast Models}

To adapt the learned baselines to the balanced contrast preference task, we fine-tune lightweight trainable heads while keeping the pretrained representation largely frozen. For general IQA models, we optimize only the prediction head. For face-quality models, we keep the pretrained feature extractor or baseline score frozen and train only a small probe head on top. This design keeps the adaptation lightweight while allowing each model to align with the target preference supervision.
Training is performed using within-subset pair sampling. Specifically, image pairs are formed only within the same $(\mathrm{Group}, \mathrm{Person})$ subset, so that the model learns preference ordering under comparable lighting and subject conditions. Each sampled pair is ordered canonically as $(\mathrm{hi}, \mathrm{lo})$, where $\mathrm{hi}$ has a higher Preference JOD than $\mathrm{lo}$.
The main objective is a pairwise ranking loss,
\begin{equation}
\mathcal{L}_{\mathrm{rank}}
=
\operatorname{softplus}\!\left(-(s_{\mathrm{hi}}-s_{\mathrm{lo}})\right),
\end{equation}
where $s_{\mathrm{hi}}$ and $s_{\mathrm{lo}}$ are the predicted preference scores for the higher- and lower-JOD images, respectively. This loss encourages the preferred image to receive a larger predicted score.
To stabilize training across subsets, we additionally use a group-wise $z$-regularization term:
\begin{equation}
\mathcal{L}_{z}
=
\frac{1}{2}
\left[
\left(s_{\mathrm{hi}}-z_{\mathrm{hi}}\right)^2
+
\left(s_{\mathrm{lo}}-z_{\mathrm{lo}}\right)^2
\right],
\end{equation}
where $z_{\mathrm{hi}}$ and $z_{\mathrm{lo}}$ denote normalized target values computed within each $(\mathrm{Group}, \mathrm{Person})$ subset. The final optimization objective is
\begin{equation}
\mathcal{L}
=
\mathcal{L}_{\mathrm{rank}} + \lambda_{z} \mathcal{L}_{z},
\end{equation}

This adaptation protocol is intended to match the pairwise supervision format of the preference task rather than to redefine the proposed contrast components.  We set $\lambda_z=0.2$ in all adapted-baseline experiments. For
feature-probe models, each frozen feature dimension is standardized using
statistics estimated from the training split only. When affine calibration is
used for evaluation, its scale and offset are also estimated exclusively from the training split and then applied unchanged to the validation or test split.

\begin{figure*}[t]
    \centering
    \includegraphics[width=0.99\textwidth]{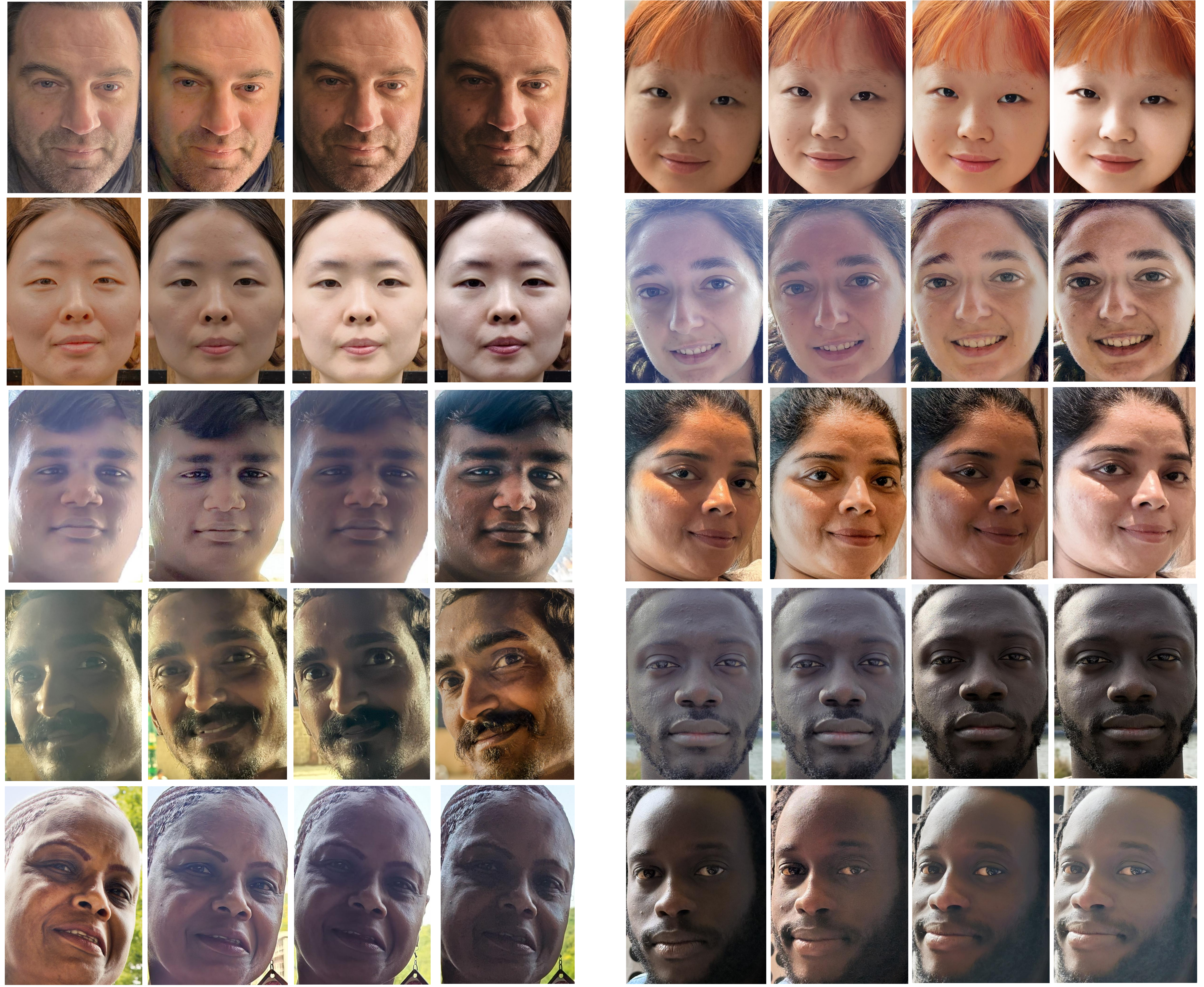}
    \caption{Real-portrait ablation study on overall contrast. A total of 444 images from 29 unique capture devices/sources are used. Each scene corresponds to a different subject and contains approximately 44 images captured with an average of five devices, together with contrast-adjusted variants generated using DxO PhotoLab 9. Human judgments were collected through pairwise comparisons for overall contrast, eye contrast, and micro-contrast. The proposed method shows strong agreement with the subjective annotations across diverse real-portrait scenes. From top to bottom, the rows correspond to: scene\_8005\_103 and scene\_1606\_031; scene\_0105\_089 and scene\_0109\_2; scene\_1119\_105 and scene\_1113\_510; scene\_3115\_305 and Riverside; and scene\_0109\_3 and scene\_5203\_106.}
    \label{fig:real_scenes}
\end{figure*}

\section{Real Portrait Generalization Study}

Figure~\ref{fig:real_scenes} shows representative examples from the real-portrait dataset, which contains 444 images across 10 scenes and 29 unique capture devices. Each scene corresponds to a different portrait subject and includes images captured with multiple devices. Table~\ref{tab:device_coverage} reports the device coverage for each scene. To support sufficient pairwise comparisons for subjective evaluation, each scene was expanded to approximately 44 images by including contrast-adjusted variants of the original captures. The resulting dataset spans diverse subjects, devices, and contrast renderings under less controlled real-world conditions. This experiment complements the controlled mannequin study by evaluating whether the proposed component-wise contrast metrics generalize to real portrait images.

\begin{table}[htbp]
\centering
\caption{Device distribution across real-portrait scenes (29 unique capture devices in total).}
\label{tab:device_coverage}
\begin{adjustbox}{max width=\linewidth}
\begin{tabular}{l c p{9cm}}
\toprule
Scene & Device Count & Devices \\
\midrule
0105\_089 & 8 & Apple iPhone 15 Pro Max, Sony Alpha 9 II, Honor Magic6 Pro, Huawei Pura 70 Ultra, Oppo Find X7 Ultra, Samsung Galaxy S24 Ultra, Vivo X100 Pro, Xiaomi 14 Ultra \\
0109\_2 & 7 & Apple iPhone 14 Pro, Google Pixel 7 Pro, Honor Magic5 Pro, Huawei P60 Pro, Oppo Find X6 Pro, Samsung Galaxy S23 Ultra, Sony A7R IV \\
0109\_3 & 7 & Apple iPhone 14 Pro, Google Pixel 7 Pro, Honor Magic5 Pro, Huawei P60 Pro, Oppo Find X6 Pro, Samsung Galaxy S23 Ultra, Sony A7R IV \\
1113\_510 & 5 & Apple iPhone 15 Pro Max, Sony Alpha 9 II, Google Pixel 8 Pro, OnePlus 12, Samsung Galaxy S24 Ultra \\
1119\_105 & 5 & Apple iPhone 15 Pro Max, Sony Alpha 9 II, Google Pixel 8 Pro, OnePlus 12, Samsung Galaxy S24 Ultra \\
1606\_031 & 6 & Apple iPhone 16 Pro Max, Honor Magic7 Pro, Huawei Pura 70 Ultra, Oppo Find X8 Pro, Vivo X200 Pro, Xiaomi 15 Pro \\
3115\_305 & 5 & Apple iPhone 15 Pro Max, Sony Alpha 9 II, Google Pixel 8 Pro, OnePlus 12, Samsung Galaxy S24 Ultra \\
5203\_106 & 3 & Apple iPhone 16 Pro, Google Pixel 9 Pro XL, Honor Magic7 Pro \\
8005\_103 & 5 & Apple iPhone 16 Pro, Google Pixel 9 Pro XL, Honor Magic7 Pro, Samsung Galaxy S25 Ultra, Xiaomi 15 Ultra \\
Riverside & 3 & Apple iPhone 17 Pro, Apple iPhone Air, Samsung Galaxy S25 Edge \\
\bottomrule
\end{tabular}
\end{adjustbox}
\end{table}

Table~\ref{tab:real_scene_all_attributes} reports scene-wise PLCC/SRCC for overall contrast, eye contrast, and micro-contrast. For each attribute, we compare the proposed zero-shot metric with IFQA fine-tuned on the corresponding mannequin-based annotations. Correlations are computed separately for each scene and macro-averaged across the 10 scenes. The proposed metrics achieve average PLCC/SRCC values of 0.792/0.825 for overall contrast, 0.753/0.768 for eye contrast, and 0.717/0.731 for micro-contrast. For overall contrast, fine-tuned IFQA performs better on average, whereas the proposed metric remains competitive without training on either mannequin or real-portrait annotations. For eye contrast, the proposed metric achieves higher average PLCC and SRCC than fine-tuned IFQA.  Performance varies across scenes for both approaches, reflecting the greater diversity of subjects and capture conditions in the real-portrait dataset.

Table~\ref{tab:real_ifqa_ft_nonft_summary} further examines the effect of mannequin-based fine-tuning on IFQA. Without task-specific adaptation, IFQA yields negative average correlations with all three contrast attributes, indicating that its off-the-shelf quality scores are not directly aligned with the attribute-specific contrast judgments considered in this work. Fine-tuning on the corresponding mannequin-based annotations substantially improves performance across all three attributes. This distinction is important when interpreting the comparison with the proposed metrics: our metrics are applied directly to real portraits without fitting to either mannequin or real-portrait annotations, whereas IFQA benefits from task-specific supervision on the mannequin dataset. In particular, the improvement after fine-tuning, including the strong performance of IFQA for overall contrast, provides additional evidence that the controlled mannequin annotations capture contrast-related information that transfers to real portraits.
Overall, these results support the proposed mannequin dataset as a controlled bridge for studying photographic face contrast, while also showing that the proposed zero-shot metrics generalize meaningfully to real portrait images.

\begin{table*}[t]
\centering
\caption{
Scene-wise PLCC and SRCC on the real-portrait dataset for overall contrast, eye contrast, and micro-contrast. The proposed metrics are applied zero-shot, while IFQA is evaluated after fine-tuning on the corresponding mannequin-based annotations. Values are reported as the point estimate $\pm$ standard deviation over 500 image-level bootstrap resamples. 
}
\label{tab:real_scene_all_attributes}
\scriptsize
\setlength{\tabcolsep}{2.2pt}
\renewcommand{\arraystretch}{1.08}
\resizebox{\textwidth}{!}{%
\begin{tabular}{lcccccccccccc}
\toprule
\multirow{3}{*}{Scene}
& \multicolumn{4}{c}{Overall}
& \multicolumn{4}{c}{Eye}
& \multicolumn{4}{c}{Micro} \\
\cmidrule(lr){2-5}
\cmidrule(lr){6-9}
\cmidrule(lr){10-13}

& \multicolumn{2}{c}{Ours}
& \multicolumn{2}{c}{IFQA (fine-tuned)}
& \multicolumn{2}{c}{Ours}
& \multicolumn{2}{c}{IFQA (fine-tuned)}
& \multicolumn{2}{c}{Ours}
& \multicolumn{2}{c}{IFQA (fine-tuned)} \\
\cmidrule(lr){2-3}
\cmidrule(lr){4-5}
\cmidrule(lr){6-7}
\cmidrule(lr){8-9}
\cmidrule(lr){10-11}
\cmidrule(lr){12-13}

& PLCC & SRCC & PLCC & SRCC
& PLCC & SRCC & PLCC & SRCC
& PLCC & SRCC & PLCC & SRCC \\
\midrule

8005\_103
& 0.914 $\pm$ 0.031 & 0.898 $\pm$ 0.046
& 0.889 $\pm$ 0.030 & 0.894 $\pm$ 0.049
& 0.487 $\pm$ 0.100 & 0.549 $\pm$ 0.125
& -0.228 $\pm$ 0.164 & -0.139 $\pm$ 0.170
& 0.512 $\pm$ 0.105 & 0.532 $\pm$ 0.145
& 0.472 $\pm$ 0.098 & 0.530 $\pm$ 0.133 \\

1606\_031
& 0.632 $\pm$ 0.079 & 0.671 $\pm$ 0.084
& 0.876 $\pm$ 0.033 & 0.868 $\pm$ 0.046
& 0.793 $\pm$ 0.039 & 0.851 $\pm$ 0.045
& 0.791 $\pm$ 0.045 & 0.837 $\pm$ 0.048
& 0.631 $\pm$ 0.095 & 0.736 $\pm$ 0.072
& 0.844 $\pm$ 0.043 & 0.749 $\pm$ 0.082 \\

0105\_089
& 0.835 $\pm$ 0.038 & 0.851 $\pm$ 0.043
& 0.835 $\pm$ 0.038 & 0.853 $\pm$ 0.043
& 0.832 $\pm$ 0.032 & 0.838 $\pm$ 0.038
& 0.748 $\pm$ 0.069 & 0.725 $\pm$ 0.076
& 0.868 $\pm$ 0.030 & 0.870 $\pm$ 0.033
& 0.861 $\pm$ 0.031 & 0.857 $\pm$ 0.043 \\

0109\_2
& 0.689 $\pm$ 0.049 & 0.717 $\pm$ 0.066
& 0.733 $\pm$ 0.071 & 0.713 $\pm$ 0.103
& 0.893 $\pm$ 0.020 & 0.923 $\pm$ 0.026
& 0.919 $\pm$ 0.017 & 0.943 $\pm$ 0.023
& 0.704 $\pm$ 0.070 & 0.670 $\pm$ 0.089
& 0.807 $\pm$ 0.053 & 0.844 $\pm$ 0.059 \\

1119\_105
& 0.800 $\pm$ 0.035 & 0.879 $\pm$ 0.045
& 0.880 $\pm$ 0.040 & 0.857 $\pm$ 0.050
& 0.764 $\pm$ 0.078 & 0.665 $\pm$ 0.100
& 0.956 $\pm$ 0.010 & 0.938 $\pm$ 0.030
& 0.316 $\pm$ 0.116 & 0.244 $\pm$ 0.136
& 0.816 $\pm$ 0.066 & 0.600 $\pm$ 0.103 \\

1113\_510
& 0.624 $\pm$ 0.090 & 0.671 $\pm$ 0.092
& 0.846 $\pm$ 0.032 & 0.863 $\pm$ 0.045
& 0.892 $\pm$ 0.048 & 0.878 $\pm$ 0.054
& 0.892 $\pm$ 0.044 & 0.887 $\pm$ 0.055
& 0.907 $\pm$ 0.019 & 0.959 $\pm$ 0.017
& 0.797 $\pm$ 0.034 & 0.843 $\pm$ 0.057 \\

3115\_305
& 0.880 $\pm$ 0.036 & 0.897 $\pm$ 0.038
& 0.949 $\pm$ 0.012 & 0.948 $\pm$ 0.020
& 0.817 $\pm$ 0.026 & 0.880 $\pm$ 0.031
& 0.693 $\pm$ 0.073 & 0.623 $\pm$ 0.103
& 0.930 $\pm$ 0.017 & 0.943 $\pm$ 0.028
& 0.772 $\pm$ 0.057 & 0.823 $\pm$ 0.061 \\

Riverside
& 0.929 $\pm$ 0.030 & 0.984 $\pm$ 0.035
& 0.957 $\pm$ 0.019 & 0.956 $\pm$ 0.063
& 0.379 $\pm$ 0.197 & 0.363 $\pm$ 0.274
& 0.688 $\pm$ 0.164 & 0.703 $\pm$ 0.235
& 0.963 $\pm$ 0.024 & 0.940 $\pm$ 0.069
& 0.541 $\pm$ 0.186 & 0.544 $\pm$ 0.194 \\

0109\_3
& 0.877 $\pm$ 0.026 & 0.898 $\pm$ 0.038
& 0.917 $\pm$ 0.029 & 0.898 $\pm$ 0.040
& 0.808 $\pm$ 0.039 & 0.885 $\pm$ 0.036
& 0.810 $\pm$ 0.044 & 0.745 $\pm$ 0.079
& 0.579 $\pm$ 0.115 & 0.566 $\pm$ 0.132
& 0.867 $\pm$ 0.032 & 0.837 $\pm$ 0.053 \\

5203\_106
& 0.744 $\pm$ 0.052 & 0.782 $\pm$ 0.064
& 0.695 $\pm$ 0.065 & 0.773 $\pm$ 0.065
& 0.863 $\pm$ 0.035 & 0.853 $\pm$ 0.051
& 0.059 $\pm$ 0.170 & 0.031 $\pm$ 0.157
& 0.762 $\pm$ 0.059 & 0.851 $\pm$ 0.063
& 0.156 $\pm$ 0.114 & 0.147 $\pm$ 0.142 \\

\midrule
\textbf{Average}
& \textbf{0.792 $\pm$ 0.016} & \textbf{0.825 $\pm$ 0.017}
& \textbf{0.858 $\pm$ 0.013} & \textbf{0.862 $\pm$ 0.017}
& \textbf{0.753 $\pm$ 0.025} & \textbf{0.768 $\pm$ 0.034}
& \textbf{0.633 $\pm$ 0.032} & \textbf{0.629 $\pm$ 0.037}
& \textbf{0.717 $\pm$ 0.024} & \textbf{0.731 $\pm$ 0.029}
& \textbf{0.693 $\pm$ 0.027} & \textbf{0.678 $\pm$ 0.032} \\
\bottomrule
\end{tabular}%
}
\end{table*}

\begin{table}[htbp]
\centering
\caption{Macro-average PLCC and SRCC across real-portrait scenes for IFQA before and after fine-tuning on the mannequin-based annotations.}
\label{tab:real_ifqa_ft_nonft_summary}
\small
\begin{tabular}{lcc|cc}
\hline
\multirow{2}{*}{Target}
& \multicolumn{2}{c|}{IFQA (fine-tuned)}
& \multicolumn{2}{c}{IFQA (non-fine-tuned)} \\
\cline{2-5}
& PLCC & SRCC & PLCC & SRCC \\
\hline
Overall contrast & 0.858 & 0.862 & -0.075 & -0.086 \\
Eye contrast     & 0.633 & 0.629 & -0.117 & -0.106 \\
Micro-contrast   & 0.693 & 0.678 & -0.113 & -0.105 \\
\hline
\end{tabular}
\end{table}

\section{Controlled Mannequin Dataset}
The original full-resolution dataset contains 508 images in total across the two mannequins, with 250 images for Diana and 258 images for Eugene. To construct the comparison groups used for perceptual annotation, we supplemented the original captures with edited images so that each group contained a sufficient number of samples for pairwise comparison. The resulting composition of original and edited images for each group is summarized in Table~\ref{tab:group_orig_edit_counts}.

\subsection{Face Cropping and Region Mask Extraction}

We first detect the face region in each image to ensure that contrast analysis focuses solely on the face rather than the background.
For the face detection, we used the MTCNN implementation~\cite{zhang2016joint} provided by the \texttt{facenet-pytorch} library~\cite{facenet_pytorch_github}. Each crop is resized to a fixed resolution of $585 \times 820$  pixels. 
This step reduces variation in face scale and alignment and standardizes the image presentation for pairwise comparison experiments.
We then extract semantic facial regions using a BiSeNet face parsing model~\cite{yu2018bisenet,yu2021bisenet}, from which we derive three masks used throughout this work: a face mask for overall contrast, an eye mask for eye contrast, and a skin mask for micro-contrast as shown in Figure~\ref{fig:guidelines}.

\begin{table}[htbp]
\centering
\caption{Image composition by group and mannequin (original vs.\ edited).}
\label{tab:group_orig_edit_counts}
\resizebox{\linewidth}{!}{%
\begin{tabular}{lcccccc}
\hline
& \multicolumn{3}{c}{Diana} & \multicolumn{3}{c}{Eugene} \\
\cline{2-4} \cline{5-7}
Group & Original & Edited & Total & Original & Edited & Total \\
\hline
Group~1 & 38 & 37 & 75  & 42 & 33 & 75  \\
Group~2 & 47 & 28 & 75  & 49 & 26 & 75  \\
Group~3 & 45 & 30 & 75  & 41 & 34 & 75  \\
Group~4 & 59 & 16 & 75  & 62 & 13 & 75  \\
Group~5 & 61 & 39 & 100 & 64 & 36 & 100 \\
\hline
Total & 250 & 150 & 400 & 258 & 142 & 400 \\
\hline
\end{tabular}%
}
\end{table}

\subsection{Mannequin Dataset Composition Evaluation.}

The complete dataset contains images captured with 29 smartphones from multiple manufacturers, covering a broad range of device categories and ISP characteristics. Table~\ref{tab:device_brands_models} summarizes the device brands and models included in the dataset. The dataset contains two mannequin subjects, Eugene and Diana, and the processed comparison subsets are organized into five groups corresponding to different lighting regimes, as described in the main paper.

We evaluate the data under two complementary conditions. \emph{Within-lighting condition} corresponds to a within-group device-disjoint split on Groups~1--4, while \emph{Mixed-lighting condition} is a cross-group evaluation in which Groups~1--4 are used as the source set and Group~5 is used as the target set. The purpose of \text{Within-lighting condition} is to measure generalization across unseen devices within the same lighting regime, whereas \emph{Mixed-lighting condition} evaluates transfer to a broader mixed group with more diverse contrast variation.

\subsubsection{\emph{Within-lighting condition.}}
This protocol is applied only to Groups~1--4. For each mannequin and each group, the split is performed by device rather than by image, so that the same smartphone never appears in both training and validation. This provides a strict device-disjoint evaluation setting and avoids optimistic correlations caused by device-specific ISP signatures appearing in both subsets.
Among the remaining samples, the validation devices are selected to satisfy three criteria: (i) strict device separation, (ii) a validation size close to the target ratio, and (iii) a reasonable balance between the training and validation subsets in JOD space. The balancing criterion uses the mean and standard deviation of the four annotation dimensions (Preference, Overall, Eye, and Micro) as a secondary constraint when constructing the split. This procedure yields a device-disjoint partition while keeping the train and validation subsets reasonably matched in perceptual-score distribution.
\begin{table*}[t]
\centering
\small
\caption{Device brands and models included in the mannequin dataset.}
\label{tab:device_brands_models}
\begin{tabular}{lcl}
\hline
Brand & \# Devices & Models \\
\hline
Samsung & 8 & A55 5G, A56 5G, S24 Ultra, S25 Ultra, Z Flip5, Z Flip6, Z Fold5, Z Fold6 \\
Google  & 7 & Pixel 8, 8 Pro, 8a, 9, 9 Pro Fold, 9 Pro XL, 9a \\
Xiaomi  & 6 & 13T Pro, 14, 14 Ultra, 15 Ultra, Redmi Note 13 Pro+, Redmi Note 14 Pro+ \\
Huawei  & 2 & Mate 60 Pro+, Pura 70 Ultra \\
Vivo    & 2 & X100 Pro, X100 Ultra \\
Honor   & 2 & Magic 6 Pro, Magic 7 Pro \\
Oppo    & 1 & Find X7 Ultra \\
Asus    & 1 & Zenfone 11 Ultra \\
\hline
\end{tabular}
\end{table*}

\begin{table*}[t]
\centering
\small
\caption{Device sets used for \emph{Within-lighting condition} on Groups~1--4. The split is strictly device-disjoint.}
\label{tab:device_split_summary}
\begin{tabular}{lp{11cm}}
\hline
Training devices (19) &
AsusZenfone11Ultra, GooglePixel8, GooglePixel9, GooglePixel9ProFold, GooglePixel9ProXL, GooglePixel9a, HuaweiPura70Ultra, SamsungGalaxyA555G, SamsungGalaxyA565G, SamsungGalaxyS25Ultra, SamsungGalaxyZFlip6, SamsungGalaxyZFold5, SamsungGalaxyZFold6, VivoX100Pro, VivoX100Ultra, Xiaomi14Ultra, Xiaomi15Ultra, XiaomiRedmiNote13ProPlus5G, XiaomiRedmiNote14ProPlus5G \\
\hline
Validation devices (10) &
GooglePixel8Pro, GooglePixel8a, HonorMagic6Pro, HonorMagic7Pro, HuaweiMate60ProPlus, OppoFindX7Ultra, SamsungGalaxyS24Ultra, SamsungGalaxyZFlip5, Xiaomi13TPro, Xiaomi14 \\
\hline
\end{tabular}
\end{table*}

Table~\ref{tab:split_device_summary_group_person} reports the number of effective training and validation devices for each group and mannequin under \emph{Within-lighting condition}. As expected, the exact number of available devices varies slightly across groups because some captures are removed during duplicate filtering, but the split remains strictly device-disjoint in all cases. Table~\ref{tab:split_summary_person} further summarizes the effective number of samples per mannequin after duplicate removal. The resulting split remains close to a two-thirds / one-third partition, with no device overlap between the training and validation sets.

\begin{table*}[t]
\centering
\small
\caption{Device split summary by group and mannequin under \emph{Within-lighting condition}.}
\label{tab:split_device_summary_group_person}
\begin{tabular}{lllll}
\hline
Group & Person & Train Dev & Val Dev & Overlap Dev \\
\hline
Group1 & Diana & 18 & 10 & 0 \\
Group1 & Eugene & 18 & 10 & 0 \\
Group2 & Diana & 16 & 10 & 0 \\
Group2 & Eugene & 17 & 10 & 0 \\
Group3 & Diana & 17 & 10 & 0 \\
Group3 & Eugene & 17 & 10 & 0 \\
Group4 & Diana & 18 & 10 & 0 \\
Group4 & Eugene & 18 & 10 & 0 \\
\hline
\end{tabular}
\end{table*}

\begin{table*}[t]
\centering
\small
\caption{Split summary aggregated by mannequin under \emph{Within-lighting condition}.}
\label{tab:split_summary_person}
\resizebox{\columnwidth}{!}{
\begin{tabular}{llllll}
\hline
Person & Total & Train Total & Val Total & Train Ratio & Val Ratio  \\
\hline
Diana & 300 & 195 & 105 & 0.6500 & 0.3500 \\
Eugene & 300 & 200 & 100 & 0.6667 & 0.3333  \\
\hline
\end{tabular}
}
\end{table*}

\subsubsection{Mixed-lighting condition.}

\emph{Mixed-lighting condition} uses Groups~1--4 as the source set and Group~5 as the target set. As Group~5 is constructed by mixing samples from the other lighting groups, any duplicated samples from Groups~1--4 are excluded from the training set and removed from the released evaluation subset to avoid leakage across protocols. Group 5 provides a broader and more heterogeneous evaluation setting than \emph{Within-lighting condition}. In this protocol, no additional train/validation split is performed within Group~5; instead, the goal is to measure how well a method transfers from the structured lighting conditions represented by Groups~1--4 to the more diverse contrast distribution represented by Group~5.

\subsection{Annotation Consistency Analysis}

\begin{figure}[t]
    \centering
    \includegraphics[width=\columnwidth]{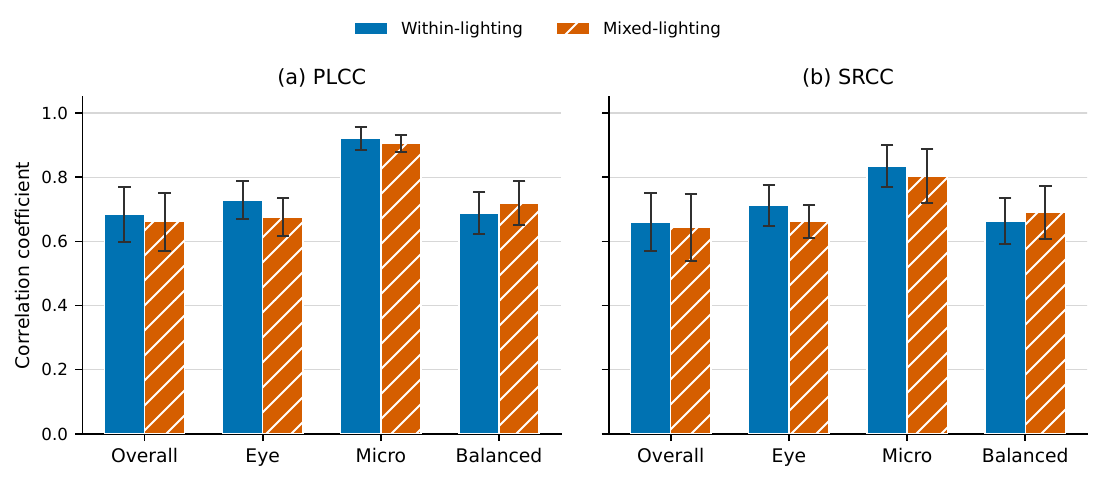}
    \caption{
    Annotation stability of the JOD scales under within-lighting
    and mixed-lighting evaluation. Each observer is excluded in turn,
    and the JOD scale reconstructed from the remaining observers is
    compared with the reference scale estimated from all annotations.
    Bars show the mean PLCC and SRCC across the four observer exclusions,
    and error bars indicate the corresponding standard deviation.
    }
    \label{fig:loo_stability}
\end{figure}

We conducted two complementary analyses to assess the consistency of the collected annotations. First, we computed binary pairwise agreement as a direct measure on the raw pairwise decisions. The component-level tasks showed moderate agreement: eye contrast achieved $0.666 \pm 0.101$, micro-contrast $0.726 \pm 0.092$, and overall contrast $0.678 \pm 0.106$. Micro-contrast produced the highest agreement, consistent with the relatively direct judgment of fine-detail visibility. Balanced contrast showed lower agreement $0.588 \pm 0.124$, reflecting the more subjective nature of this holistic task. 
Second, we also evaluated the stability of the recovered perceptual scales using a leave-one-observer-out analysis. For each task, we repeatedly removed one observer, recomputed the JOD scale from the remaining annotations, and measured its correlation with the JOD scale obtained from the full annotation set. The results are shown in Figure~\ref{fig:loo_stability}. The resulting PLCC and SRCC values show that the inferred scales remain correlated after observer removal, especially for micro-contrast, which achieves the highest correlations under both evaluation settings. This indicates that the recovered JOD scales retain meaningful stability when individual observers are
excluded and can serve as perceptual references for metric validation.

\subsection{Device-Level Contrast Analysis}

\begin{figure*}[p]
    \centering
    \captionsetup{font=small}
    \setlength{\abovecaptionskip}{4pt}
    \setlength{\belowcaptionskip}{0pt}

    \includegraphics[width=0.98\textwidth,height=0.24\textheight,keepaspectratio]{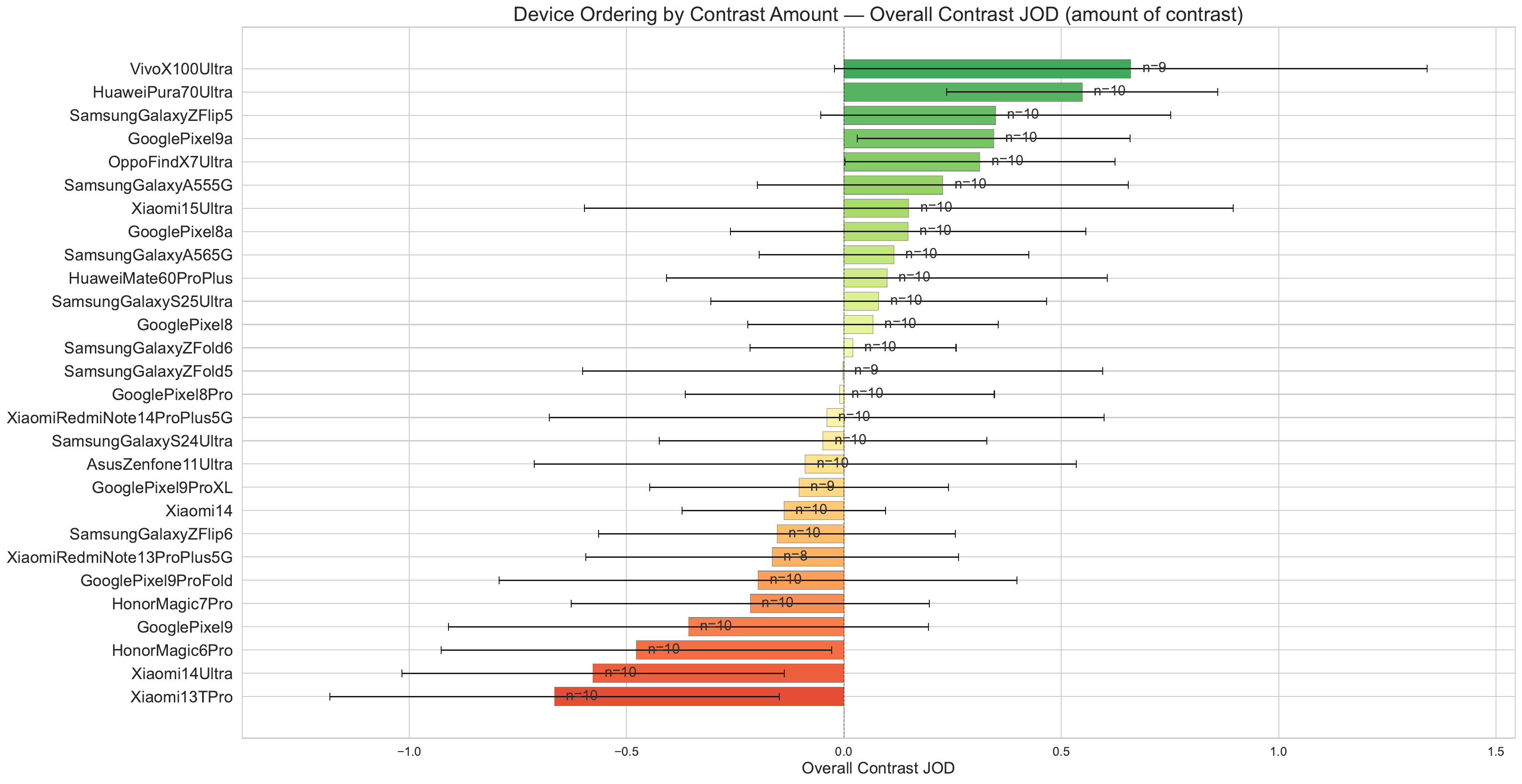}\\[-1mm]
    {\scriptsize (a) Overall Contrast JOD ranking}\\[0.3mm]

    \includegraphics[width=0.98\textwidth,height=0.24\textheight,keepaspectratio]{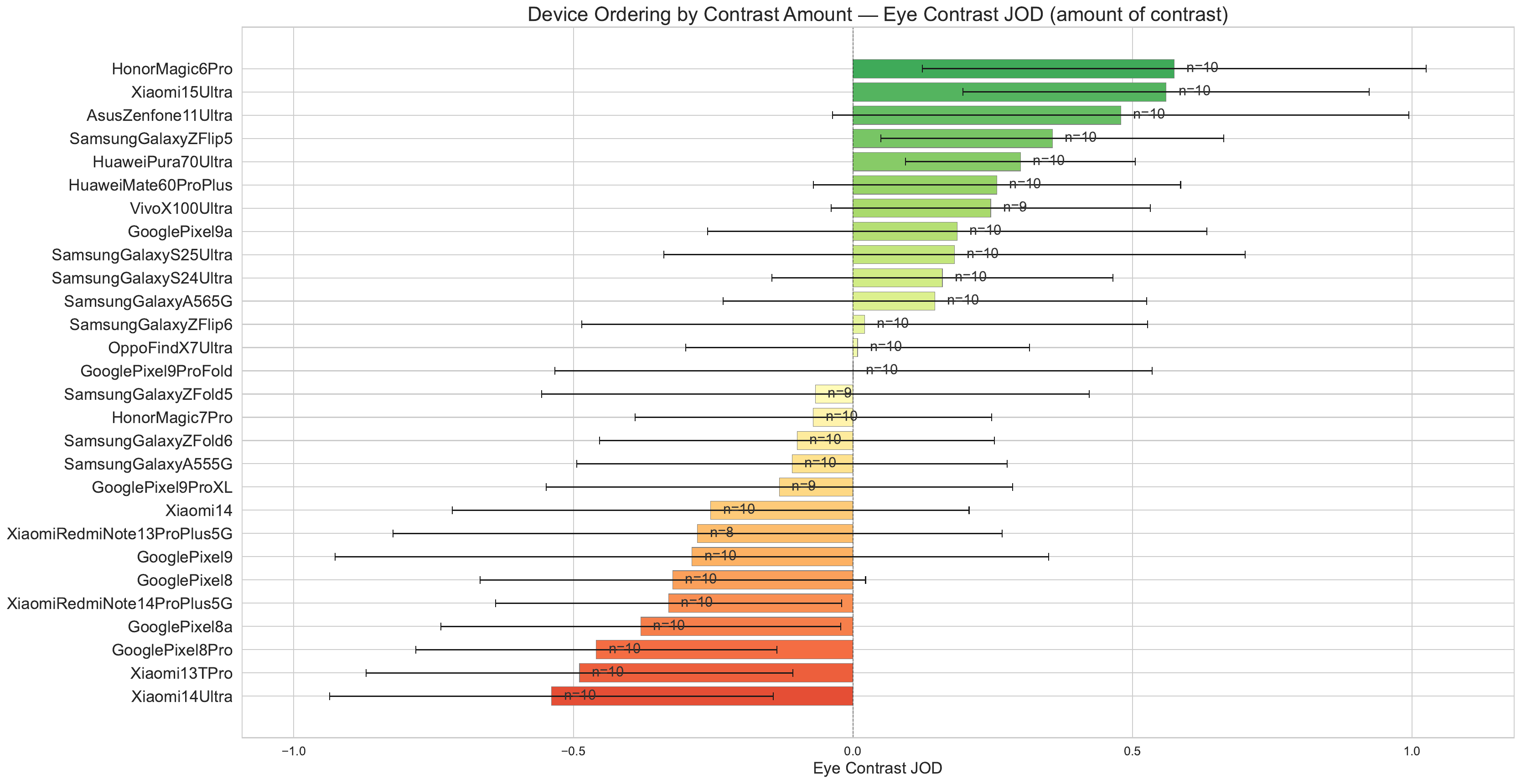}\\[-1mm]
    {\scriptsize (b) Eye Contrast JOD ranking}\\[0.3mm]

    \includegraphics[width=0.98\textwidth,height=0.24\textheight,keepaspectratio]{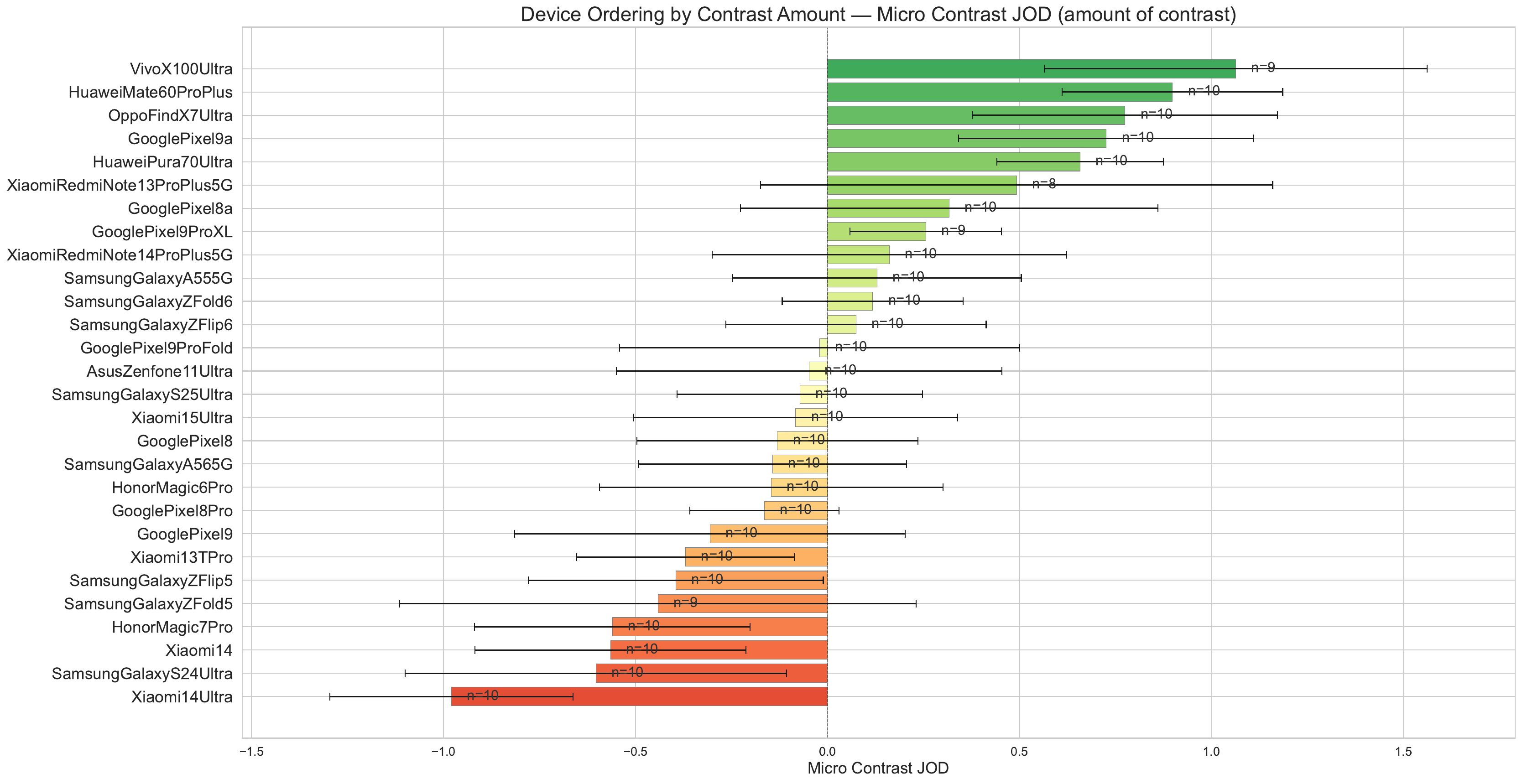}\\[-1mm]
    {\scriptsize (c) Micro Contrast JOD ranking}\\[0.3mm]

    \includegraphics[width=0.98\textwidth,height=0.24\textheight,keepaspectratio]{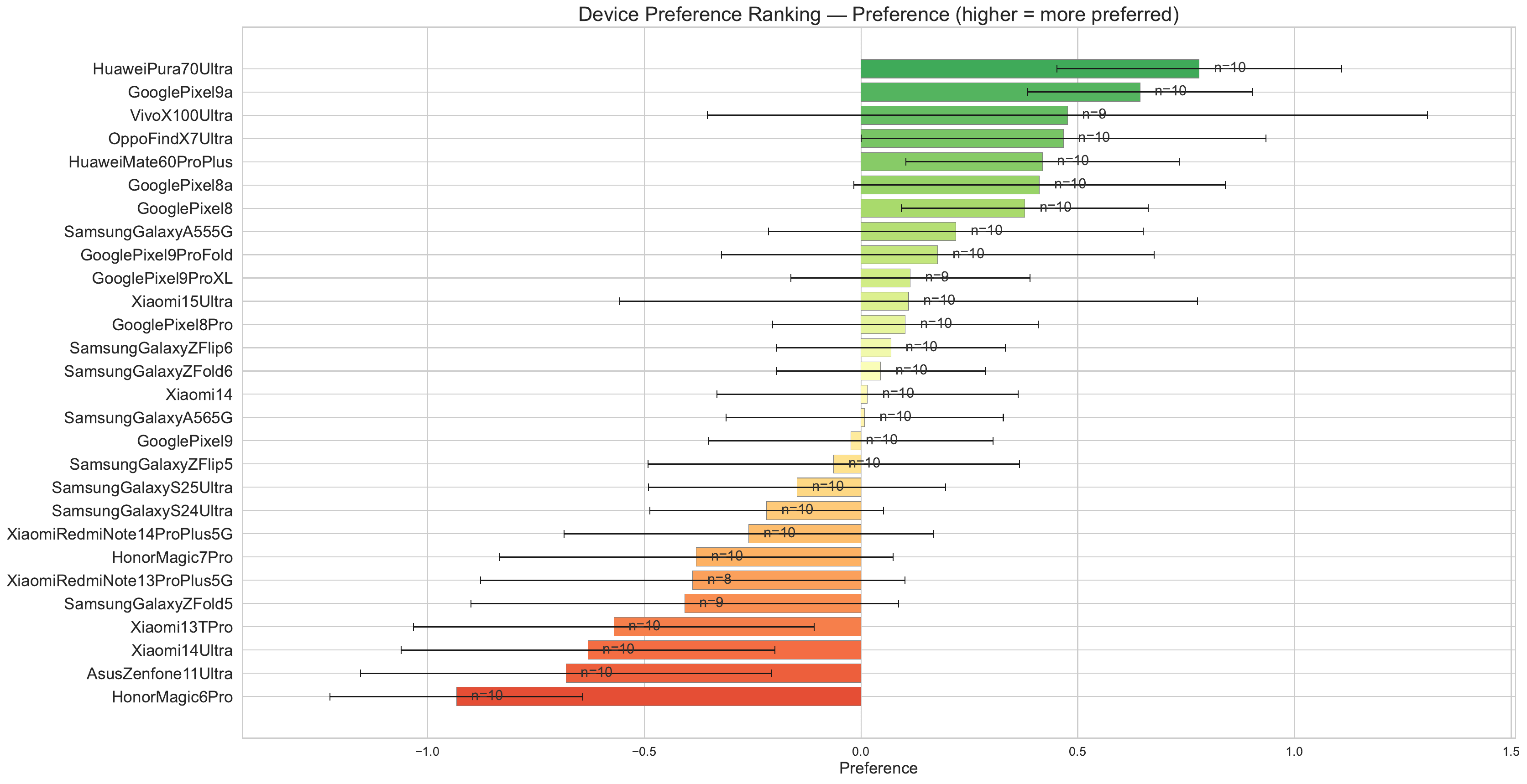}\\[-1mm]
    {\scriptsize (d) Balanced Contrast JOD ranking}

    \caption{Device-level pooled rankings with 95\% confidence intervals for four JOD targets. Overall, Eye, and Micro represent amount of contrast dimensions, while Balanced contrast reflects subjective utility. Scores are macro-aggregated subgroup-normalized JOD values expressed in standard-deviation units.}
    \label{fig:device_ranking_jod_vertical}
\end{figure*}

This analysis illustrates how the proposed annotations can be used to compare smartphone rendering behavior at the device level.
Figure~\ref{fig:device_ranking_jod_vertical} shows the device-level contrast analysis.  
To remove cross-subgroup scale differences, each JOD target was normalized within each \((\text{group},\text{mannequin})\) subgroup. For sample \(i\) in subgroup \(c=(g,p)\), we computed

\begin{equation}
z_i^{(m)}=\frac{x_i^{(m)}-\mu_c^{(m)}}{\sigma_c^{(m)}},
\end{equation}

where \(x_i^{(m)}\) is the raw JOD value for metric \(m\), and \(\mu_c^{(m)}\), \(\sigma_c^{(m)}\) are the subgroup-wise mean and standard deviation.

To avoid bias from non-uniform sample counts across groups and mannequins, we used \emph{macro} aggregation.  
For each device \(d\), metric \(m\), and subgroup \(c\), we first computed a subgroup-level device mean

\begin{equation}
\mu_{d,c}^{(m)}=\frac{1}{N_{d,c}}\sum_{i\in\mathcal{I}_{d,c}} z_i^{(m)},
\end{equation}

where \(\mathcal{I}_{d,c}\) is the set of samples captured by device \(d\) in subgroup \(c\), and \(N_{d,c}=|\mathcal{I}_{d,c}|\).  
Then we averaged these subgroup means with equal weight:
\begin{equation}
\bar z_d^{(m)}=\frac{1}{|C_d|}\sum_{c\in C_d}\mu_{d,c}^{(m)},
\end{equation}
where \(C_d\) is the set of available \((\text{group},\text{mannequin})\) subgroups for device \(d\). Therefore, each bar in the ranking plots represents the macro-aggregated device mean of normalized JOD scores. Error bars indicate the \(95\%\) confidence interval estimated across the available subgroup-level means for each device.

\begin{figure*}[t]
    \centering
    \begin{minipage}{0.7\textwidth}
        \centering
        \includegraphics[width=\linewidth]{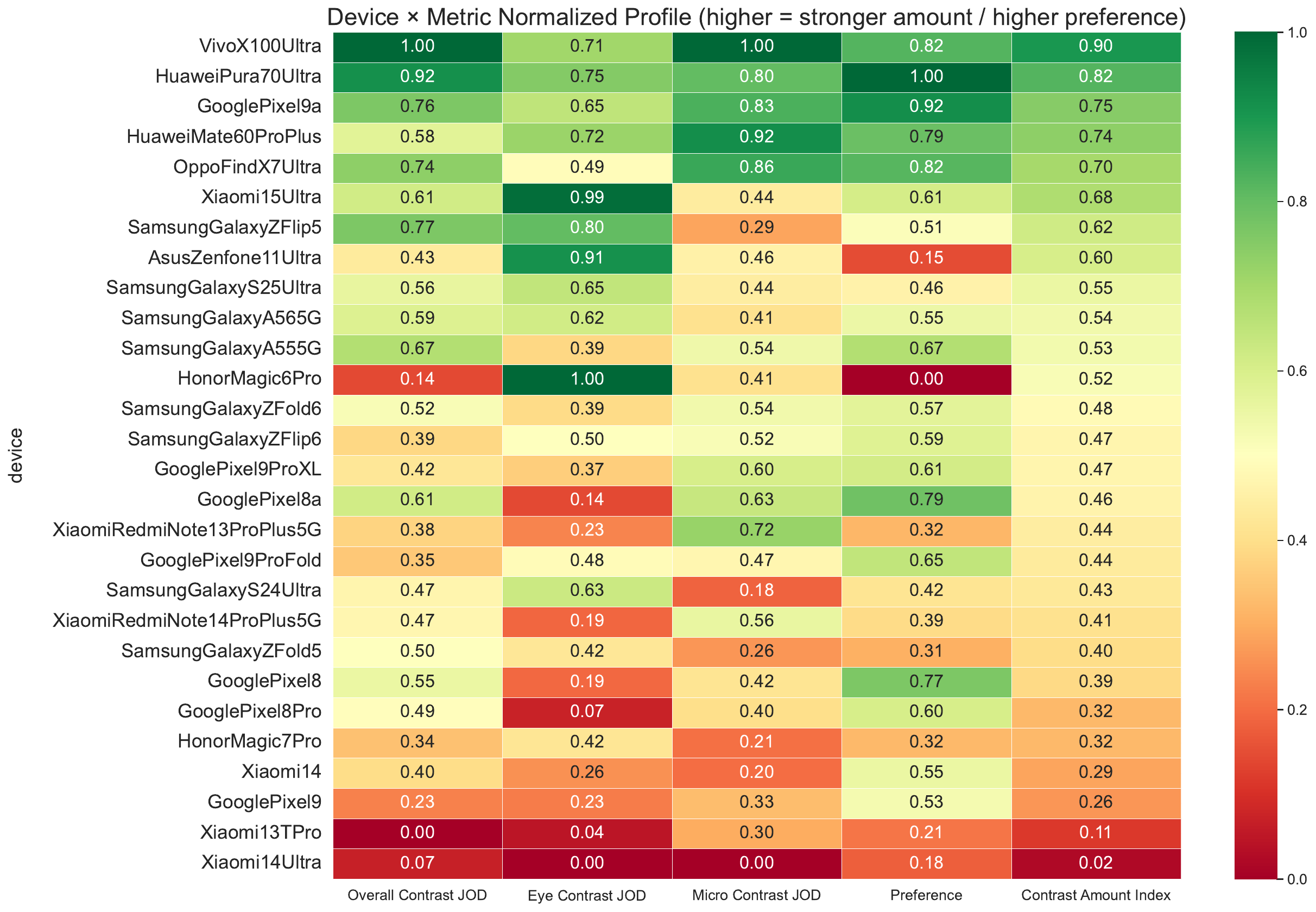}
        {\small (a) Device scorecard across Overall/Eye/Micro/Preference (balanced contrast).}
    \end{minipage}
    \hfill
    \begin{minipage}{0.8\textwidth}
        \centering
        \includegraphics[width=\linewidth]{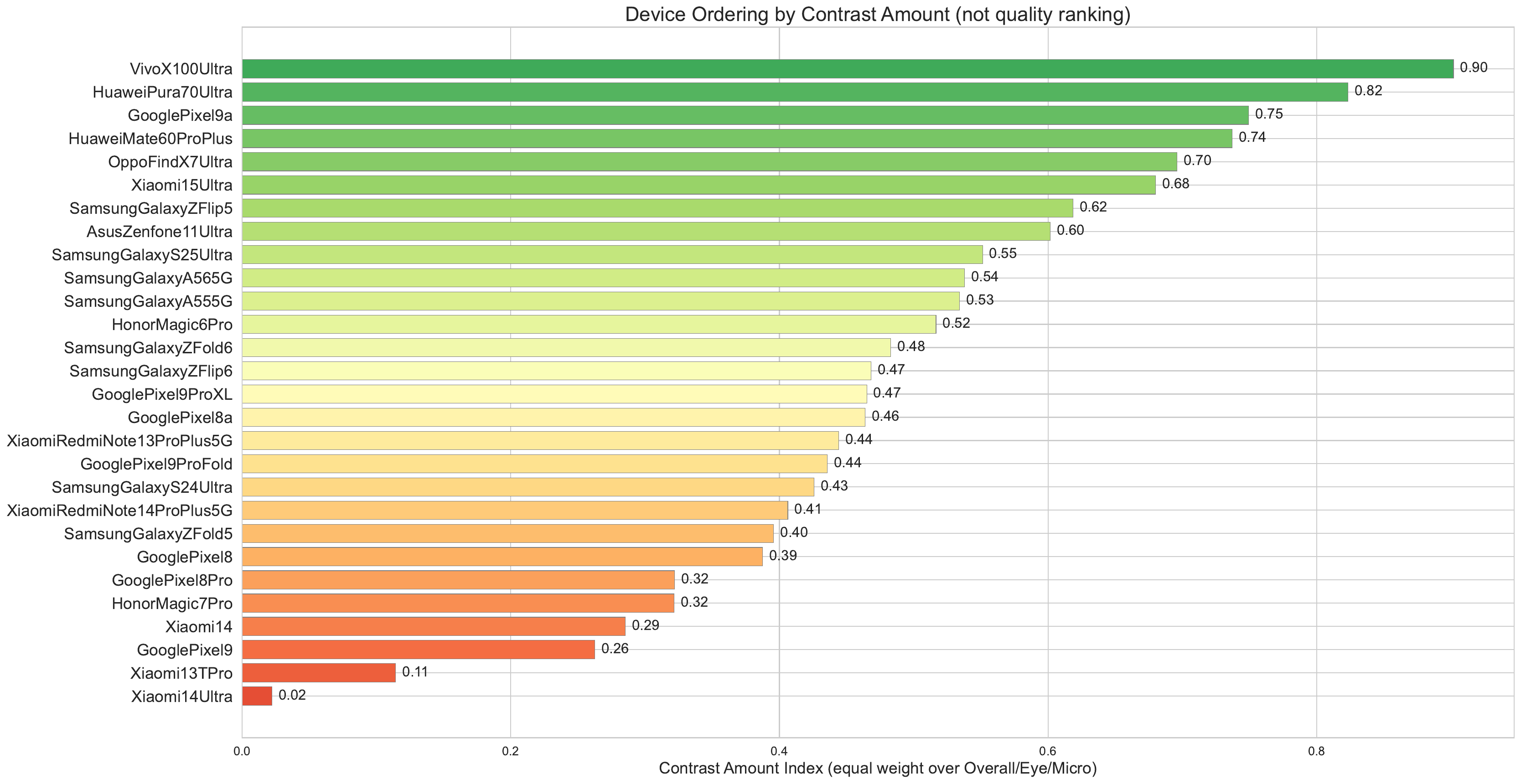}
        {\small (b) Device ordering by contrast-amount index \(C_d\) (equal-weight average of Overall/Eye/Micro).}
    \end{minipage}
    \caption{Device-level summary under z-score-based analysis. Panel (a) provides a compact cross-metric profile per device, and panel (b) highlights relative ordering by contrast amount.}
    \label{fig:scorecard_amount}
\end{figure*}

\paragraph{Device score and Contrast Amount Ordering}
Using the macro-aggregated score \(\bar z_d^{(m)}\), we then applied metric-wise min--max scaling across devices for visualization:
\begin{equation}
\tilde s_d^{(m)}=
\frac{\bar z_d^{(m)}-\min_{d'} \bar z_{d'}^{(m)}}
{\max_{d'} \bar z_{d'}^{(m)}-\min_{d'} \bar z_{d'}^{(m)}} \in [0,1].
\end{equation}
Figure~\ref{fig:scorecard_amount} (a) shows this normalized profile for Overall, Eye, Micro, and 
Balanced contrast. We define the contrast-amount index as the equal-weight average of the three amount dimensions:
\begin{equation}
C_d=\frac{1}{3}\Big(\tilde s_d^{(\mathrm{Overall})}
+\tilde s_d^{(\mathrm{Eye})}
+\tilde s_d^{(\mathrm{Micro})}\Big).
\end{equation}
Figure~\ref{fig:scorecard_amount} (b) ranks devices by \(C_d\). This index captures \emph{amount of contrast} rather than perceptual quality; therefore, a higher \(C_d\) does not necessarily imply a higher Preference score. We treat balanced contrast Preference as the primary utility endpoint, and \(C_d\) as an explanatory descriptor of physical contrast strength.

\section{Pairwise Comparison Tool}
\begin{figure*}[t]
    \centering
    \includegraphics[width=0.99\textwidth]{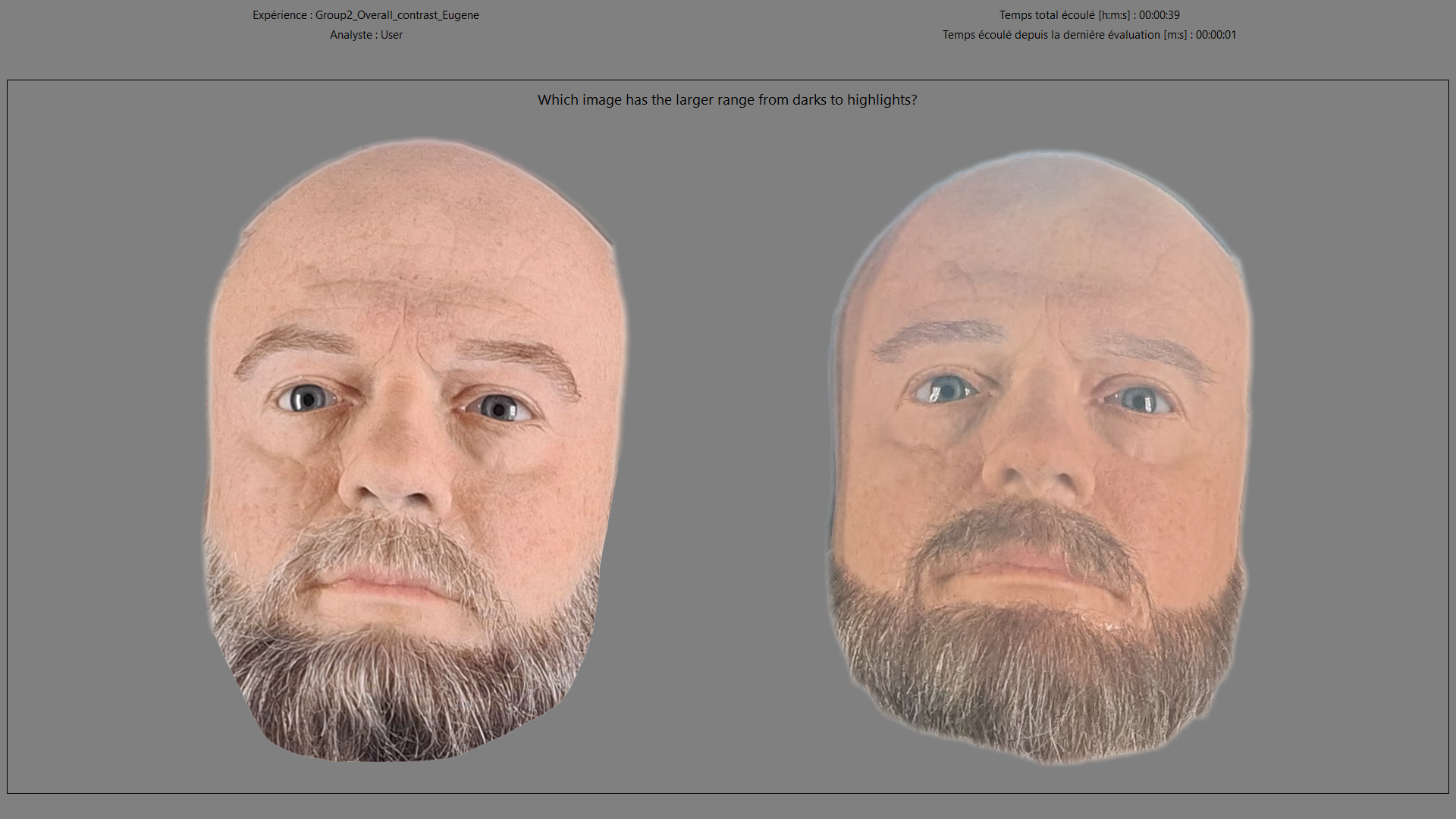}
    \caption{Example of the annotation tool used to acquire pairwise comparisons (Overall contrast, Eugene)}
    \label{fig:Example_overall_Eugene}
\end{figure*}

\begin{figure*}[t]
    \centering
    \includegraphics[width=0.99\textwidth]{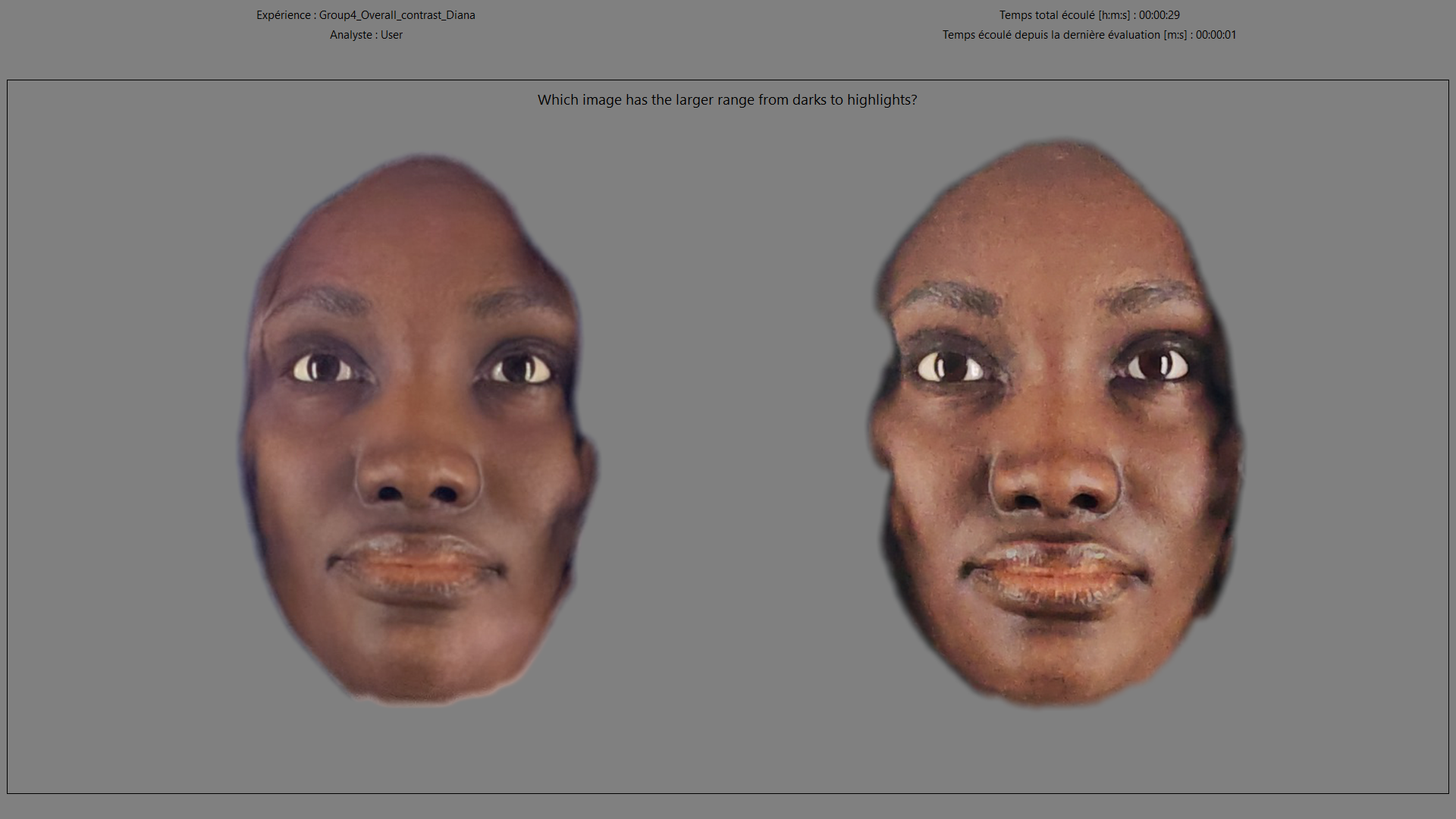}
    \caption{Example of the annotation tool used to acquire pairwise comparisons (Overall contrast, Diana)}
    \label{fig:Example_overall_Diana}
\end{figure*}

\begin{figure*}[t]
    \centering    \includegraphics[width=0.99\textwidth]{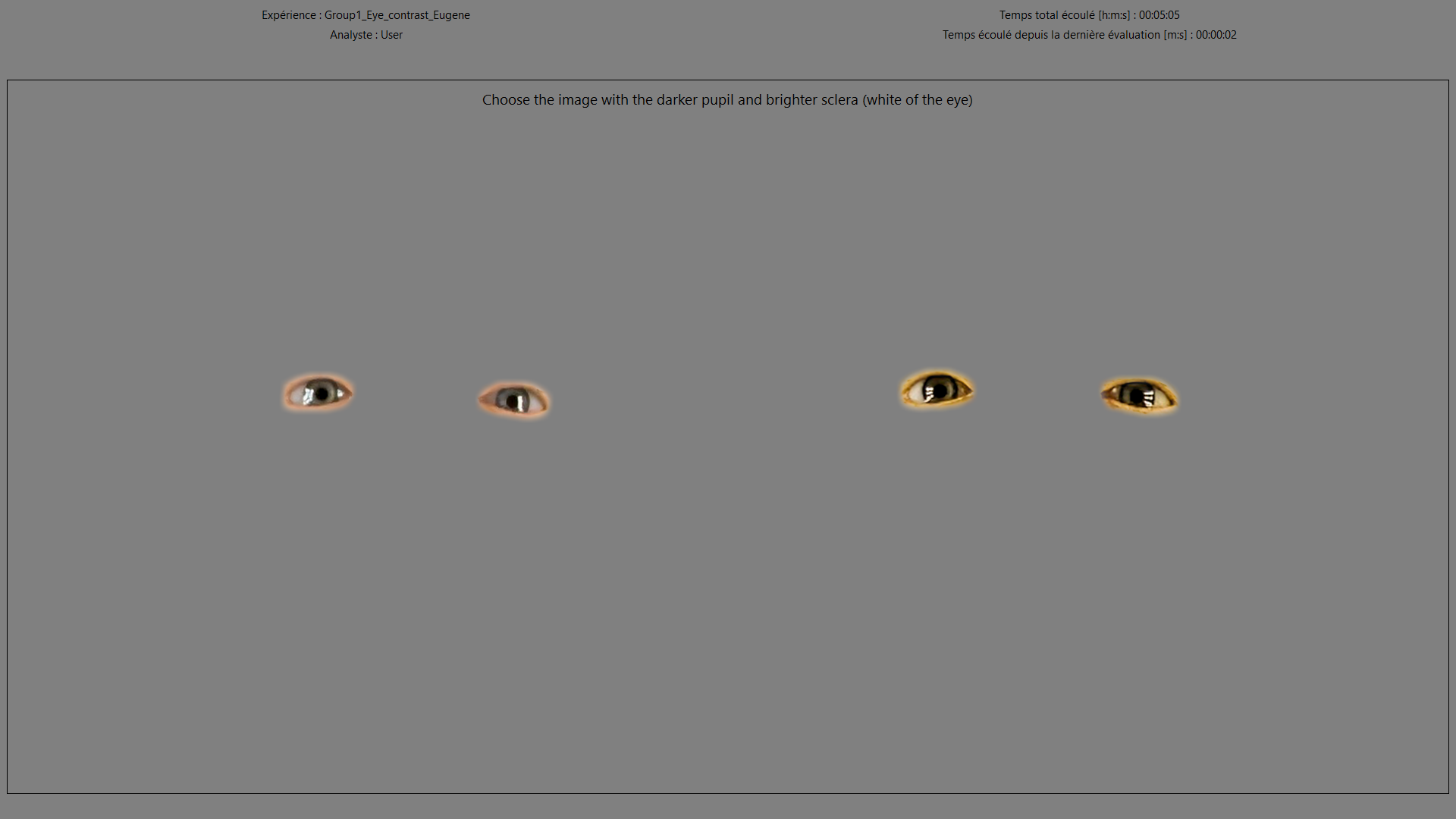}
    \caption{Example of the annotation tool used to acquire pairwise comparisons (Eye contrast, Eugene)}
    \label{fig:Example_Eye_Eugene}
\end{figure*}

\begin{figure*}[t]
    \centering    \includegraphics[width=0.99\textwidth]{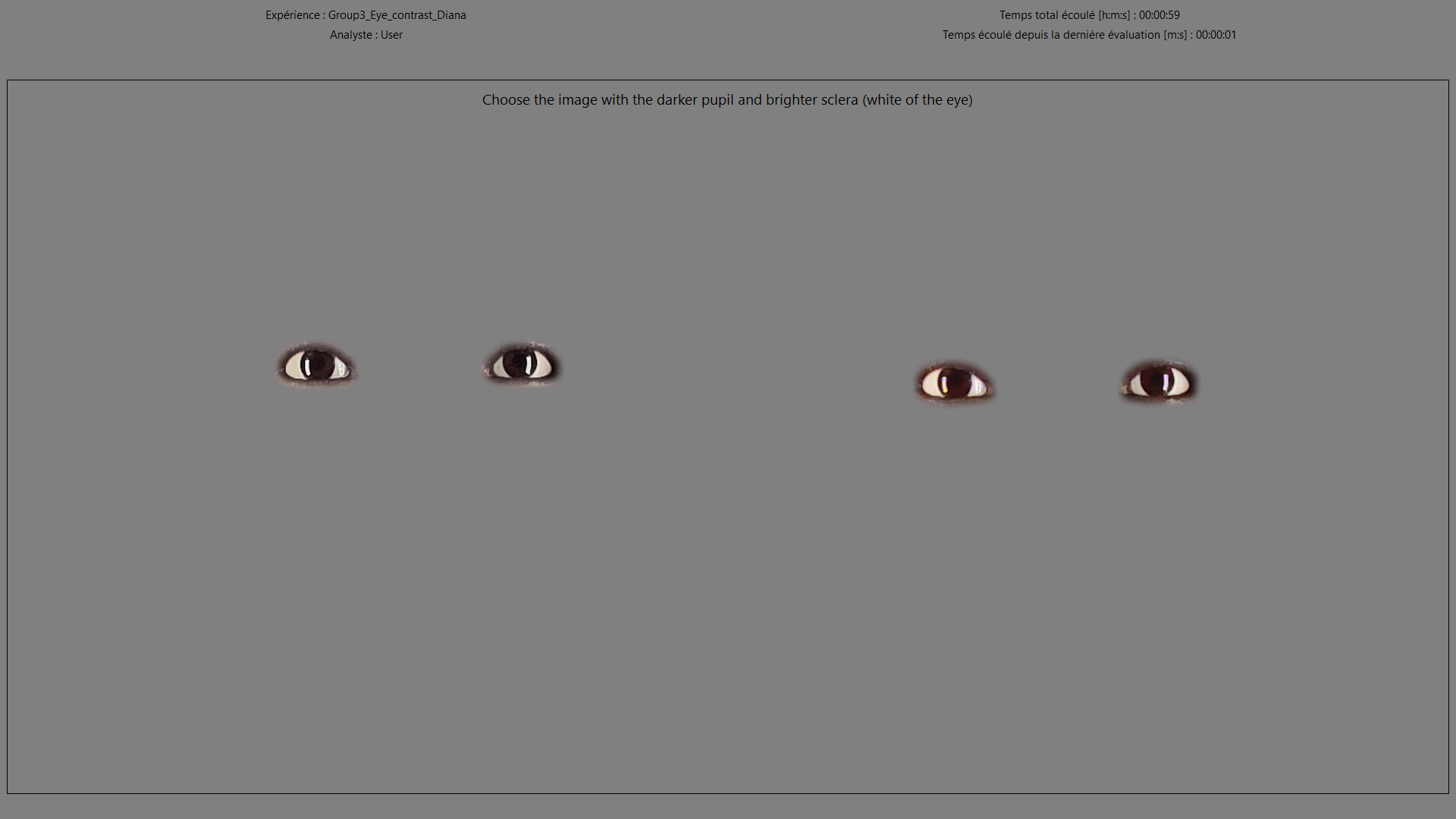}
    \caption{Example of the annotation tool used to acquire pairwise comparisons (Eye contrast, Diana)}
    \label{fig:Example_Eye_Diana}
\end{figure*}

\begin{figure*}[t]
    \centering    \includegraphics[width=0.99\textwidth]{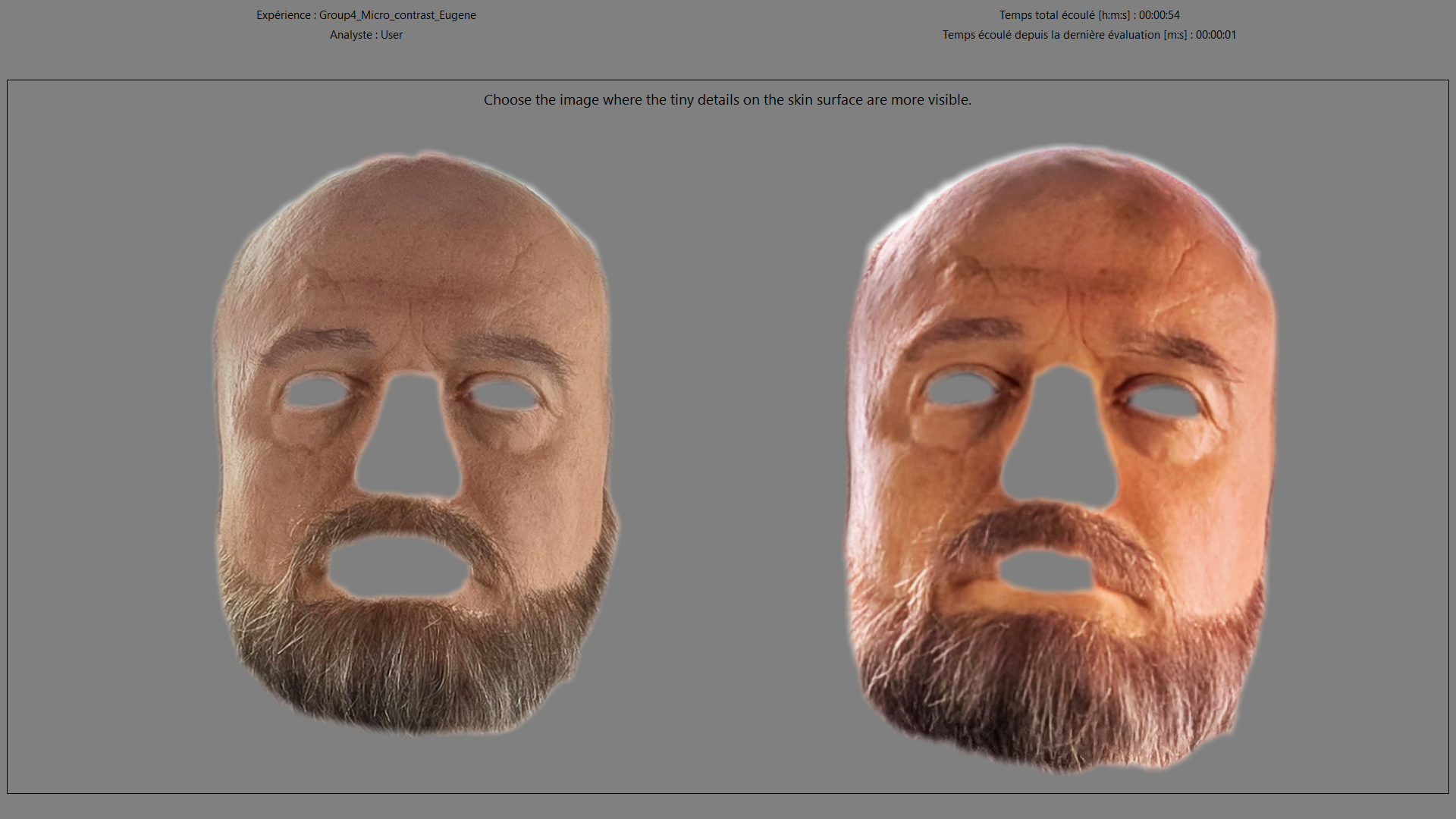}
    \caption{Example of the annotation tool used to acquire pairwise comparisons (Micro-contrast, Eugene)}
    \label{fig:Example_Micro_Eugene}
\end{figure*}

\begin{figure*}[t]
    \centering    \includegraphics[width=0.99\textwidth]{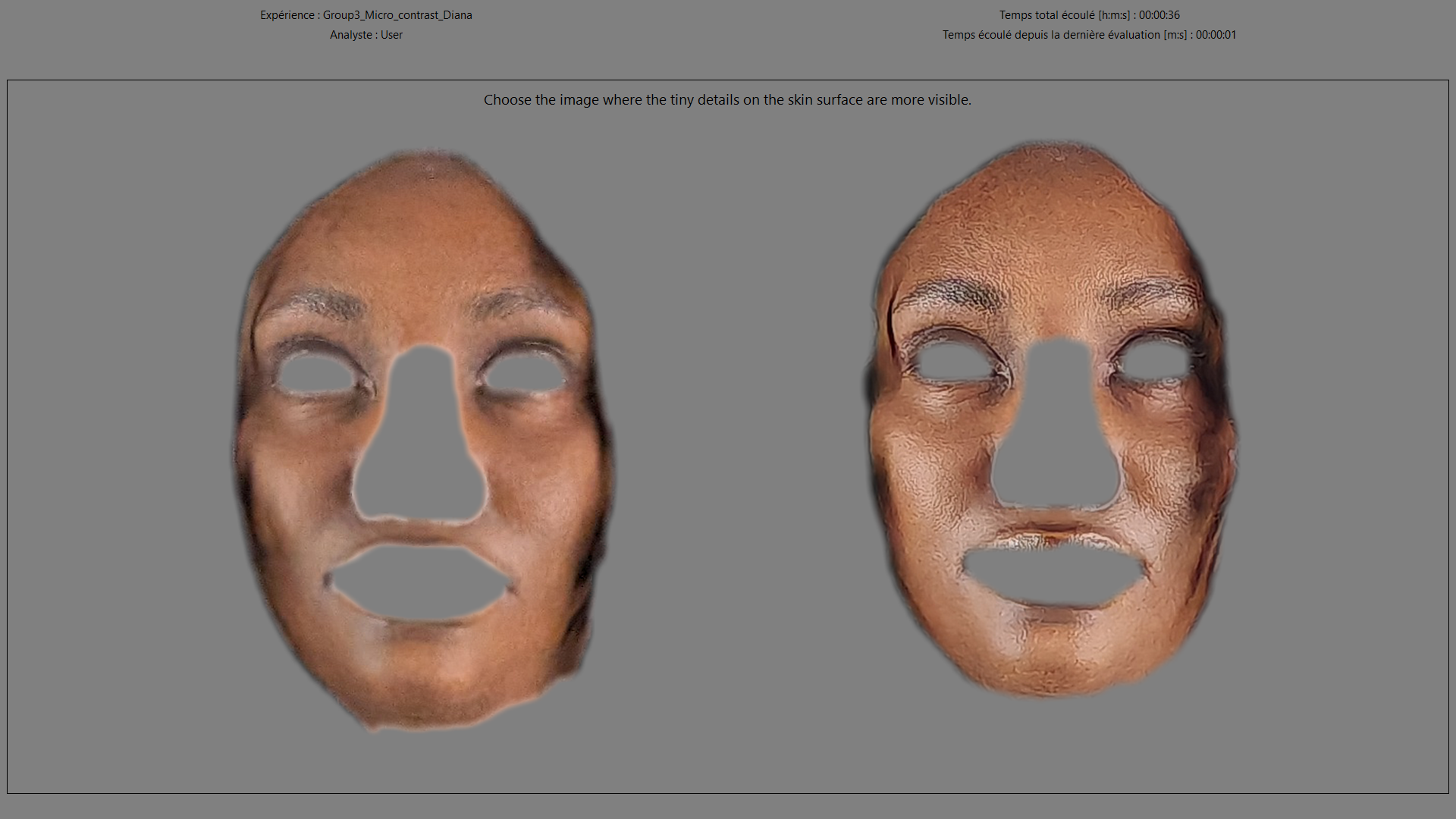}
    \caption{Example of the annotation tool used to acquire pairwise comparisons (Micro-contrast, Diana)}
    \label{fig:Example_Micro_Diana}
\end{figure*}

\begin{figure*}[t]
    \centering    \includegraphics[width=0.99\textwidth]{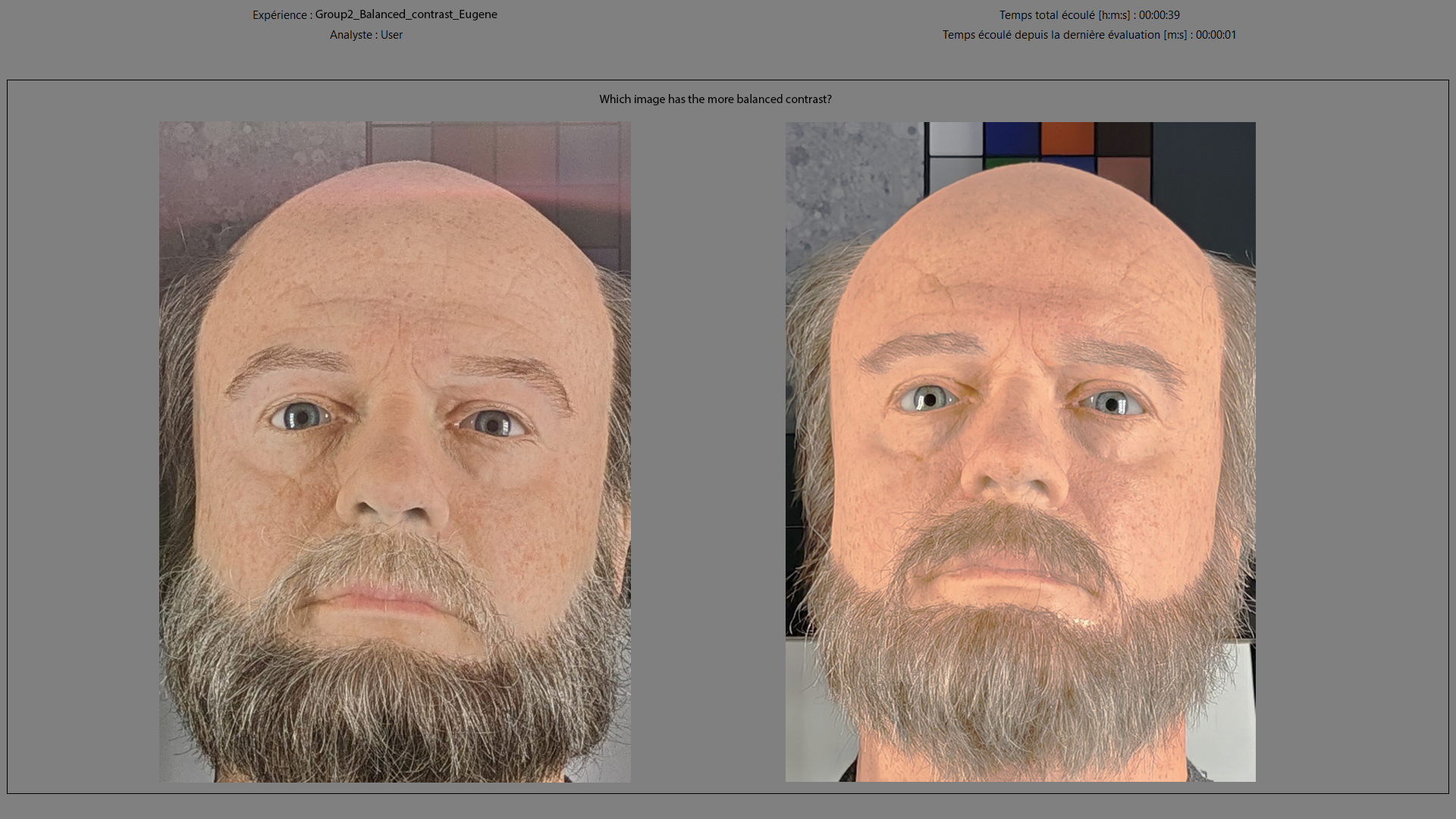}
    \caption{Example of the annotation tool used to acquire pairwise comparisons (Balanced contrast, Eugene)}    \label{fig:Example_Preference_Eugene}
\end{figure*}

\begin{figure*}[t]
    \centering     \includegraphics[width=0.99\textwidth]{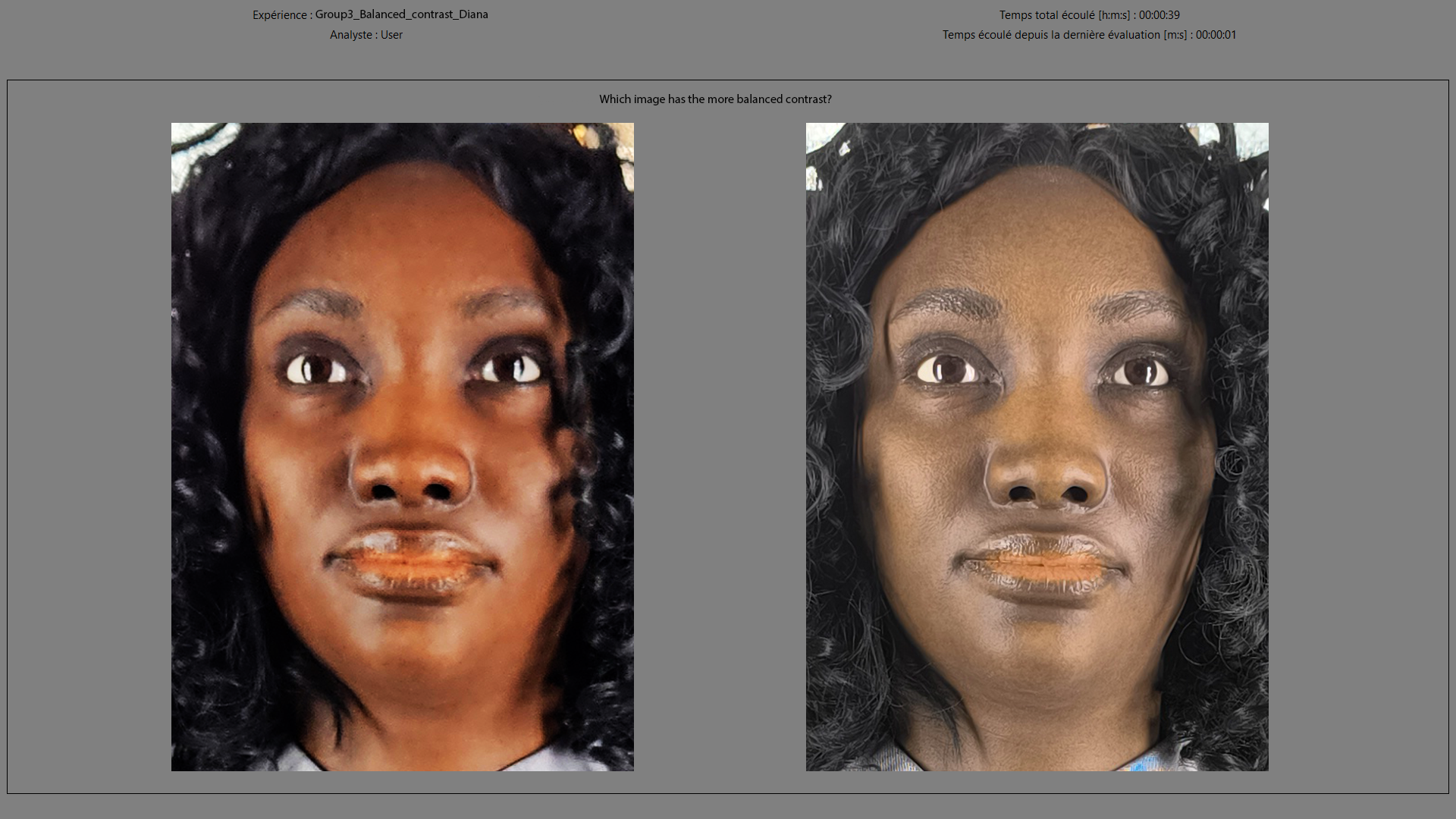}
    \caption{Example of the annotation tool used to acquire pairwise comparisons (Balanced contrast, Diana)}
    \label{fig:Example_Preference_Diana}
\end{figure*}

During the experiments, the annotators performed pairwise comparisons using a custom annotation tool designed to present two images side by side and record their judgments for different perceptual attributes. For each comparison, the annotators selected the image that best matched the target criterion, depending on the annotation task. Figures~\ref{fig:Example_overall_Eugene} and~\ref{fig:Example_overall_Diana} show example interfaces used for overall tonal contrast annotation. Figures~\ref{fig:Example_Eye_Eugene} and~\ref{fig:Example_Eye_Diana} are for eye contrast annotation. Figures~\ref{fig:Example_Micro_Eugene} and~\ref{fig:Example_Micro_Diana} illustrate example interfaces for micro-contrast annotation. Lastly, Figures~\ref{fig:Example_Preference_Eugene} and~\ref{fig:Example_Preference_Diana} show example interfaces for balanced contrast annotation. Before the main experiment, all annotators were provided with detailed annotation guidelines and received two hours of prior training to ensure a consistent understanding of the evaluation criteria. Each annotation session took approximately one hour per attribute. 
The total annotation time amounted to 160 hours across all annotations.

\subsection{JOD-Ordered Examples for Each Attribute}

\begin{figure*}[t]
    \centering
    \includegraphics[width=0.99\textwidth]{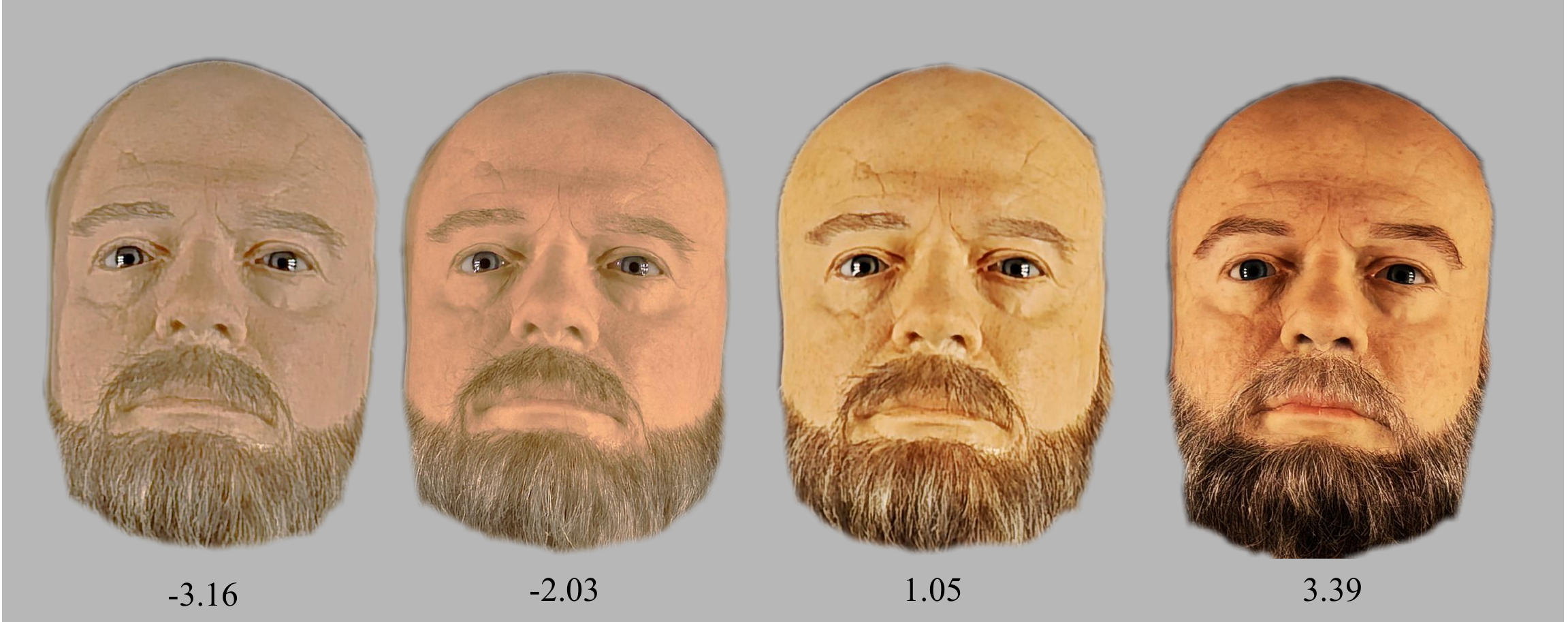}
    \caption{Examples of Eugene images sorted by JOD for overall contrast, from lower to higher values.}
    \label{fig:JOD_Overall_Eugene}
\end{figure*}

\begin{figure*}[t]
    \centering
    \includegraphics[width=0.99\textwidth]{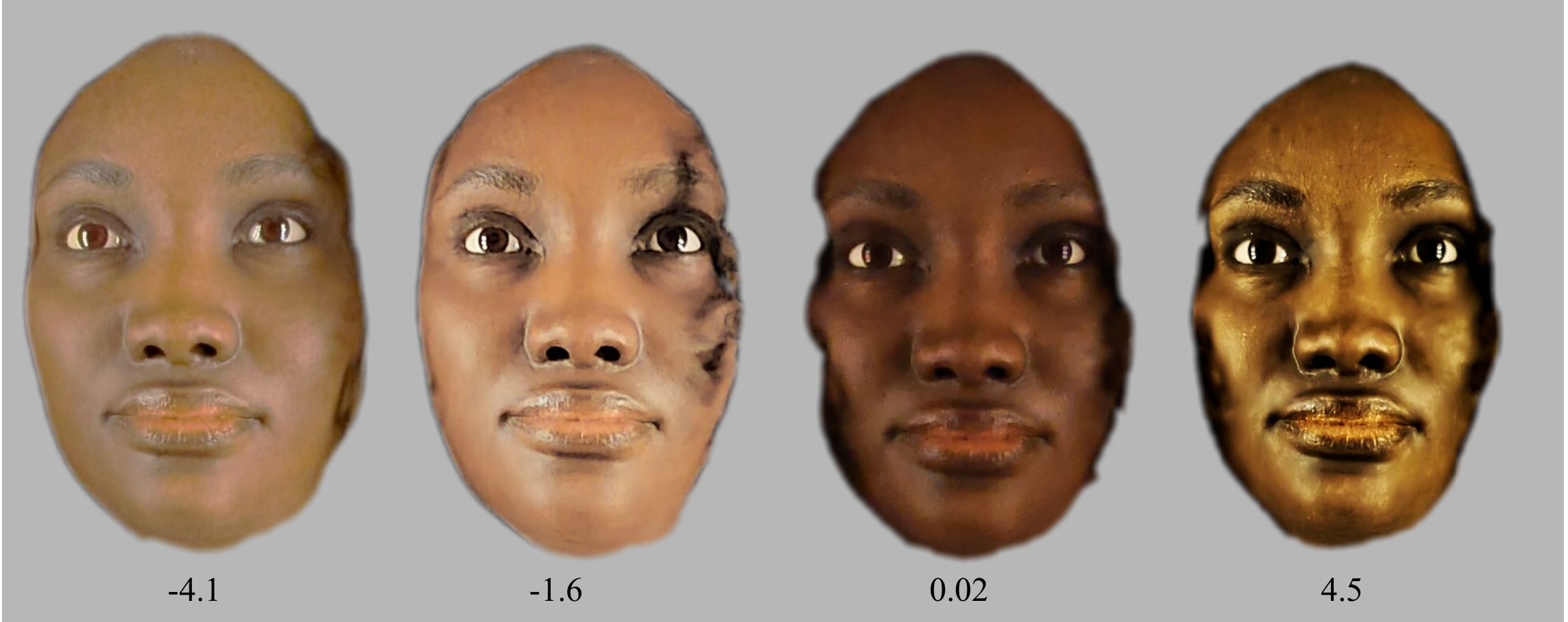}
    \caption{Examples of Diana images sorted by JOD for overall contrast, from lower to higher values.}
    \label{fig:JOD_Overall_Diana}
\end{figure*}

\begin{figure*}[t]
    \centering
    \includegraphics[width=0.99\textwidth]{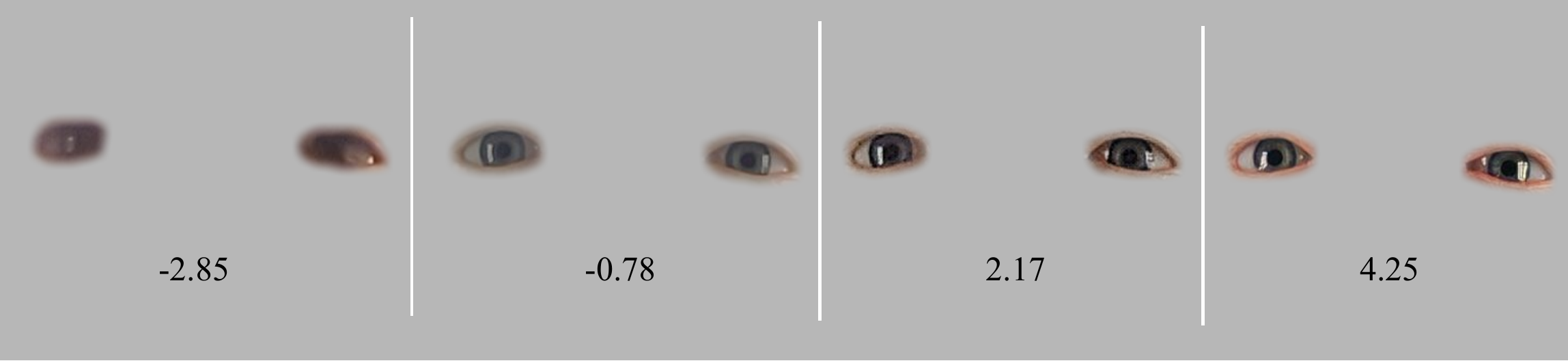}
    \caption{Examples of Eugene images sorted by JOD for eye contrast, from lower to higher values.}
    \label{fig:JOD_Eye_Eugene}
\end{figure*}

\begin{figure*}[t]
    \centering
    \includegraphics[width=0.99\textwidth]{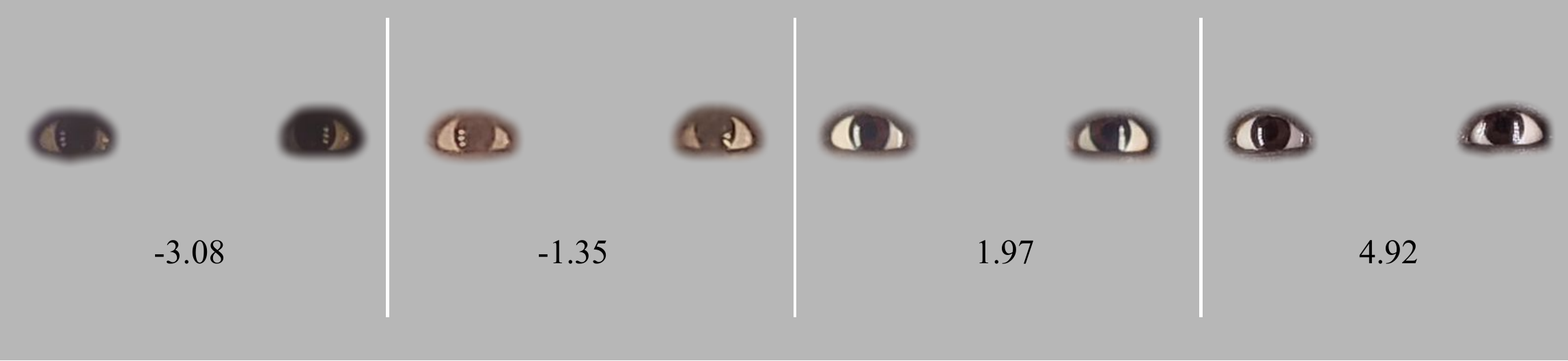}
    \caption{Examples of Diana images sorted by JOD for eye contrast, from lower to higher values.}
    \label{fig:JOD_Eye_Diana}
\end{figure*}

\begin{figure*}[t]
    \centering
    \includegraphics[width=0.99\textwidth]{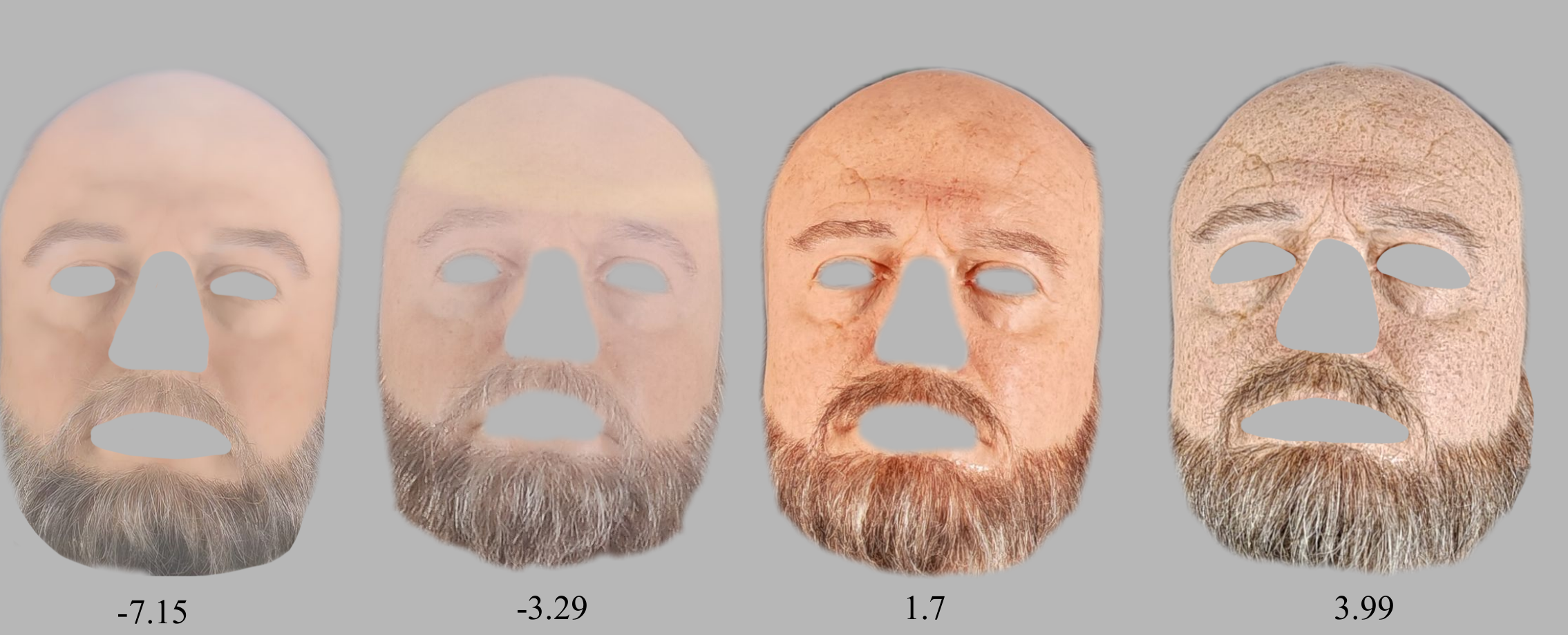}
    \caption{Examples of Eugene images sorted by JOD for micro-contrast, from lower to higher values.}
    \label{fig:JOD_Micro_Eugene}
\end{figure*}

\begin{figure*}[t]
    \centering
    \includegraphics[width=0.99\textwidth]{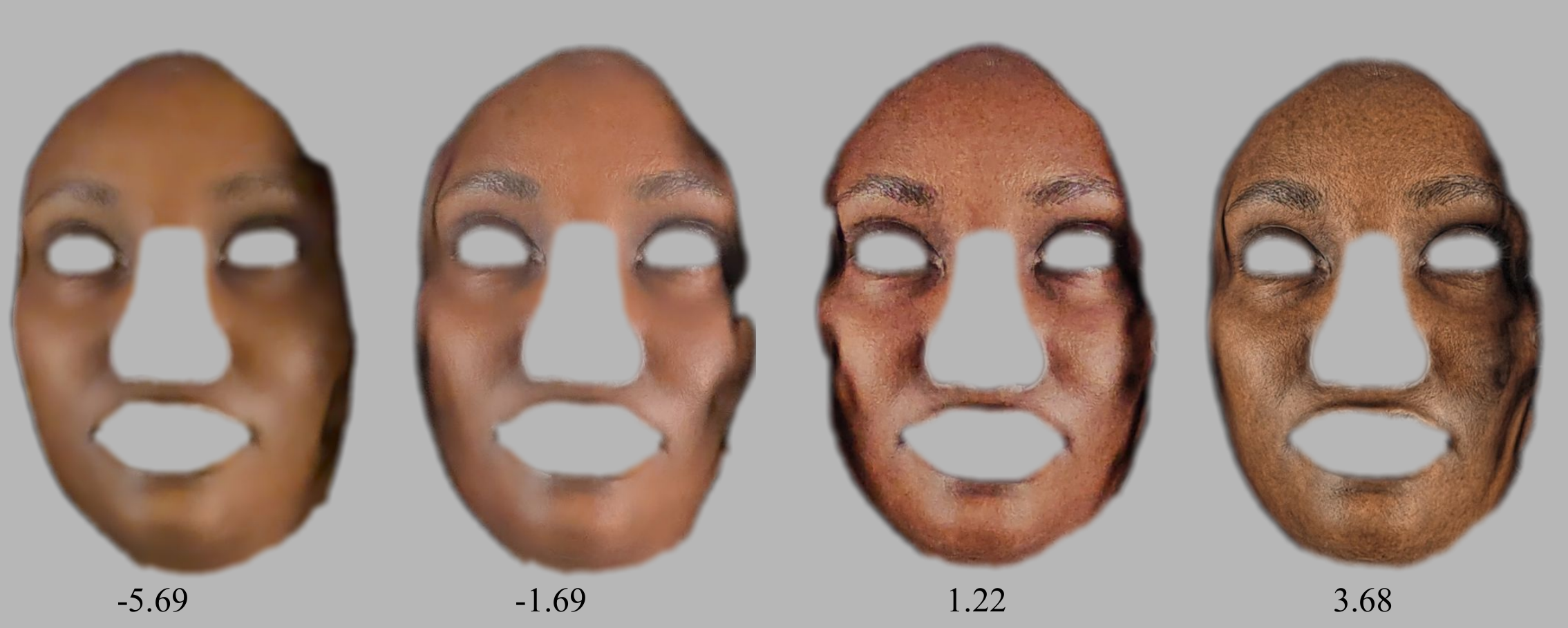}
    \caption{Examples of Diana images sorted by JOD for micro-contrast, from lower to higher values.}
    \label{fig:JOD_Micro_Diana}
\end{figure*}

\begin{figure*}[t]
    \centering
    \includegraphics[width=0.99\textwidth]{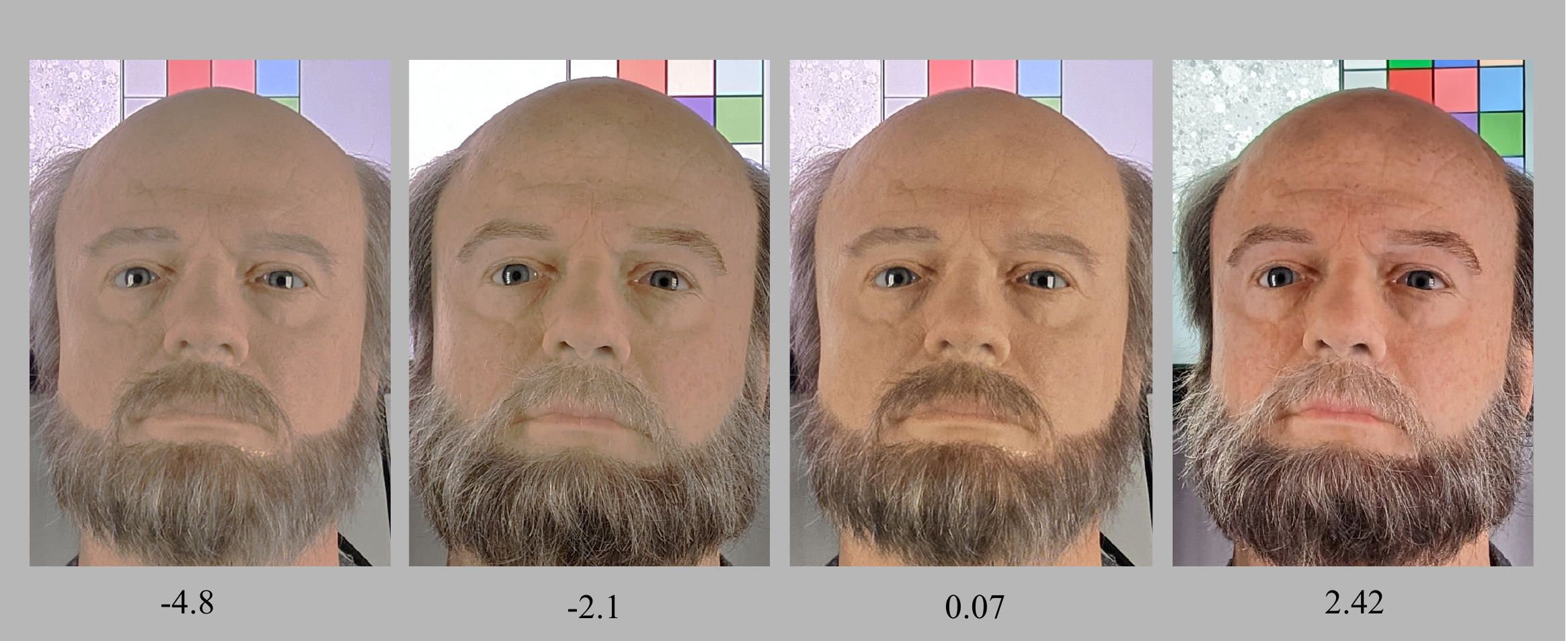}
    \caption{Examples of Eugene images sorted by JOD for balanced contrast preference, from lower to higher values.}
    \label{fig:JOD_Preference_Eugene}
\end{figure*}

\begin{figure*}[t]
    \centering
    \includegraphics[width=0.99\textwidth]{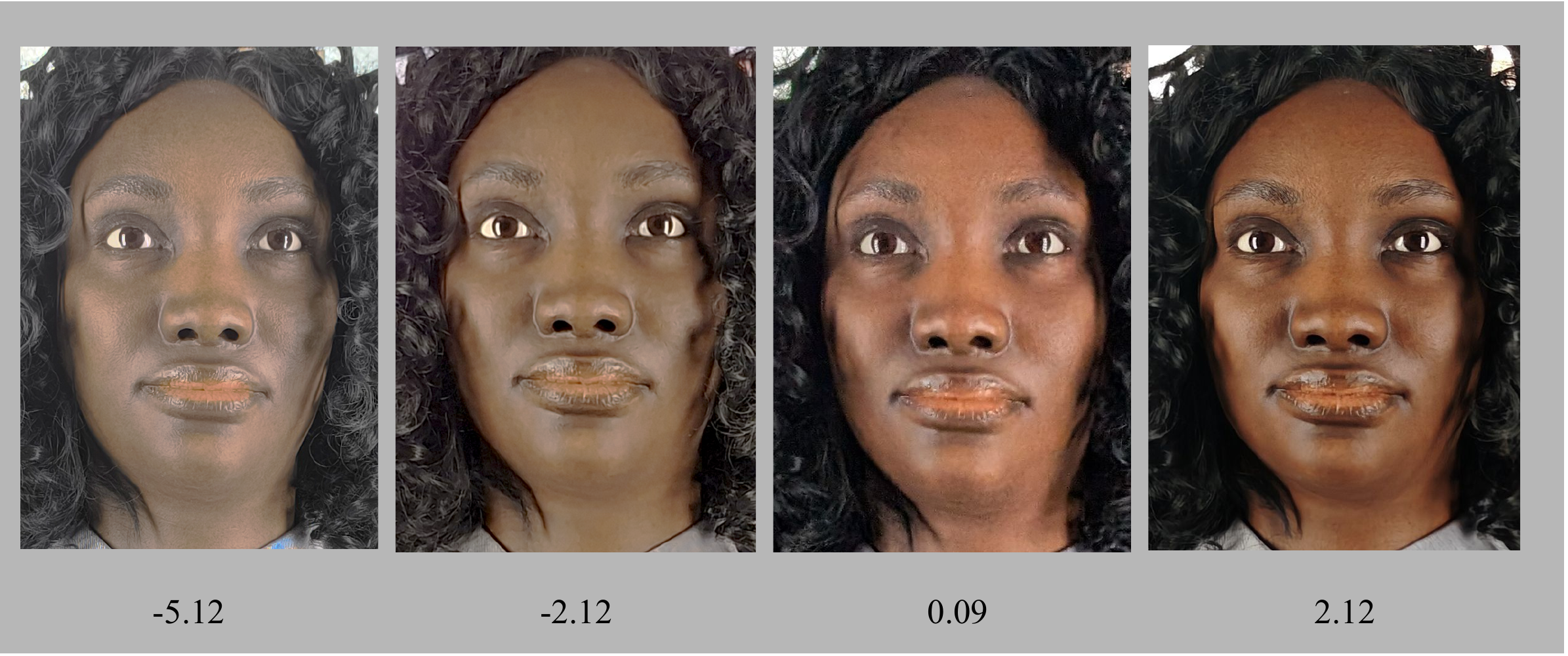}
    \caption{Examples of Diana images sorted by JOD for balanced contrast preference, from lower to higher values.}
    \label{fig:JOD_Preference_Diana}
\end{figure*}

To illustrate the perceptual scales recovered from the pairwise annotations, we show example images sorted by their JOD values for each attribute. In all figures, the examples are ordered from lower to higher JOD. For overall contrast, eye contrast, and micro-contrast, higher JOD corresponds to more of the corresponding contrast attribute. For balanced contrast preference, higher JOD indicates a stronger human preference for the corresponding image. This ordering allows a visual inspection of how each recovered perceptual scale varies across the dataset. Figures~\ref{fig:JOD_Overall_Eugene} and~\ref{fig:JOD_Overall_Diana} show overall contrast examples for Eugene and Diana. Figures~\ref{fig:JOD_Eye_Eugene} and~\ref{fig:JOD_Eye_Diana} show eye-contrast examples. Figures~\ref{fig:JOD_Micro_Eugene} and~\ref{fig:JOD_Micro_Diana} show micro-contrast examples. Finally, Figures~\ref{fig:JOD_Preference_Eugene} and~\ref{fig:JOD_Preference_Diana} show examples sorted by balanced contrast preference JOD.

\end{document}